%% file: main.tex
\documentclass{article}
\usepackage{microtype}
\usepackage{graphicx}
\usepackage{subfig}
\usepackage{float}
\usepackage{multirow}
\usepackage{booktabs}
\usepackage{paralist}
\usepackage{amsmath}
\usepackage{amssymb}
\usepackage{mathtools}
\usepackage{amsthm}
\usepackage{bm}
\usepackage[linesnumbered,ruled,vlined]{algorithm2e}
\usepackage{hyperref}
\usepackage{url}
\usepackage[capitalize,noabbrev]{cleveref}
\usepackage[textsize=tiny]{todonotes}

\input{savespace}

\usepackage[accepted]{icml2024}

\icmltitlerunning{Probabilistic Routing for Graph-Based Approximate Nearest Neighbor Search}

\input{macros}

\begin{document}

\twocolumn[
\icmltitle{Probabilistic Routing for Graph-Based Approximate Nearest Neighbor Search}

\input{author}

\icmlkeywords{nearest neighbor search, graph-based ANNS, probabilistic routing}

\vskip 0.3in
]



\printAffiliationsAndNotice{}  

\input{abstract}
\input{intro}
\input{related}
\input{problem}
\input{baseline}
\input{peos}
\input{analysis}
\input{exp}
\input{concl}


\input{ack}


\bibliography{references}
\bibliographystyle{icml2024}

\input{appendix}

\end{document}

%% file: savespace.tex
\usepackage[all=normal, floats, bibnotes, wordspacing, charwidths, indent, lists]{savetrees}

\skip\footins 8.1pt plus 4pt minus 2pt 

%% file: macros.tex
\newcommand{\size}[1]{| #1 |}
\newcommand{\norm}[1]{\Vert #1 \Vert}

\newcommand{\set}[1]{\{\, #1 \,\}}

\newcommand{\argmax}{\mathop{\rm argmax}}

\newcommand{\sgn}{sgn}
\newcommand{\erf}{erf}
\newcommand{\opt}{opt}

\newcommand{\mycomment}[1]{}

\theoremstyle{plain}
\newtheorem{theorem}{Theorem}[section]

\newtheorem{lemma}[theorem]{Lemma}

\theoremstyle{definition}
\newtheorem{definition}[theorem]{Definition}

\theoremstyle{remark}

\SetFuncSty{text}
\SetArgSty{text}
\SetKw{OR}{or}
\SetKw{AND}{and}
\SetKw{NOT}{not}
\SetKw{TRUE}{true}
\SetKw{FALSE}{false}
\SetKw{NULL}{nil}
\SetKw{Break}{break}
\SetKw{Continue}{continue}
\SetKwInOut{Input}{Input}
\SetKwInOut{Output}{Output}

\SetCommentSty{mycommfont}
\newcommand{\State}[1]{#1\;}
\newcommand{\StateCmt}[2]{
  #1 \tcc*[r]{#2}
}

%% file: author.tex


\icmlsetsymbol{equal}{*}

\begin{icmlauthorlist}
\icmlauthor{Kejing Lu}{aff1}
\icmlauthor{Chuan Xiao}{aff1,aff2}
\icmlauthor{Yoshiharu Ishikawa}{aff1}

\end{icmlauthorlist}

\icmlaffiliation{aff1}{Graduate School of Informatics, Nagoya University, Japan}
\icmlaffiliation{aff2}{Graduate School of Information Science and Technology, Osaka University, Japan}

\icmlcorrespondingauthor{Kejing Lu}{lu@db.is.i.nagoya-u.ac.jp}

%% file: abstract.tex
\begin{abstract}
  Approximate nearest neighbor search (ANNS) in high-dimensional spaces is a pivotal challenge in the field of machine learning. In recent years, graph-based methods have emerged as the superior approach to ANNS, establishing a new state of the art. Although various optimizations for graph-based ANNS have been introduced, they predominantly rely on heuristic methods that lack formal theoretical backing. This paper aims to enhance routing within graph-based ANNS by introducing a method that offers a probabilistic guarantee when exploring a node's neighbors in the graph. We formulate the problem as probabilistic routing and develop two baseline strategies by incorporating locality-sensitive techniques. Subsequently, we introduce PEOs, a novel approach that efficiently identifies which neighbors in the graph should be considered for exact distance calculation, thus significantly improving efficiency in practice. Our experiments demonstrate that equipping PEOs can increase throughput on commonly utilized graph indexes (HNSW and NSSG) by a factor of 1.6 to 2.5, and its efficiency consistently outperforms the leading-edge routing technique by 1.1 to 1.4 times. The code and datasets used for our evaluations are publicly accessible at \url{https://github.com/ICML2024-code/PEOs} .
\end{abstract}

%% file: intro.tex
\section{Introduction}
\label{sec:intro}

Nearest neighbor search (NNS) is the process of identifying the vector in a dataset that is closest to a given query vector. This technique has found widespread application in machine learning, with numerous solutions to NNS having been proposed. Given the challenge of finding exact answers, practical efforts have shifted towards approximate NNS (ANNS) for efficiency. While some approaches provide theoretical guarantees for approximate answers -- typically through locality-sensitive hashing (LSH) \cite{simhash,DatarIIM04:p-stable,falconn} -- others prioritize faster search speeds for a desired recall level by utilizing quantization~\cite{PQ,OPQ} or graph indexes~\cite{hnsw,nsg,nssg}.


Graph-based ANNS, distinguished by its exceptional empirical performance, has become the leading method for ANNS and is widely implemented in vector search tools and databases. During the indexing phase, graph-based ANNS constructs a proximity graph where nodes represent data vectors and edges connect closely located vectors. In the search phase, it maintains a priority queue and a result list while exploring the graph. Nodes are repeatedly popped from the priority queue until it is empty, calculating the distance from their neighbors to the query. Neighbors that are closer than the furthest element in the result list are added to the queue, and the results are accordingly updated.


To further improve the performance of graph-based ANNS, practitioners have introduced various empirical optimization techniques, including routing~\cite{TOGG-KMC,HCNNG,learn-to-route,FINGER}, edge selection~\cite{nssg,DiskANN}, and quantization~\cite{hvs,ngtqg}. The primary objective is to minimize distance calculations during neighbor exploration. However, most of these optimizations are heuristic, based on empirical observations (e.g., over 80\% of data vectors are less relevant than the furthest element in the results list and thus should be pruned before exact distance calculations~\cite{FINGER}), making them challenging to quantitatively analyze. Although analyses elucidate the effectiveness of graph-based ANNS~\cite{practice-to-theory, worst-case-of-graph}, they concentrate on theoretical aspects rather than empirical improvements.


In this paper, we investigate routing in graph-based ANNS, aiming to identify which neighbors should be evaluated for distance to the query during the search phase efficiently. Our objective is to bridge the theoretical and practical aspects of ANNS by providing a theoretical guarantee. While achieving an LSH-like guarantee for graph-based ANNS is challenging, we demonstrate that it is feasible to establish a probabilistic guarantee for exploring a node's neighbors. Specifically, we introduce probabilistic routing in graph-based ANNS: for a given top node $v$ in the priority queue, an error bound $\epsilon$, and a distance threshold $\delta$, any neighbor $u$ of $v$ with a distance to the query less than $\delta$ will be calculated for distance with a probability of at least $1-\epsilon$. Addressing the probabilistic routing problem yields several benefits: First, it ensures that ANNS explores the most promising neighbors (less than 20\% of all neighbors, as observed in \cite{FINGER}) with high probability, facilitates quantitative analysis of a search algorithm's effectiveness. Second, by devising probabilistic routing algorithms that accurately and efficiently estimate distances, we can significantly enhance practical efficiency. Third, the theoretical framework ensures consistent performance across different datasets, contrasting with heuristic approaches that may result in high estimation errors and consequently, lower recall rates.

To address the probabilistic routing problem, we initially integrate two foundational algorithms from existing research -- SimHash~\cite{simhash} and (reverse) CEOs~\cite{ceos} -- into graph-based ANNS. Subsequently, we introduce a novel routing algorithm, namely \textbf{P}artitioned \textbf{E}xtreme \textbf{O}rder \textbf{S}tatistics (PEOs), characterized by the following features:


(1) PEOs utilizes space partitioning and random projection techniques to estimate a random variable  which represents the angle between each neighbor and the query vector. By aggregating projection data from multiple subspaces, we substantially reduce the variance of the estimated random variable's distribution, thereby enhancing the accuracy of neighbor-to-query distance estimations.




(2) Through comprehensive analysis, we show that PEOs addresses the probabilistic routing problem within a user-defined error bound $\epsilon$ ($\epsilon \le 0.5$). The algorithm introduces a parameter $L$, denoting the number of subspaces in partitioning. An examination of $L$'s influence on routing enables us to identify an optimal parameter configuration for PEOs. Comparative analysis with baseline algorithms reveals that an appropriately selected $L$ value yields a variance of PEOs' estimated random variable than that obtained via SimHash-based probabilistic routing and the reverse CEOs-based probabilistic routing is a special case of PEOs with $L = 1$. 



(3) The implementation of PEOs is optimized using pre-calculated data and lookup tables, facilitating fast and accurate estimations. The use of SIMD further enhances processing speed, allowing for the simultaneous estimation of 16 neighbors and leveraging the data locality, a significant improvement over conventional methods that require accessing raw vector data stored disparately in memory.



Our experiments encompass tests on several publicly accessible datasets. By integrating PEOs with HNSW~\cite{hnsw} and NSSG~\cite{nssg}, two predominant graph indexes for ANNS, we achieve a reduction in the necessity for exact distance calculations by 70\% to 80\%, thereby augmenting queries per second (QPS) by 1.6 to 2.5 times under various recall criteria. Moreover, PEOs demonstrates superior performance to FINGER~\cite{FINGER}, a leading-edge routing technique, consistently enhancing efficiency by 1.1 to 1.4 times while reducing space costs.

%% file: related.tex
\section{Related Work}
\label{sec:related}
Various types of ANNS approaches have been proposed, encompassing tree-based approaches~\cite{cover_tree}, hashing-based approaches~\cite{AndoniI08:near-optimal-hashing, IPL, lei2019sublinear, R2LSH,falconn,Falconn++}, quantization based approaches~\cite{PQ, OPQ, imi, scann, SOAR}, learn-to-index-based approaches~\cite{Bliss, BATLearn}, and graph-based approaches~\cite{hnsw,DiskANN,nsg,nssg}. 
Among these, graph-based methods are predominantly considered state-of-the-art (SOTA). To enhance the efficiency of graph-based ANNS, optimizations can be broadly categorized into: (1) routing, (2) edge-selection, and (3) quantization, with these optimizations generally being orthogonal to one another. Given our focus on routing, we briefly review relevant studies in this domain. TOGG-KMC~\cite{TOGG-KMC} and HCNNG~\cite{HCNNG} employ KD trees to determine the direction of the query, thereby restricting the search to vectors within that specified direction. Despite fast estimation, it tends to yield suboptimal query accuracy, limiting its effectiveness. FINGER~\cite{FINGER} estimates the distance of each neighbor to the query. Specifically, for each node, it generates promising projected vectors locally to establish a subspace, and then applies collision counting, as in SimHash, to approximate the distance in each visited subspace. Learn-to-route~\cite{learn-to-route} learns a routing function, utilizing additional representations to facilitate optimal routing from the starting node to the nearest neighbor. In addition to these methods that accelerate ANNS itself, learning representations with a coarse granularity~\cite{kusupati2022matryoshka} may help reduce the dimensionality and yield a faster search. Such method is orthogonal to the techniques proposed in this paper and can be further applied to speed up ANNS.

%% file: problem.tex
\section{Problem Definition}
\label{sec:problem}
\begin{definition} [Nearest Neighbor Search (NNS)]
  Given a query vector $\bm{q} \in \mathbb{R}^d$ and a dataset of vectors $\mathcal{O}$, find the vector $\bm{o}^{\ast} \in \mathcal{O}$ such that $dist(\bm{q}, \bm{o}^{\ast})$ is the smallest. 
\end{definition}
For distance function $dist(\cdot, \cdot)$, two widely utilized metrics are $\ell_2$ distance and angular distance. Maximum inner product search (MIPS), closely related to NNS, aims to identify the vector that yields the maximum inner product with the query vector. We elaborate on extension to MIPS in Appendix~\ref{extension-to-mips} and plan to assess its performance in future work.


There is a significant interest not only in identifying a singular nearest neighbor but also in locating the top-$K$ nearest neighbors, a task referred to as $K$-NN search. It is a prevailing view that calculating exact NNS results poses a considerable challenge, whereas determining approximate results suffices for addressing many practical applications~\cite{qin2021high}. Notably, many SOTA ANNS algorithms leverage a graph index, where each vector in $\mathcal{O}$ is linked to its nearby vectors.


\begin{algorithm}[!t]
  \small
  \caption{Graph-based ANNS with routing}
  \label{alg:anns}  
  \Input{Query $\bm{q}$, \# results $K$, graph index $G$}
  \Output{$K$-NN of $\bm{q}$}
  \StateCmt{$R \gets \emptyset$}{an ordered list of results, $\size{R} \leq efs$}
  \StateCmt{$P \gets \set{$ entry node $v_0 \in G}$}{a priority queue}
  \While{$P \neq \emptyset$}{
    \State{$v \gets P.pop()$}
    \ForEach{unvisited neighbor $u$ of $v$}{
      \lIf{$\size{R} < efs$}{$\delta \gets \infty$} \label{alg:ln:delta-1}
      \lElse{$p \gets R[efs]$, $\delta \gets dist(\bm{p}, \bm{q})$} \label{alg:ln:delta-2}
      \If{RoutingTest($\bm{u}, \bm{v}, \bm{q}, \delta$) = \TRUE}{ \label{alg:ln:routing}
        \If{$dist(\bm{u}, \bm{q}) < \delta$}{
          \State{$R.push(u)$, $P.push(u)$}
        }
      }
    }
  }
  \Return($\set{R[1], \ldots, R[K]}$)
\end{algorithm}


The construction of a graph index can be approached in various ways, e.g., through a KNN graph~\cite{dong2011efficient}, HNSW~\cite{hnsw}, NSG~\cite{nsg}, and the improved variant NSSG~\cite{nssg}. Among these, HNSW stands out as the most extensively adopted model, implemented in many ANNS platforms such as Faiss and Milvus. Given a graph index $G = (V, E)$ built upon $\mathcal{O}$, traversal of this graph enables the discovery of ANNS results, as delineated in Algorithm~\ref{alg:anns}. The search initiates from an entry node $v_0 \in G$, maintaining $R$, an ordered list that contains no more than $efs$ ($efs \ge K$) -- a list size parameter -- results identified thus far. Neighbors of the entry node in the graph are examined against $\bm{q}$ for proximity and added to a priority queue if they are closer to the query than the most distant element in $R$ or if $R$ is not full. Nodes are popped from the priority queue to further explore their adjacent nodes in the graph, until the priority queue becomes empty. It is noteworthy that practical search operations may extend beyond those depicted in Algorithm~\ref{alg:anns}, e.g., by employing pruning strategies to expedite the termination of the search process~\cite{hnsw}. Algorithms designed for graph-based ANNS with angular distance can be adapted to accommodate $\ell_2$ distance, since calculating the $\ell_2$ distance between $\bm{q}$ and $\bm{o} \in \mathcal{O}$ merely involves determining the angle between them during graph traversal: $\ell_2(\bm{q}, \bm{o})^2 = \norm{\bm{q}}^2 + \norm{\bm{o}}^2 - 2\bm{q}^{\top}\bm{o}$, where $\norm{\bm{q}}$ is fixed and $\norm{\bm{o}}$ can be calculated beforehand.

While naive graph exploration entails calculating the exact distance for all neighbors, a \emph{routing} test can be applied to assess whether a neighbor warrants exact distance calculation. An efficient routing algorithm can substantially enhance the performance of graph-based ANNS. We study routing algorithms with the following probability guarantee. 


\begin{definition} [Probabilistic Routing]
  \label{def:probabilistic-routing}
  Given a query vector $\bm{q}$, a node $v$ in the graph index, an error bound $\epsilon$, and a distance threshold $\delta$, for an arbitrary neighbor $u$ of $v$ such that $dist(\bm{u}, \bm{q}) < \delta$, if a routing algorithm returns true for $u$ with a probability of at least $1-\epsilon$, then the algorithm is deemed to be $(\delta, 1-\epsilon)$-routing. 
\end{definition}

Given our interest in determining whether a neighbor has the potential to refine the temporary results, our focus narrows to the case when $\delta$ equals the distance between $\bm{q}$ and the most distant element in the result list $R$, with $\size{R} = efs$ (Lines~\ref{alg:ln:delta-1} --~\ref{alg:ln:delta-2}, Algorithm~\ref{alg:anns}).




%% file: baseline.tex
\section{Baseline Algorithms}
\label{sec:baseline}
To devise a baseline algorithm for probabilistic routing, we adapt SimHash~\cite{simhash} and CEOs~\cite{ceos} for routing test, which were designed for the approximation with recall guarantee for ANNS and MIPS, respectively. 


\subsection{SimHash Test}
\label{sec:simhash}
SimHash, a classical random projection-based LSH method for the approximation of angular distance, has the following result~\cite{simhash}. 
\begin{lemma}
  \label{lem:simhash}
  \textbf{(SimHash)} Given $\bm{u}$, $\bm{q}$, and $m$ random vectors $\set{\bm{a}_i}^m_{i=1} \sim \mathcal N(0, I^d)$, the angle $\theta$ between $\bm{u}$ and $\bm{q}$ can be estimated as 
  \begin{equation}
    \hat \theta = \frac{\pi}{{m}}\sum\limits_i {[{\mathop{\rm \sgn}} (\bm{u}^{\top} \bm{a}_i) \ne {\mathop{\rm \sgn}} (\bm{q}^{\top} \bm{a}_i)]}.
  \end{equation}
\end{lemma}
Based on the above lemma, we design a routing test: 
\begin{equation}
  \label{eq:simhash-criterion}
  {\rm{\textbf{SimHash}}} \; {\rm{\textbf{Test:}}} \quad \#Col(\bm{u}, \bm{q}) \ge T_{\epsilon}^{\text{SimHash}}(\bm{u}, \bm{q}, \delta, m)
\end{equation}
where $\#Col$ denotes the collision number of the above random projection along $\set{\bm{a}_i}^m_{i=1}$, and $T_{\epsilon}^{\text{SimHash}}(\bm{u}, \bm{q}, \delta, m)$ denotes a threshold determined by $\epsilon$ and $\delta$. A neighbor $u$ of $v$ passes the routing test iff. it satisfies the above condition. By careful setting of the threshold, we can obtain the desired probability guarantee. Besides probabilistic routing, SimHash has also been used in FINGER~\cite{FINGER} to speed up graph-based ANNS, serving as one of its building blocks but utilized in a heuristic way. 

It can be seen that the result of the above SimHash test is regardless of $v$. This means that if a node $u$ fails in a test, it will never be inserted into the result set or the priority queue. Since this will compromise the recall of ANNS, we design the following remedy by regarding the neighbors of $v$ as residuals w.r.t. $v$: for each neighbor $u$ of $v$, let $\bm{e} = \bm{u} - \bm{v}$, and this residual can be associated with the edge $e$ from $v$ to $u$ in the graph index. $\bm{e}$ is then used instead of $\bm{u}$ in the SimHash test. As such, the routing test becomes dependent on $\bm{v}$, and a threshold $T_{\epsilon}^{\text{SimHash}}(\bm{u}, \bm{v}, \bm{q}, \delta, m)$ can be derived by the Hoeffding's inequality applied to the Binominal distribution. Hence a neighbor $u$ may fail to pass the test w.r.t. $v$ but succeed in the test w.r.t. node $v'$ in the graph index. Moreover, to model  the angle between $\bm{e}$ and $\bm{q}$ in the routing test, we normalize $\bm{q}$ to $\bm{q}'$ (and optionally, $\bm{e}$ to $\bm{e}'$) to simplify calculation. 
We discuss algorithms in the context of the above remedy hereafter.

\subsection{RCEOs Test}
\label{sec:rceos}
In the realm  of LSH, Andoni et al. proposed Falconn~\cite{falconn} for angular distance whose basic idea is to find the closest or furthest projected vector to the query and record such vector as a hash value, leading to a better search performance than SimHash. Pham and Liu~\cite{Falconn++} employed Concomitants of Extreme Order Statistics (CEOs) \cite{ceos} to record the minimum or maximum projection value, further improving the performance of Falconn. By swapping the roles of query and data vectors in CEOs, we have Reverse CEOs (RCEOs) and it can be applied to graph-based ANNS with probabilistic routing:  
\begin{lemma}
  \label{lem:rceos}
  \textbf{(RCEOs)} Given two normalized vectors $\bm{e}'$, $\bm{q}'$, and $m$ random vectors $\{\bm{a}_i\}^m_{i=1} \sim \mathcal N(0, I^d)$, and $m$ is sufficiently large, assuming that $\bm{a}_1 = {\argmax_{\bm{a}_i}} |\bm{e}'^{\top}\bm{a}_i|$, we have the following result:
  \begin{equation}
    \label{eq:rceos}
    \bm{q}'^{\top}\bm{a}_1 \sim \mathcal N({\mathop{\rm sgn}} (\bm{e}'^{\top}\bm{a}_1) \cdot \bm{e}'^{\top}\bm{q}'\sqrt {2\ln m}, 1 - (\bm{e}'^{\top} \bm{q}')^2). 
  \end{equation}
\end{lemma}
Despite an asymptotic result, it has been shown that $m$ does not need to be very large ($\geq 128$) to ensure an enough closeness to the objective normal distribution~\cite{ceos}. 
   
Based on Lemma~\ref{lem:rceos}, we design a routing test: 
\begin{equation}
  \label{eq:ceos-criterion}
  {\rm{\textbf{RCEOs}}} \; {\rm{\textbf{Test:}}} \quad \bm{q}'^\top \bm{a}_1 \ge T_{\epsilon}^{\text{RCEOs}}(\bm{u}, \bm{v}, \bm{q}, \delta, m)
\end{equation}
where $\bm{a}_1 = {\argmax_{\bm{a}_i}} |\bm{e}'^{\top}\bm{a}_i|$ and $T_{\epsilon}^{\text{RCEOs}}(\bm{u}, \bm{v}, \bm{q}, \delta, m)$ denotes a threshold related to $\epsilon$ and $\delta$.

%% file: peos.tex
\section{Partitioned Extreme Order Statistics (PEOs)}
\label{sec:peos}


\subsection{Space Partitioning}
In PEOs, the data space is partitioned into $L$ subspaces. Let $M \subseteq \mathbb{R}^d$ be the original data space. We decompose $M$ into $L$ orthogonal $d'$-dimensional subspaces $M_1, M_2, \ldots, M_L$, $d' = d/L$. When $L > 1$, we can significantly decrease the variance of the normal distribution in RCEOs (\eqref{eq:rceos}), hence delivering a better routing test. Specifically, (1) RCEOs test is a special case of PEOs test with $L = 1$, and (2) by choosing an appropriate $L$ ($L > 1$), PEOs test outperforms SimHash test while RCEOs test cannot (Appendix~\ref{sec:different-tests}). 




\subsection{PEOs Test}
\label{sec:process-of-peos}
Following the space partitioning, for each neighbor $u$ of $v$, $\bm{e}$ is partitioned to $[\bm{e}_1, \bm{e}_2, \ldots, \bm{e}_L]$, where $\bm{e}_i$ is the sub-vector of $\bm{e}$ in $M_i$. The PEOs test consists of the following steps. 

(1) (\textbf{Orthogonal Decomposition})
We decompose $\bm{e}$ as $\bm{e} = \bm{e}_{reg} + \bm{e}_{res}$ such that $\bm{e}_{reg} \perp \bm{e}_{res}$ and the direction of $\bm{e}_{reg}$ is determined as follows: 
\begin{equation}
  \frac{\bm{e}_{reg}}{\norm{\bm{e}_{reg}}} = [\frac{\bm{e}_1}{\sqrt{L} \norm{\bm{e}_1}}, \frac{\bm{e}_2}{\sqrt{L} \norm{\bm{e}_2}}, \ldots, \frac{\bm{e}_L}{\sqrt{L} \norm{\bm{e}_L}}].
\end{equation}
We call $\bm{e}_{reg}$ the regular part of $\bm{e}$ and $\bm{e}_{res}$ the residual part of $\bm{e}$. Besides, we introduce two weights $w_{reg}$ and $w_{res}$ such that $w_{reg}= \norm{\bm{e}_{reg}} / \norm{\bm{e}}$ and $w_{res}= \norm{\bm{e}_{res}} / \norm{\bm{e}}$.

(2) (\textbf{Generating Projected Vectors}) In each $M_i$, we independently generate $m$ projected vectors $\{\bm{a^i_j}\}^m_{j=1}$, where $\bm{a^i_j} \sim \mathcal N(0,I^{d' \times d'})$. In the original space $M$, we independently generate $m$ projected vectors $\{\bm{b_j}\}^m_{j=1}$ such that $\bm{b_j} \sim \mathcal N(0,I^{d \times d})$. 

(3) (\textbf{Collection of Extreme Values}) In each $M_i$, we collect $L + 1$ extreme values that yield the greatest inner products with the projected vectors: $e[i] = {\rm{sgn}}(\bm{e}_i^{\top}\bm{a^i_j})j$, where $j = \argmax_j |\bm{e}_i^{\top}\bm{a^i_j}|$, and $e[0] = {\rm{sgn}}(\bm{e}_{res}^{\top}\bm{b_j})j$, where $j = \argmax_j |\bm{e}_{res}^{\top}\bm{b_j}|$.

(4) (\textbf{CDF of Normal Distribution}) Let $\mathcal N^{\bm{e},x}_{\min}$ be the following normal distribution associated with $\bm{e}$:
\begin{equation}
  \label{eq:N_ex_min}
  \mathcal N^{\bm{e},x}_{\min} = \mathcal N(x\sqrt{2L\ln m} , w_{reg}^2 + Lw_{res}^2 - \frac{L x^2}{L+1})
\end{equation}
where $0 < x < 1$. Let $F_{\bm{e},x}$ be the CDF of $\mathcal N^{\bm{e},x}_{\min}$. We define $F^{-1}_{\bm{e}}(x,z)$ such that $F_{\bm{e},x}(F^{-1}_{\bm{e}}(x,z)) = z$, where $0 < z \le 0.5$. Note that $F^{-1}_{\bm{e}}$ can be well-defined since $F_{\bm{e},x}(z)$ is a monotone function of $x$ when $z$ is fixed. When setting $z = \epsilon$, we write $F^{-1}_{\bm{e},\epsilon}(x) = F^{-1}_{\bm{e}}(x,\epsilon)$.


(5) (\textbf{Query Projection}) Given query $\bm{q}$, we normalize it to $\bm{q}'$ and calculate the inner products with the projected vectors to obtain two values $H_1(\bm{e})$ and $H_2(\bm{e})$ w.r.t. $\bm{e}$: 
\begin{equation}
  H_1(\bm{e}) = \sum\limits_i {\rm{sgn}}(e[i]) (\bm{q}_i'^{\top} \bm{a}^i_{|e[i]|}),
\end{equation}
\begin{equation}
  H_2(\bm{e}) = {\rm{sgn}}(e[0]) (\bm{q}'^{\top} \bm{b}_{|e[0]|}).
\end{equation}

(6) (\textbf{Routing Test}) With $H_1(\bm{e})$ and $H_2(\bm{e})$, we calculate $A_r(\bm{e})$ as follows for $\ell_2$ distance: 
\begin{equation}
\label{eq:A_r(e)}
  A_r(\bm{e}) = \frac{\norm{\bm{u}}^2/2 - r -\bm{v}^{\top}\bm{q}}{\norm{\bm{q}} \norm{\bm{e}}}
\end{equation}
where $r = \norm{\bm{p}}^2/2 - \bm{p}^{\top}\bm{q}$ and $p$ is the furthest element to $\bm{q}$ in the temporary result list $R$. It is easy to see that $\delta^2 - 2r = \norm{\bm{q}}^2$ when $\delta$ captures the $\ell_2$ distance from $\bm{q}$ to the furthest element in $R$. For angular distance, we remove the norms $\norm{\bm{u}}^2/2$ and $\norm{\bm{p}}^2/2$ in $r$ to obtain $A_r(\bm{e})$. 

With $A_r(\bm{e})$, we design a routing test for $u$: If $A_r(\bm{e}) \ge 1$, it returns false. If $A_r(\bm{e}) \le 0$, it returns true. In case $0 < A_r(\bm{e}) < 1$, we calculate $H(\bm{e})$ and $T_{r}(\bm{e})$ as follows:
\begin{equation}
  H(\bm{e}) = w_{reg} H_1(\bm{e}) + \sqrt{L}w_{res} H_2(\bm{e}),
\end{equation}
\begin{equation}
  \label{eq:LB}
  T_{r}(\bm{e}) = F^{-1}_{\bm{\bm{e}},\epsilon}(A_r(\bm{e})).
\end{equation}
Then, the test returns true iff. the following condition is met. 
\begin{equation}
  \label{eq:criterion}
  \quad H(\bm{e}) - T_{r}(\bm{e}) \ge 0.
\end{equation}

\subsection{Implementation of PEOs}
\label{sec:peos-implementation}
\begin{figure}[!t]
  \centering
  \includegraphics[width=\linewidth]{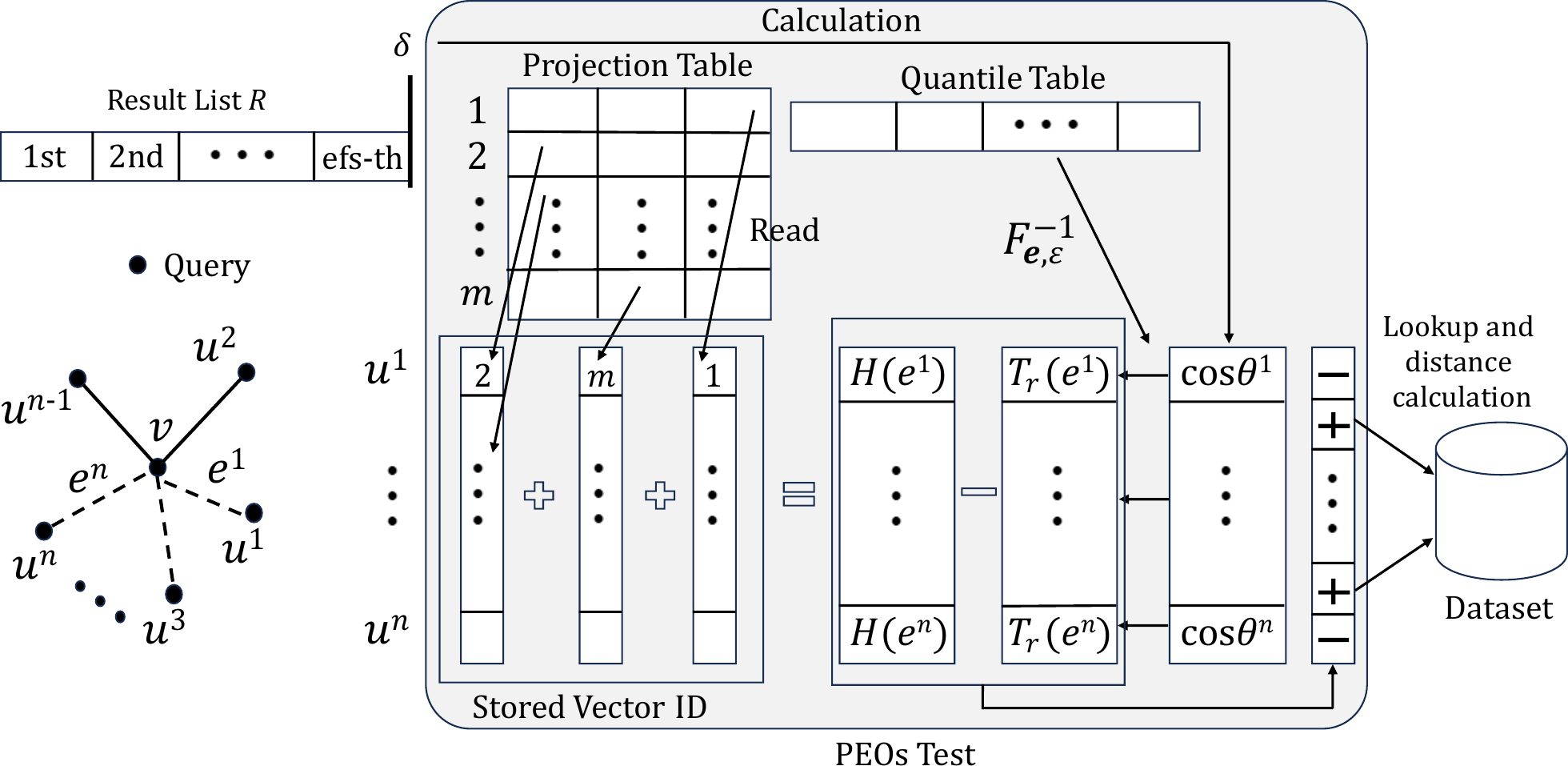}
  \caption{Illustration of the PEOs test. There are $n$ neighbors of $v$. $\theta^1, \ldots, \theta^n$ denote the angles between $\bm{e}^1, \ldots, \bm{e}^n$ and $\bm{q}$, respectively. $u^2$ and $u^{n-1}$ pass the test (indicated by ``+''). We access their raw vectors from the dataset and calculate their distances to $\bm{q}$.}
  \label{fig:illustration}
\end{figure}

In the PEOs test, Steps (1), (2), and (3) can be pre-calculated during index construction. Since space partitioning may result in unbalanced norms in subspaces, we can optionally permute the dimensions so that the norms of all $\bm{e}_i$'s ($1 \le i \le L, e \in E$) in the graph are as close to each other as possible (Appendix~\ref{sec:dimension-permutation}). Such permutation does not affect the topology of the graph or the theoretical guarantee of PEOs. Steps (4), (5), and (6) are calculated during the search. $H(\bm{e})$ and $T_{r}(\bm{e})$ can be calculated efficiently because (1) $\bm{q}_i'^{\top} \bm{a}^i_{|e[i]|}$ and $\bm{q}'^{\top} \bm{b}_{|e[0]|}$ can be calculated for $\bm{q}$ only once and stored in a projection table, and (2) although the online calculation of $F^{-1}_{\bm{e},\epsilon}(x)$ is costly, we can build a lookup table containing the quantiles corresponding to the different values of the variance in $\mathcal N^{\bm{e},x}_{\min}$ since such variance is bounded. From the lookup table, we can choose a quantile slightly smaller than the true quantile by employing the monotonicity, which does not affect the correctness of the probability guarantee. An illustration of the implementation of the PEOs test is depicted in Figure~\ref{fig:illustration}, with pseudo-code given in Algorithm~\ref{alg:peos}. 

Another optimization is the use of SIMD, where 16 edges can be processed at a time. This will significantly accelerate ANNS, because the raw vectors of neighbors are stored separately in the memory and loading them into CPU is costly. 

\begin{algorithm}[!t]
  \small
  \caption{PEOs Test}
  \label{alg:peos}
  \Input{query $\bm{q}$, edge $e = (v, u)$, threshold $r$, projected vectors $\set{\bm{a}^i_j}$ and $\set{\bm{b}_j}$ ($1 \le i \le L$, $1 \le j \le m$), quantile table $Q$} 
  \Output{whether $u$ passes the routing test}
  \State{Calculate $A_{r}(\bm{e})$}
  \lIf{$A_{r}(\bm{e}) \ge 1$}{\Return(\FALSE)}
  \lIf{$A_{r}(\bm{e}) \le 0$}{\Return(\TRUE)}
  \State{Normalize $\bm{q}$ to $\bm{q}'$}
  \StateCmt{Build a projection table $T_I$ for all $\bm{q}_i'^{\top}\bm{a}^i_j$ and $\bm{q}'^{\top}\bm{b}_j$} {only once and used throughout the search}
  \State{Calculate $H(\bm{e})$ with $T_I$}
  \State{Calculate $T_{r}(\bm{e})$ with $Q$}
  \lIf{$H(\bm{e}) \ge T_{r}(\bm{e})$}{
    \Return(\TRUE)
  }
  \lElse{
    \Return(\FALSE)
  }
\end{algorithm}

%% file: analysis.tex
\section{Analysis of PEOs}
\label{sec:analysis-of-peos}

\subsection{Probability Guarantee of PEOs}
\label{sec:guarantee-of-peos}
We assume $\norm{\bm{e}} = \norm{\bm{q}} = 1$ and let $\theta$ denote the angle between them. By the independence of projected vectors in different subspaces and the result in Lemma~\ref{lem:rceos}, we have $H_1(\bm{e}) \sim \mathcal N^{\theta}_{\bm{e}, \bm{q}}$, where $\mathcal N^{\theta}_{\bm{e}, \bm{q}}$ is defined as follows: 
\begin{equation}
  \label{eq:combination}
  \mathcal N^{\theta}_{\bm{e}, \bm{q}} = \mathcal N(\eta \sum\limits_i\norm{\bm{q}_i} \cos \theta_i , 1 - \sum\limits_i\norm{\bm{q}_i}^2 \cos^2 \theta_i)
\end{equation}
where $\eta=\sqrt{2\ln m}$ and $\theta_i$ denotes the angle between $\bm{e}_i$ and $\bm{q}_i$. Next, we analyze the relationship between $\theta$ and $\mathcal N^{\theta}_{\bm{e}, \bm{q}}$. To this end, we introduce the following definition.

\begin{definition}
  We define two partial orders $\prec$ and $\preceq$ such that, for two normal distributions $\mathcal N(\mu_1, \sigma_1^2)$ and $\mathcal N(\mu_2, \sigma_2^2)$, $\mathcal N(\mu_1, \sigma_1^2) \prec \mathcal N(\mu_2, \sigma_2^2)$ iff.
  $\mu_1 \le \mu_2$ and $\sigma_1^2 \ge \sigma_2^2$, and 
  $\mathcal N(\mu_1, \sigma_1^2) \preceq \mathcal N(\mu_2, \sigma_2^2)$ iff. $\mu_1 = \mu_2$ and $\sigma_1^2 \ge \sigma_2^2$.
\end{definition}

With the notations defined above, we want to find an appropriate normal distribution $\mathcal {\tilde N}^{\theta}_{\bm{e}, \bm{q}}$ such that $\mathcal {\tilde N}^{\theta}_{\bm{e}, \bm{q}} \prec \mathcal N^{\theta}_{\bm{e}, \bm{q}}$ ($\mathcal {\tilde N}^{\theta}_{\bm{e}, \bm{q}} \preceq \mathcal N^{\theta}_{\bm{e}, \bm{q}}$ is more favorable) for all adequate pairs $(\bm{e}, \bm{q})$'s. We define $\mathcal {\tilde N}^{\theta}_{\bm{e}, \bm{q}}$ as follows, where $e_{\min}= \min_{1 \le i \le L} {\norm{\bm{e}_i}}$ and $e_{\max}= \max_{1 \le i \le L} {\norm{\bm{e}_i}}$:
\begin{equation}
  \mathcal {\tilde N}^{\theta}_{\bm{e}, \bm{q}} = \mathcal N(\frac{{(\cos \theta + (e_{\min} - e_{\max}) \sum\limits_i{\norm{\bm{q}_i}}  }) \eta}{e_{\max}}, 1- \cos^2 \theta ).
\end{equation}
Then, we have the following lemma.
\begin{lemma}
  \label{lemma:comparison-of-distributions}
  $\mathcal {\tilde N}^{\theta}_{\bm{e}, \bm{q}} \prec \mathcal N^{\theta}_{\bm{e}, \bm{q}}$. 
  When $\norm{\bm{e}_1} = \cdots = \norm{\bm{e}_L}$, $\mathcal {\tilde N}^{\theta}_{\bm{e}, \bm{q}} = \mathcal N(\cos \theta \sqrt{2 L \ln m}, 1- \cos^2 \theta) \preceq \mathcal N^{\theta}_{\bm{e}, \bm{q}}$.
\end{lemma}

From Lemma~\ref{lemma:comparison-of-distributions}, we can see that, for the case when $e_{\max} - e_{\min}$ is large, we can only get a loose lower bound of $\mathbb E [\mathcal N^{\theta}_{\bm{e}, \bm{q}}]$ due to the impact of unknown $\theta_i$'s (although it is possible to get a strict lower bound by solving a linear programming, the calculation is too costly for a fast test), while the estimation of $\mathbb E [\mathcal N^{\theta}_{\bm{e}, \bm{q}}]$ is always accurate when $\norm{\bm{e}_1} = \cdots = \norm{\bm{e}_L}$ holds. This explains why we decompose vector $\bm{e}$ into $\bm{e}_{reg}$ and $\bm{e}_{res}$ and deal with them separately. 
Then, we have the following theorem for the probability guarantee of PEOs.

\begin{theorem}
  \label{theorem}
  (1) (\textbf{Probabilistic Guarantee}) Suppose that $m$ is sufficiently large. The PEOs test is $(\delta, 1-\epsilon)$-routing. 

  (2) (\textbf{False Positives}) Consider a neighbor $u$ whose distance to $\bm{q}$ is at least $\delta$. If $\cos \theta \le \tilde F^{-1}_{\tilde \theta}(\epsilon)$ ($\epsilon \le 0.5$), 
  where $\cos \tilde \theta = A_r(\bm{e})$ and $\tilde F_{\theta}$ is the CDF of distribution $\mathcal N^{\bm{e}, \cos \theta}_{\min}/\sqrt{2 L \ln m}$. Then the probability that $u$ passes PEOs test is at most $1-\tilde F_{\theta}(\tilde F^{-1}_{\tilde \theta}(\epsilon))$.

  (3) (\textbf{Variance of Estimation and Comparison to RCEOs}) Suppose that $H(\bm{e})/\sqrt{2L\ln m} \sim \mathcal N_H$, where $\mathcal N_H$ is an unknown normal distribution, and let $ \mathcal N^{\theta}_{\opt} = \mathcal N(\cos \theta, \sin^2 \theta/(2L\ln m) )$. When $w_{res} \le 1/(L+1)$,
  \begin{equation}
   - \frac{L+2}{L(L+1)^2\ln m} \le {\rm{Var}}[\mathcal N_H]  - {\rm{Var}}[\mathcal N^{\theta}_{\opt}] \le \frac{1}{(L+1)^2 \ln m} 
  \end{equation}
  where $\theta$ is the (unknown) angle between $\bm{e}$ and $\bm{q}$.
\end{theorem}

\textbf{Remarks.} (1) The first statement of Theorem~\ref{theorem} guarantees that promising neighbors are explored with a high confidence. 

(2) The second statement shows that the routing efficiency is determined by the variance of $\mathcal N^{\bm{e}, \cos \theta}_{\min}/\sqrt{2 L \ln m}$. Such variance is expected to be as small as possible since a smaller variance leads to a smaller probability of a false positive. 

(3) It is easy to see that, for RCEOs with $m^L$ projected vectors, the distribution associated with $\cos \theta$ is $\mathcal N^{\theta}_{\opt}$. For a comparison, we use $\mathcal N_H$ to denote the distribution of $H(\bm{e})/\sqrt{2L\ln m}$. Clearly, $\mathbb E[\mathcal N_H] = \mathbb E[\mathcal N^{\theta}_{\opt}]$. On the other hand, the third statement shows that, if $w_{res}$ is a small value, the variances of such two distributions are very close. In this situation, the effect of PEOs is close to that of RCEOs with $m^L$ projected vectors, which explains why PEOs can perform much better than RCEOs empirically. 

\subsection{Impact of $L$}
\label{sec:impact-of-L}
Based on the second and the third statements in Theorem~\ref{theorem}, $w_{reg}$ and $w_{res}$ are critical values which control the routing efficiency. First, we want to show that $w_{reg}$ is generally close to 1. To this end, we calculate $\mathbb E[w_{reg}]$ under the assumption that vector $\bm{e}$ obeys an isotropic distribution. Because prevalent graph indexes (e.g., HNSW) diversify the selected edges in the indexing phase, 
such assumption is not very strong for the real datasets. Besides, we can permute the dimensions (Sec.~\ref{sec:peos-implementation}) to make $\bm{e}$ follow an isotropic distribution. 
Let $\bar w_{reg}(L,d)$ denote $\mathbb E_{\bm{e} \sim U(\mathbb S^{d-1})}[w_{reg}]$ ($\bar w_{reg}(L,d)$ means that $w_{reg}$ is affected by $L$ and $d$). Then, we have the following lemma ($d' > 3$). 
\begin{lemma}
  \label{lemma:bounds-of-wreg}
  Given $L$, $d'$, and $d' = d/L$, 
  \begin{equation} 
    \bar w_{reg}(L, d)  \ge \frac{(d'-1)\sqrt{2Ld-3L}}{(d-1) \sqrt{2d'+2\sqrt{3}-6}}. 
  \end{equation}
\end{lemma}



As an example, $\bar w_{reg}(L, d) \ge 0.978$, when $d = 128$ and $L=8$. For other reasonable choice of $L$ w.r.t. $d$, we can also obtain a $\bar w_{reg}(L, d)$ close to 1. 
Next, we analyze the relationship between $L$ and $w_{res}$ more accurately. To this end, we consider the distribution $\mathcal N^{\bm{e}, \cos \theta}_{\min}/\sqrt{2 L \ln m}$. Its expected value is $\cos \theta$ and its variance, which is expected to be as small as possible, as shown in Theorem~\ref{theorem}, is of great interest to us. For its variance, we particularly focus on the remaining part $J_{rel}$ by removing the part regrading $\cos \theta$, which is a value close to 0 for most of $\bm{e}$'s. Specifically, $J_{rel}$ is defined as follows: 
\begin{equation}
  J_{rel}(L) = \frac{1 + (L-1) \mathbb E[w_{res}^2(L,d)]}{L}.
\end{equation}
On the other hand, for $\mathcal N^{\theta}_{\opt}$, we have $J_{\opt}(L) = 1/L$, which corresponds to RCEOs with $m^L$ projected vectors. Due to the effect of $w_{res}$, there is a difference between $J_{\opt}(L)$ and $J_{rel}(L)$. Thus, an appropriate value of $L$ should satisfies the following two requirements:

(1) (\textbf{Major}) $J_{rel}(L)$ should be as small as possible.

(2) (\textbf{Minor}) $w_{res}$ is close to $1/(L+1)$.

Here, the first requirement is to improve the routing efficiency of PEOs, which is more important, and the second one is to measure the deviation in the condition in the third statement (Theorem~\ref{theorem}). In Figure~\ref{fig:simulation-sift-var} -- \ref{fig:simulation-gist-var}, under the assumption of isotropic distribution, we plot the curves of $J_{rel}(L)$, $J_{\opt}(L)$ and $\Delta = |w_{res}-[1/(L+1)]|$, for $d = 128$, $384$ and $960$, respectively. It can be seen that (1) $w_{res}$ increases as $L$ grows, which means that $L$ should not be overlarge because $J_{rel}(L') > J_{rel}(L)$ may occur when $L' > L$, and (2) due to the closeness of $J_{\opt}$ and $J_{rel}$, the effect of PEOs is very close to that of RCEOs with $m^L$ projected vectors when $L$ is small (e.g. $L \le 8$). Based on the analysis above, we set $L$ to 8, 15, 16 for these three dimensions. By varying the value of $L$, the performance of PEOs on the real datasets with these dimensions is consistent with our analysis (Figures~\ref{fig:simulation-sift-L} -- \ref{fig:simulation-gist-L}), which will be elaborated in Sec.~\ref{exp:effect-of-L}. 


\begin{figure*}[!t]
  \centering
  \subfloat[GloVe200-angular, $K = 100$]{\includegraphics[width=.3\textwidth]{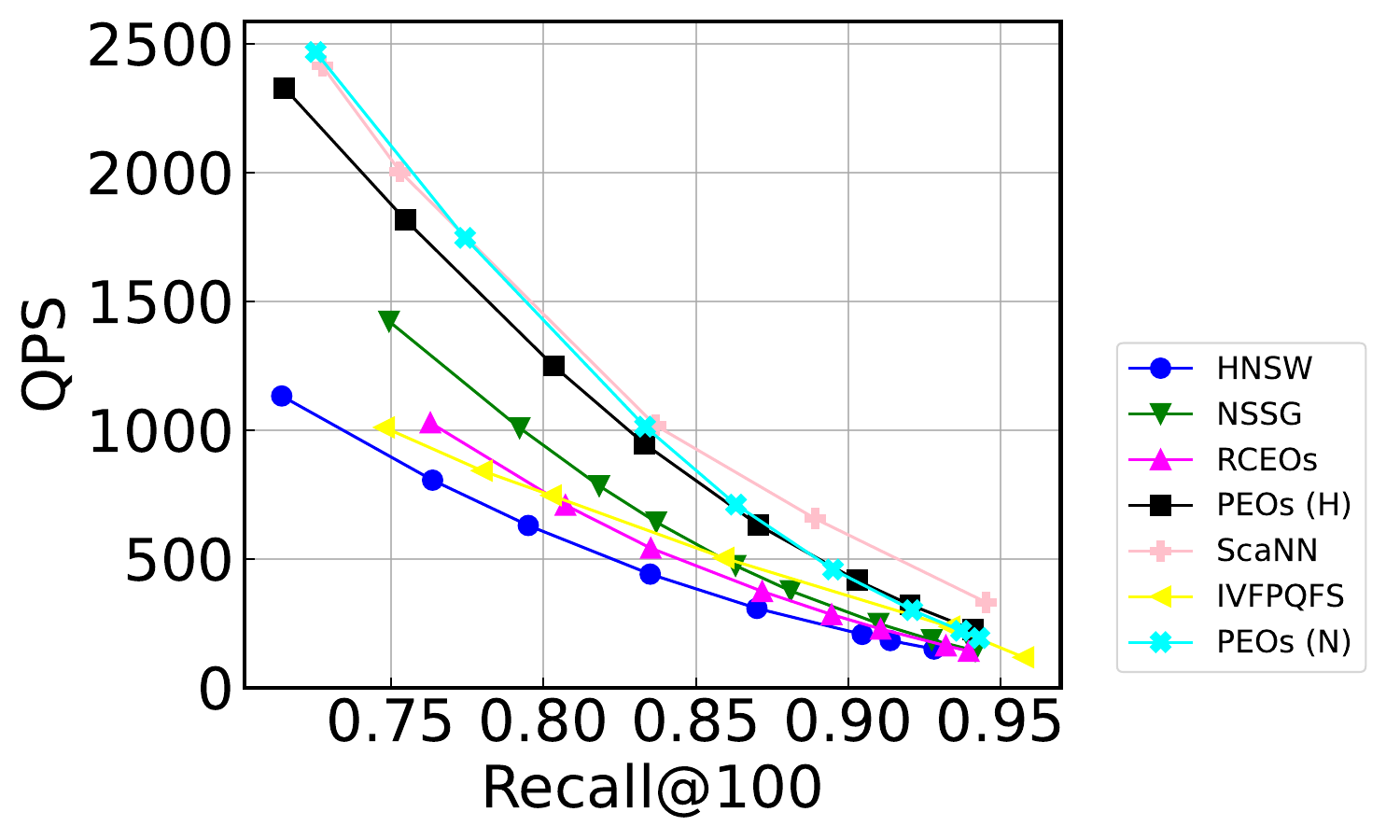}}
  \subfloat[GloVe300-$\ell_2$, $K = 100$]{\includegraphics[width=.3\textwidth]{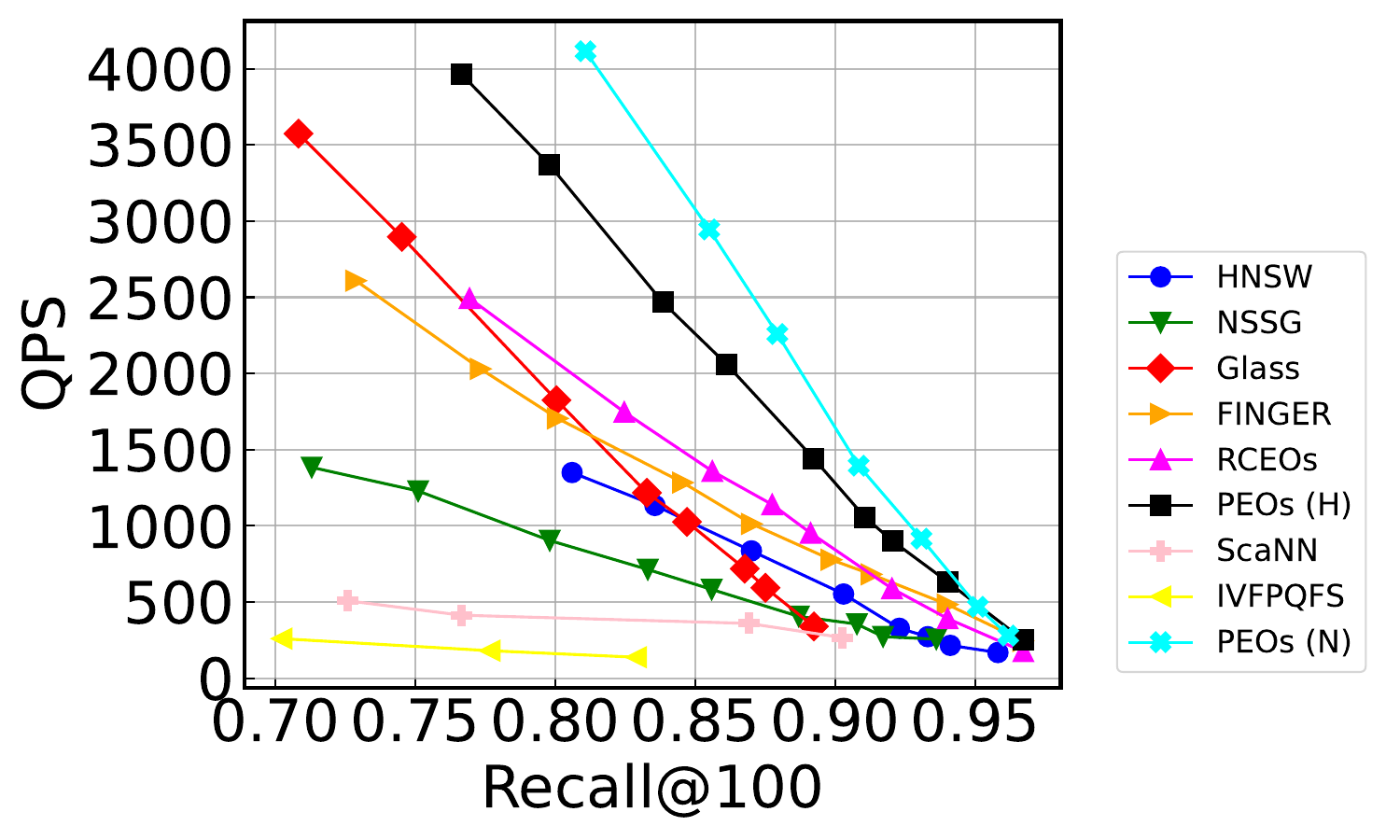}}
  \subfloat[DEEP10M-angular, $K = 100$]{\includegraphics[width=.3\textwidth]{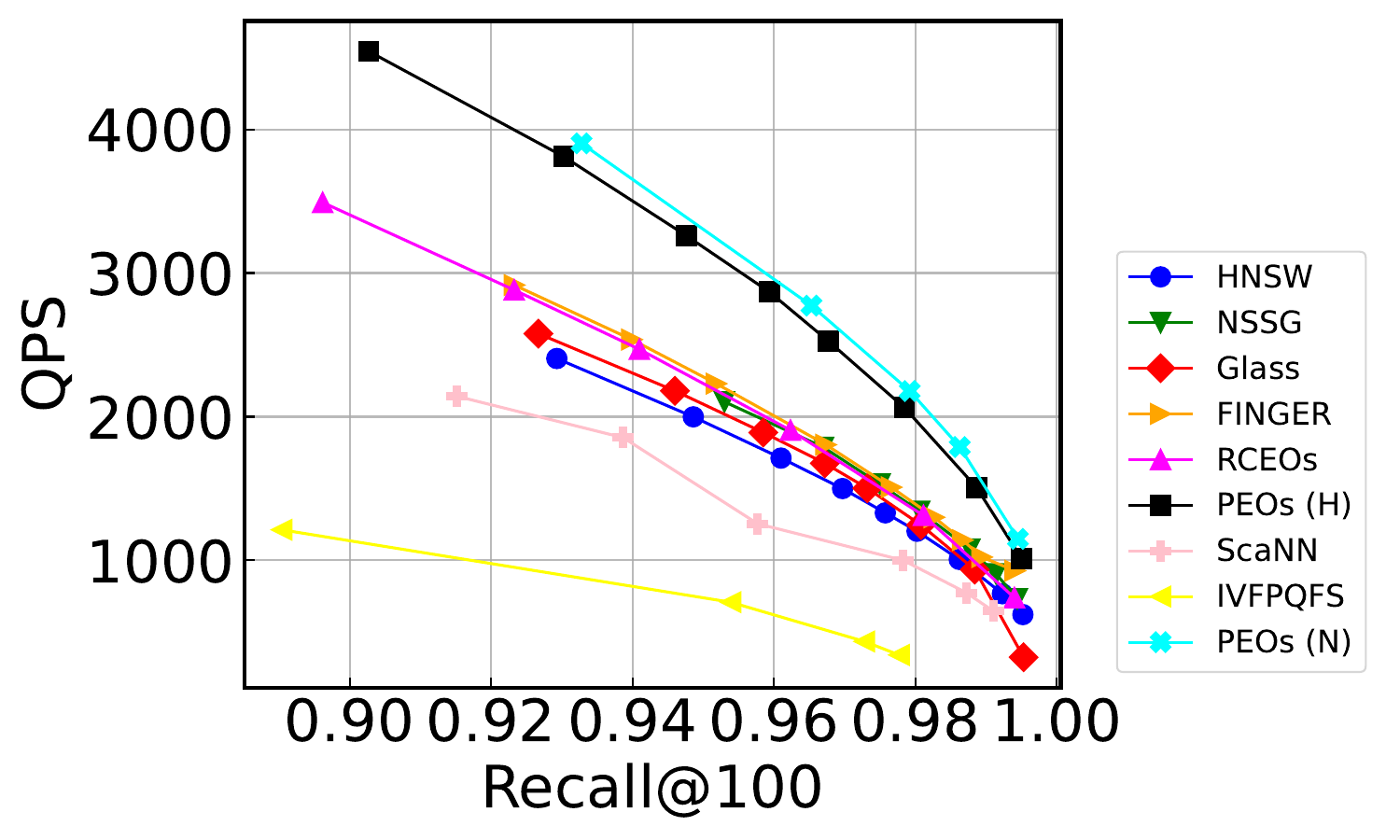}}
  
  \subfloat[SIFT10M-$\ell_2$, $K = 100$]{\includegraphics[width=.3\textwidth]{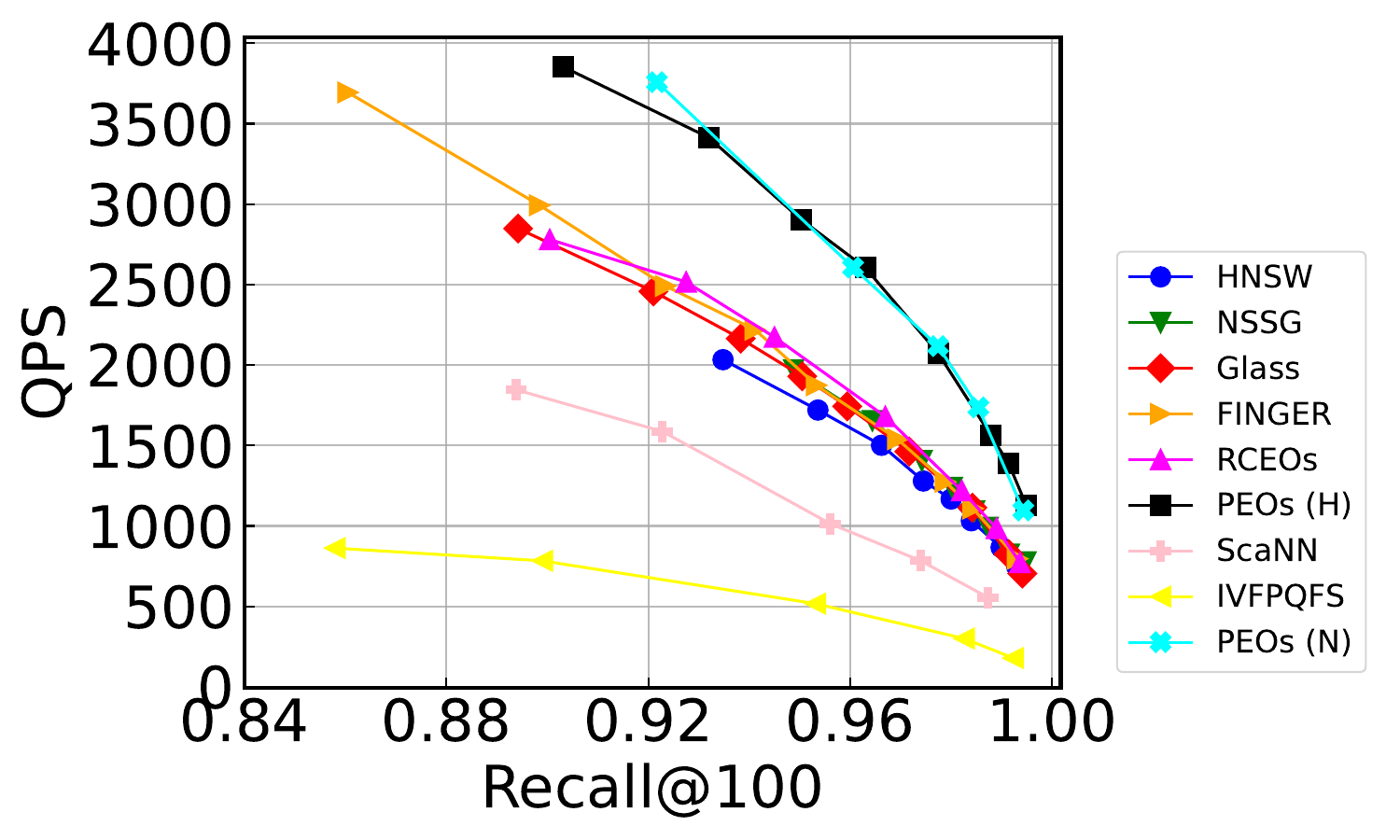}}
  \subfloat[Tiny5M-$\ell_2$, $K = 100$]{\includegraphics[width=.3\textwidth]{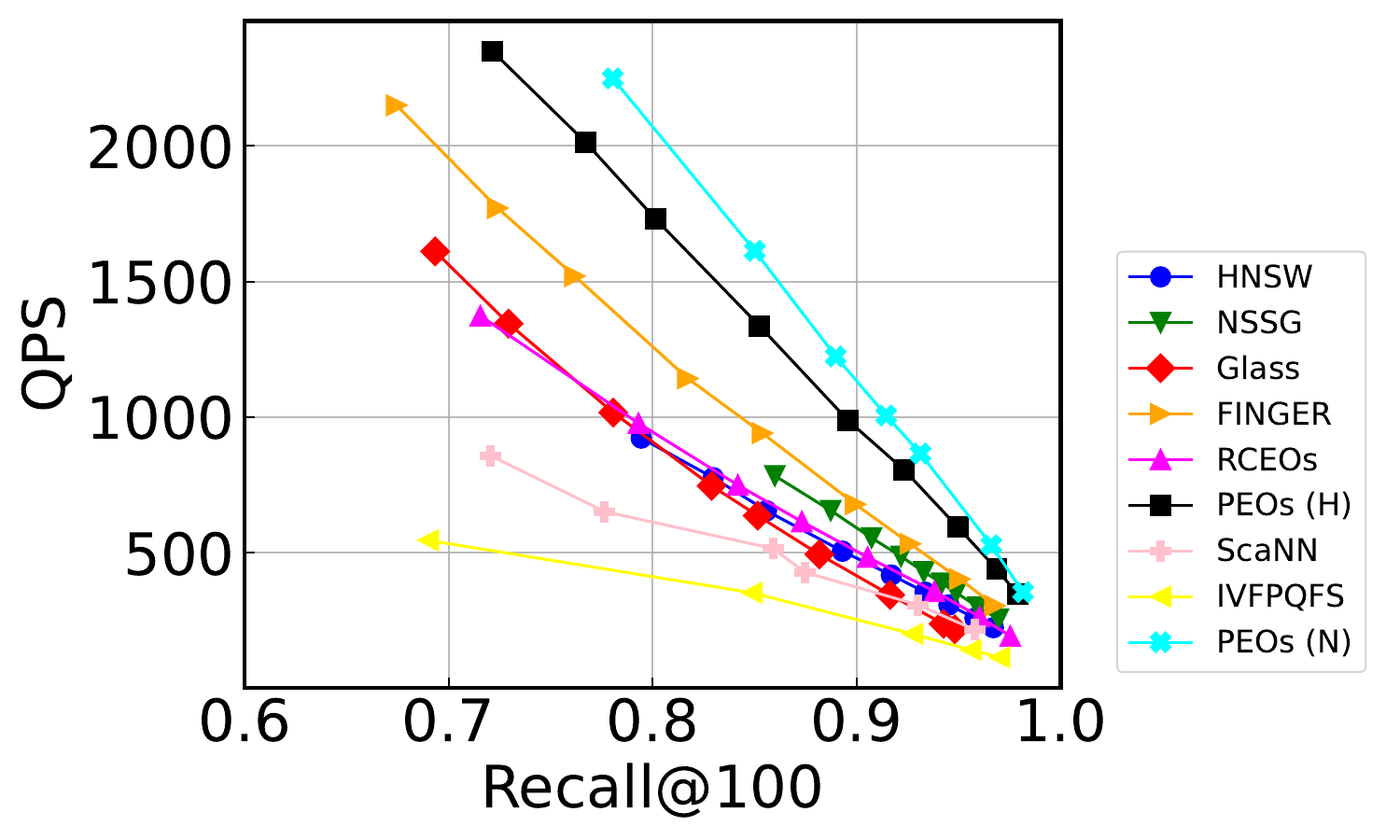}}
  \subfloat[GIST-$\ell_2$, $K = 100$]{\includegraphics[width=.3\textwidth]{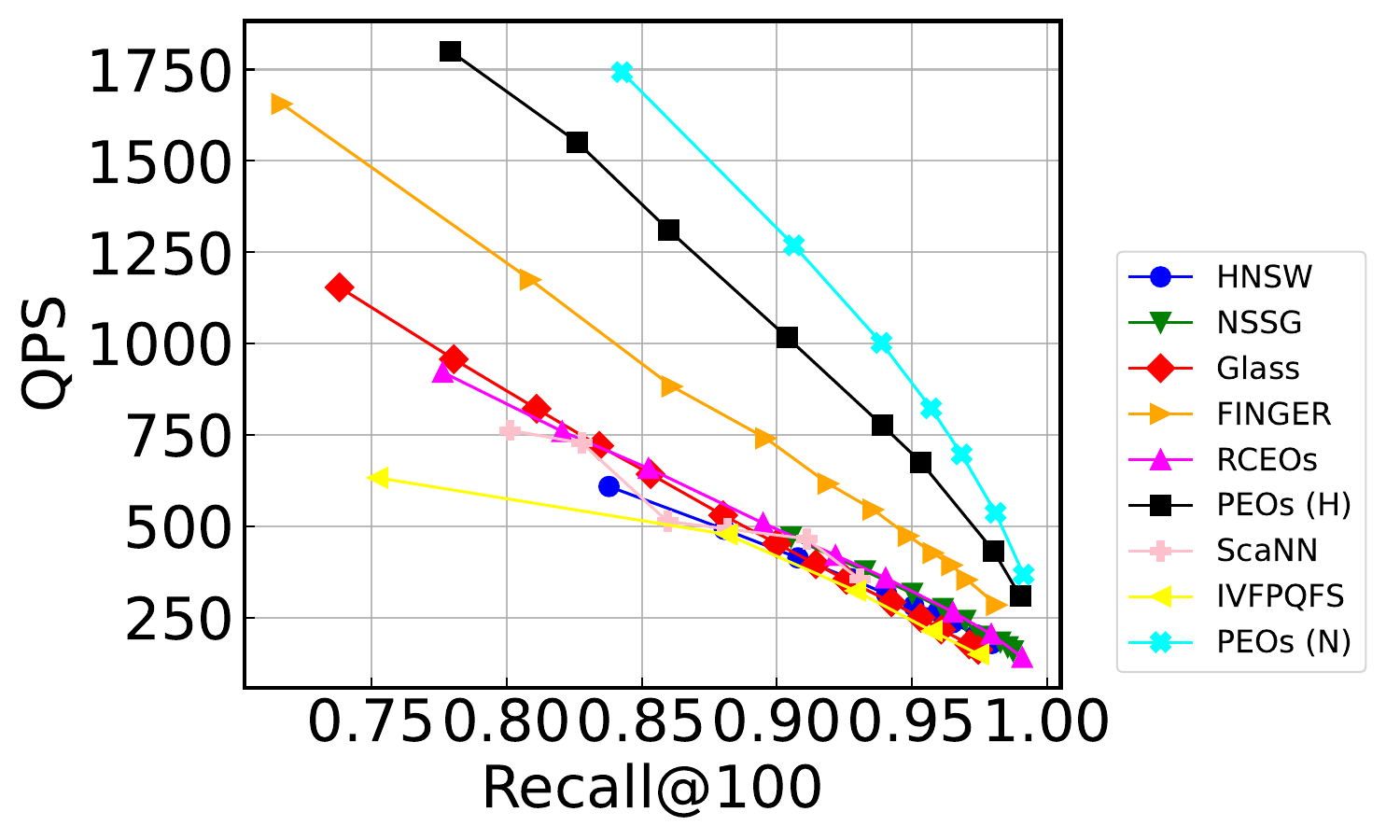}}
  \caption{Recall-QPS evaluation. PEOs (H) denotes HNSW+PEOs and PEOs (N) denotes NSSG+PEOs. The recalls of Glass and FINGER are lower than 30\% on GloVe200 and thus not shown.}
  \label{fig:performance-evaluation}
\end{figure*}

\subsection{Computational Cost and Space Cost}
\label{sec:discussion}

We assume that PEO is applied on HNSW. The analysis of other graph indexes (e.g., NSSG) is similar.

For the time complexity, it is difficult to provide a strict analysis since currently no graph-based method can offer such a guarantee. Nonetheless, if we assume that the length of the search path is $\omega$, where $\omega$ can be roughly estimated as $O(\log n)$ in practice, where $n$ is the number of data points, then the complexity of HNSW is $O(vd\omega)$, where $v$ is the average out-degree and $d$ is dimension. Thus, the complexity of HNSW+PEOs is $O(v\omega(L+4) + \rho vd\omega )$ (without the consideration of SIMD), where $4$ is the number of scalar operations and $\rho$ denotes the filtering ratio, which depends on the data distribution and is around 0.25 in practice (Appendix~\ref{sec:distance-reall-curves}).

As for the space cost of PEOs, take $M=32$ as an example (the maximum out-degree is 64). We use $a_{16}$, $a_{32}$, $a_{48}$, $a_{64}$ to denote the number of nodes whose out-degrees lie in $[1,16]$, $[17,32]$, $[33,48]$, and $[49,64]$, respectively (for the utilization of SIMD). Let $X=16 \times a_{16} + 32 \times a_{32} + 48 \times a_{48} + 64 \times a_{64}$. Let $n$ denote the number of data points and $d$ denote the dimension. Then, the space complexity is $O(X(L+4) + 4nd)$, assuming that we use 8 bits to compress every scalar. If we use 16 bits, the complexity is $O(X(L+8) + 4nd)$. $X$ is dependent on the edge-selection strategy and data distribution. For most cases, $X$ is around $32n$. Such space cost is affordable for 10M-scale and smaller datasets on a single PC (Sec.~\ref{sec:index-size-and-time}). For 100M-scale and larger datasets, we can slightly sacrifice efficiency to significantly reduce space consumption (Sec.~\ref{sec:100M}). In addition to the index, space consumption also includes the projection table and quantile lookup table. Both are are very small and can be stored in CPU cache -- one is 100 or 1000 floats and the other is $256L$ floats, where $L$ is typically in $[8, 32]$.





%% file: exp.tex
\begin{figure*}[!t]
  \centering
  \subfloat[Dimension: 128 (SIFT10M)]{\includegraphics[width=.3\textwidth]{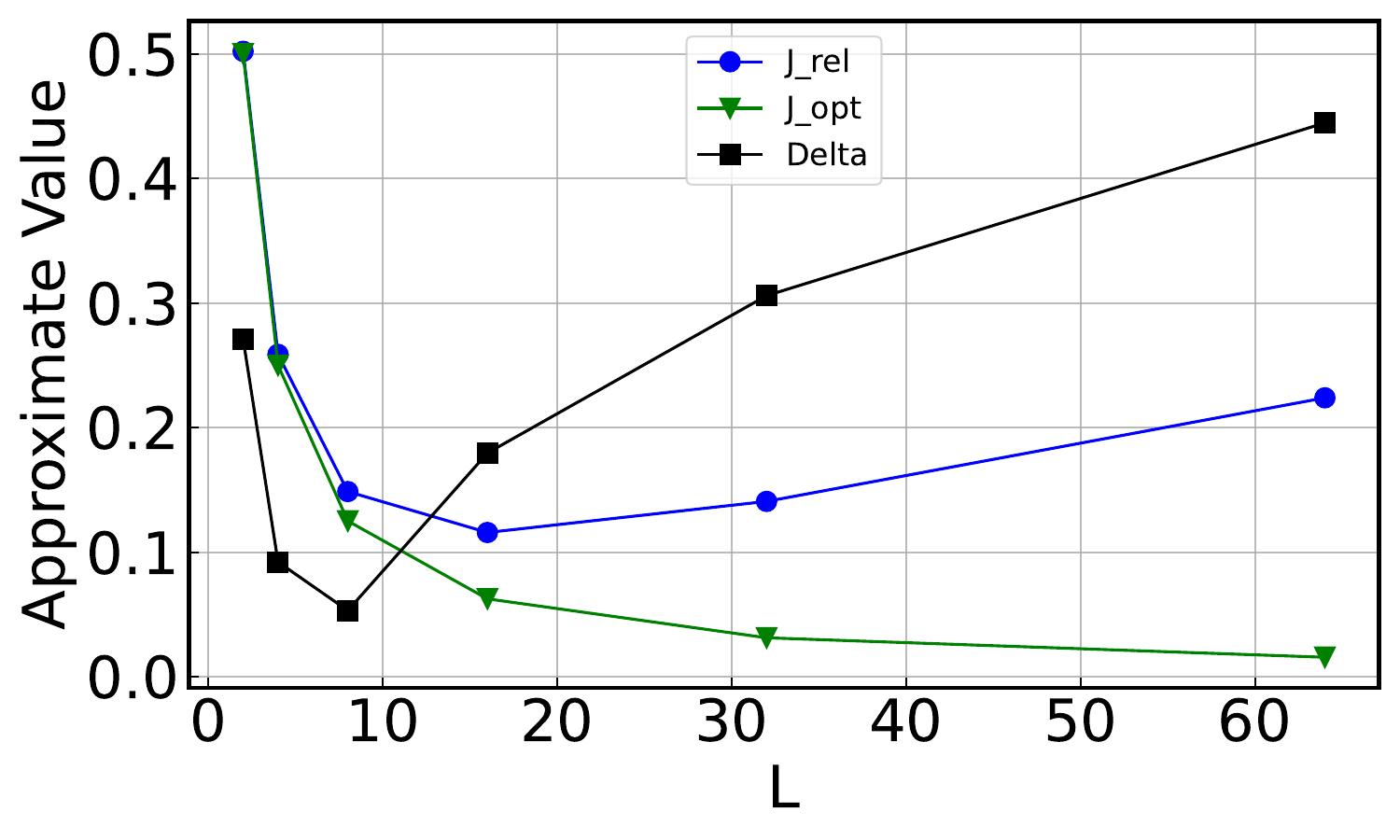}\label{fig:simulation-sift-var}}
  \subfloat[Dimension: 384 (Tiny5M)]{\includegraphics[width=.3\textwidth]{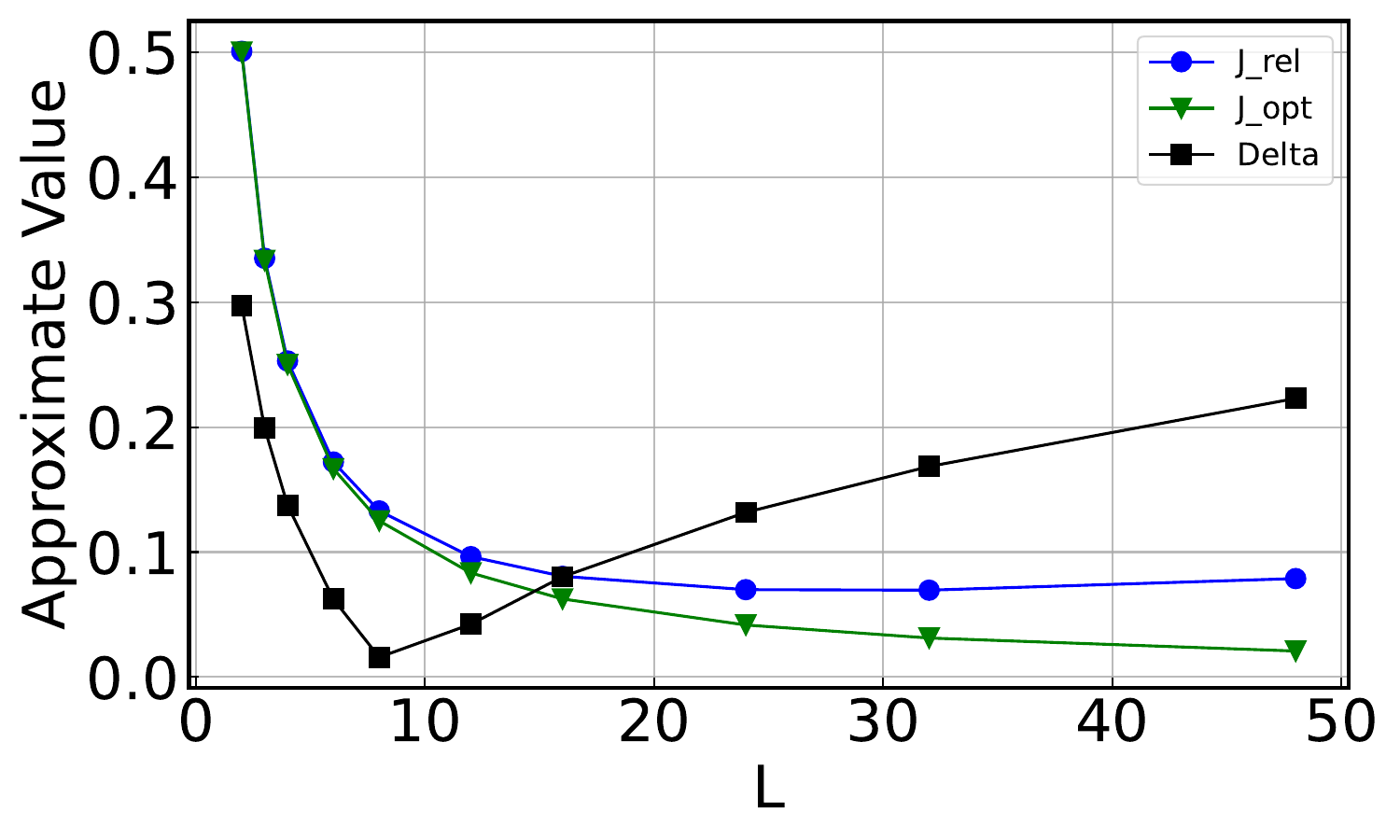}\label{fig:simulation-tiny-var}}
  \subfloat[Dimension: 960 (GIST)]{\includegraphics[width=.3\textwidth]{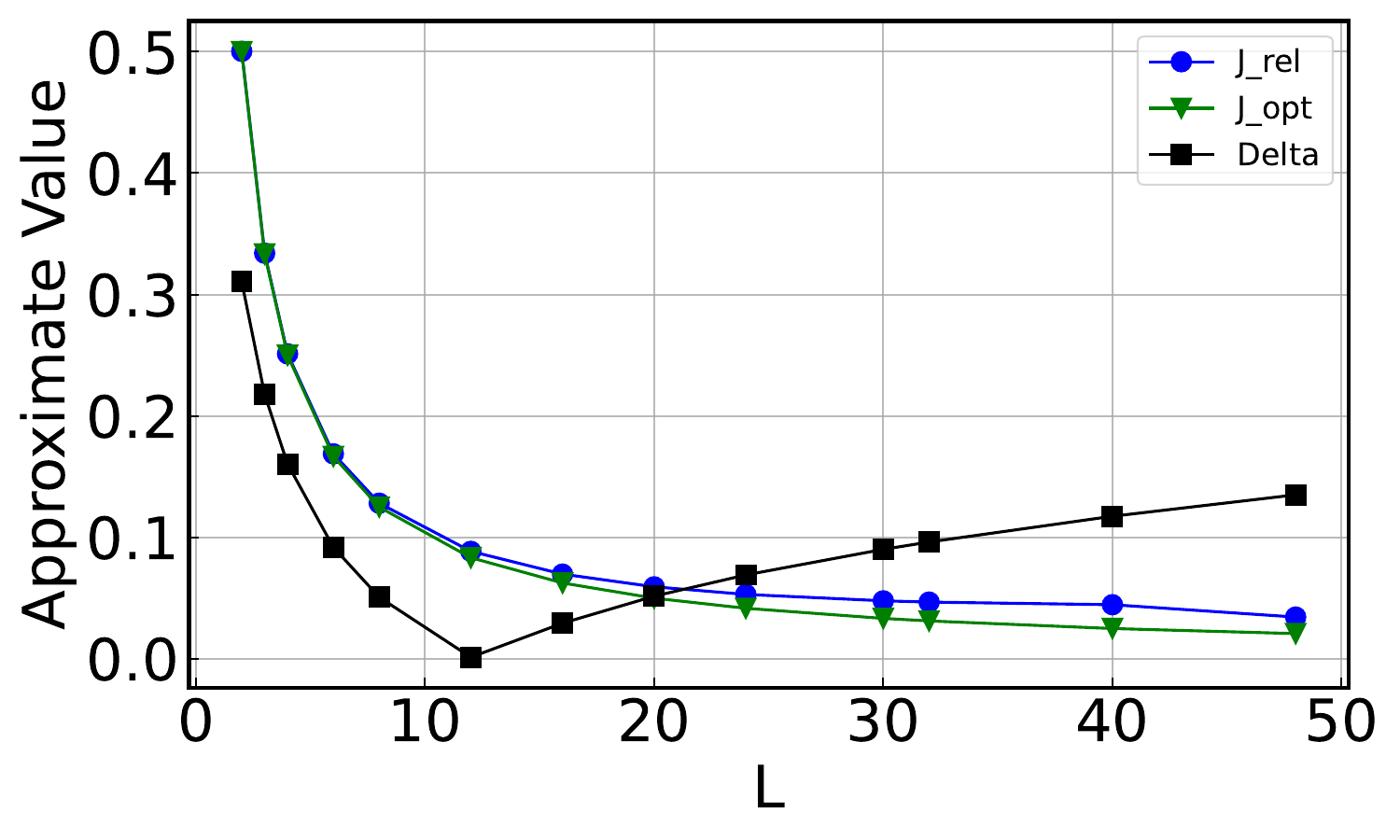}\label{fig:simulation-gist-var}} 

  \subfloat[SIFT10M-$\ell_2$, $K=100$]{\includegraphics[width=.3\textwidth]{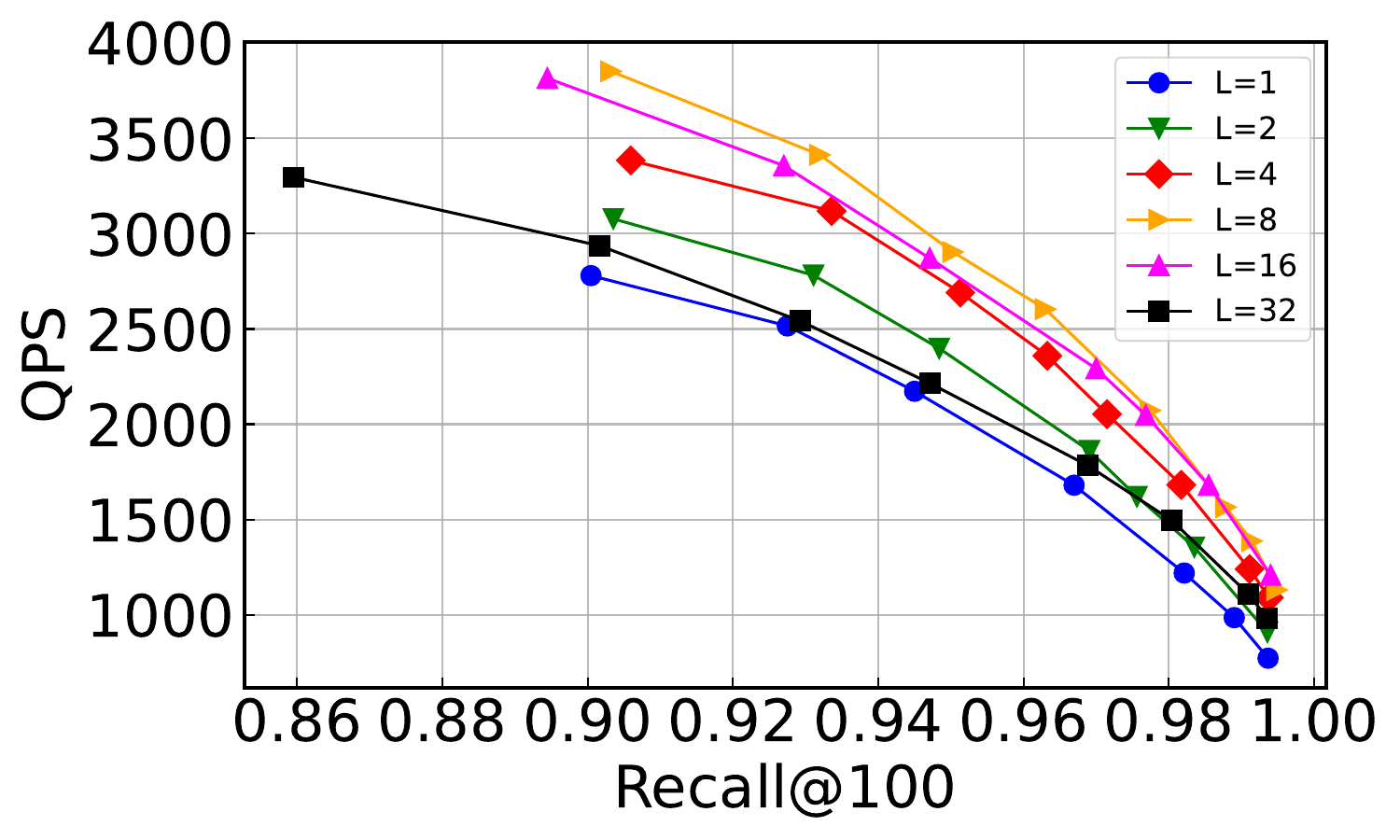}\label{fig:simulation-sift-L}}
  \subfloat[Tiny5M-$\ell_2$, $K=100$]{\includegraphics[width=.3\textwidth]{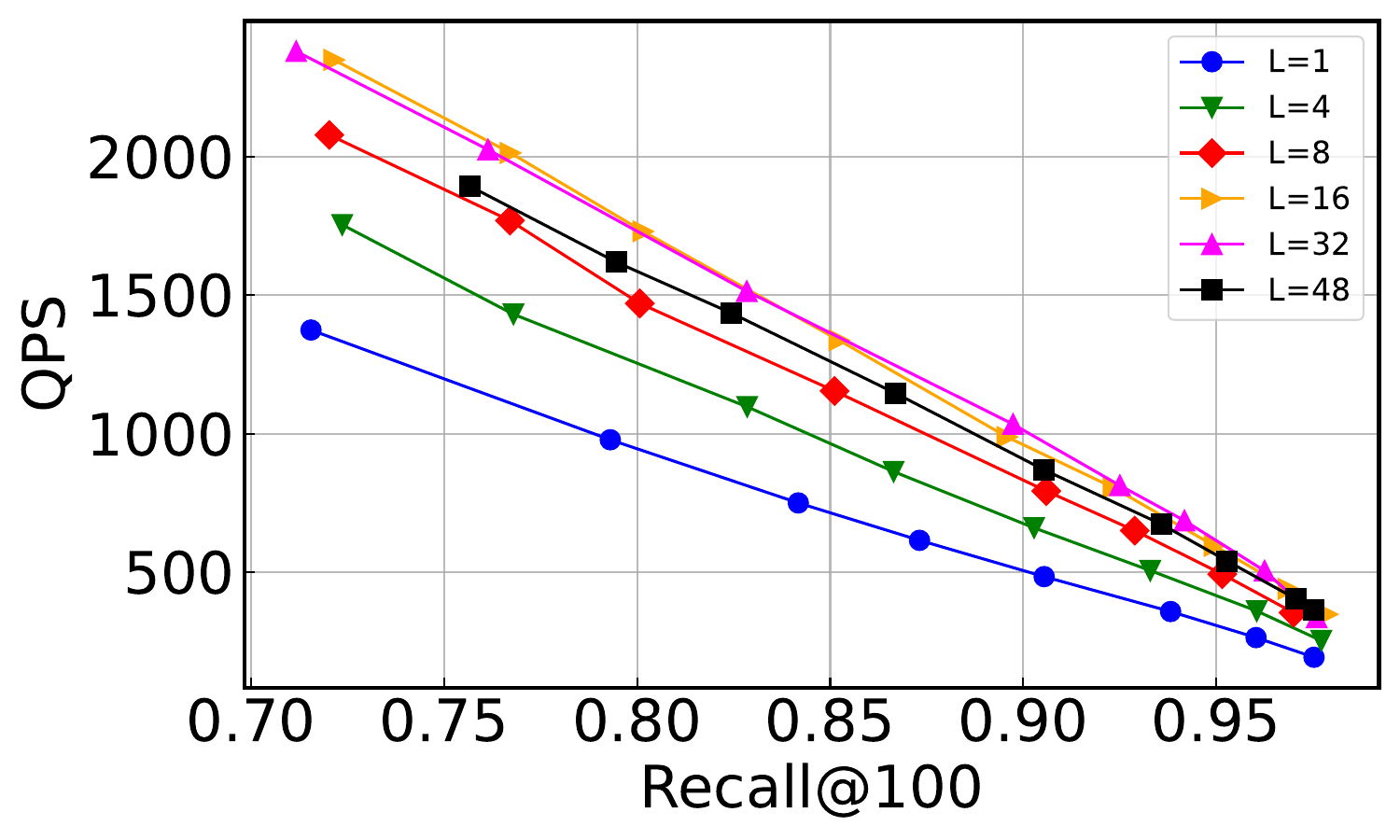}\label{fig:simulation-tiny-L}}
  \subfloat[GIST-$\ell_2$, $K=100$]{\includegraphics[width=.3\textwidth]{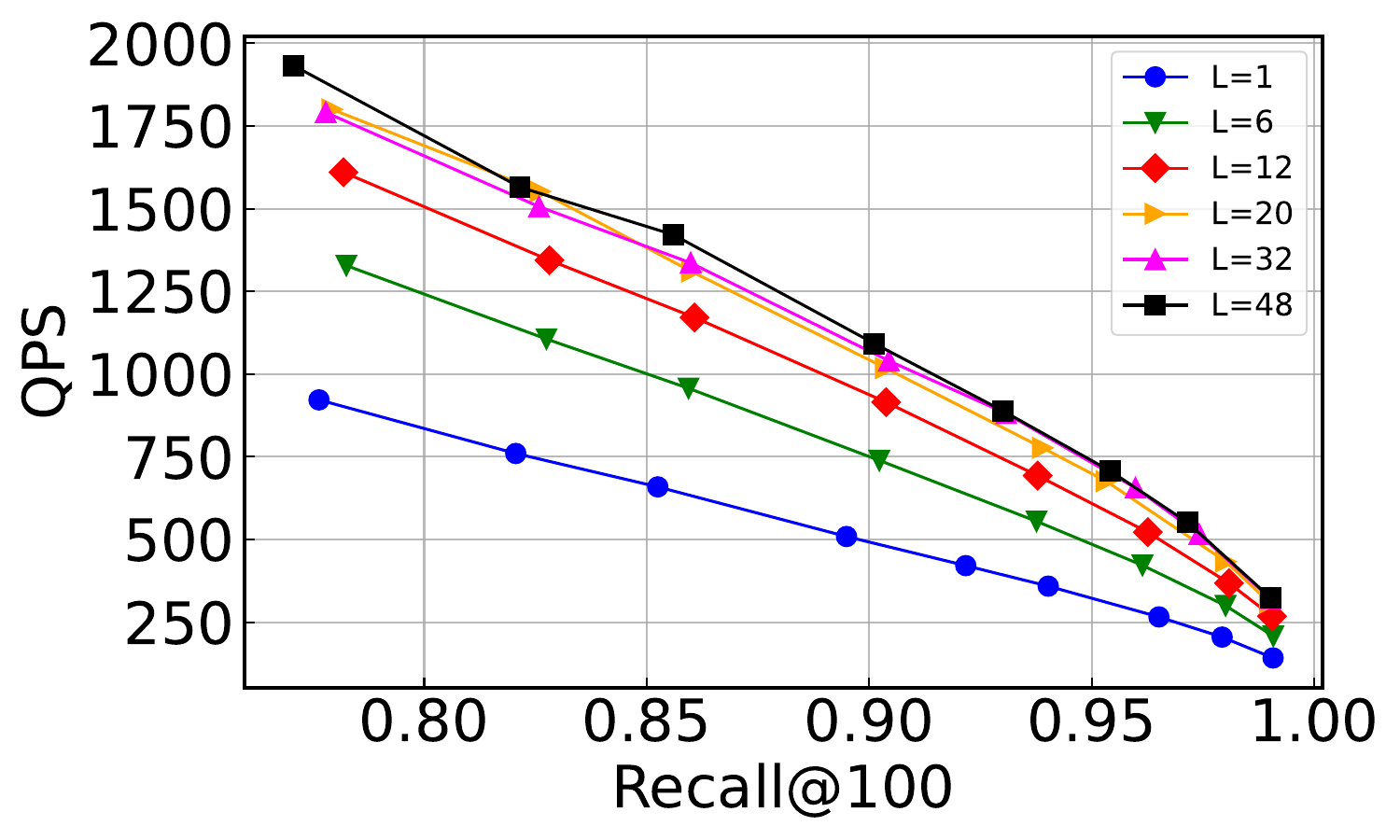}\label{fig:simulation-gist-L}}

  \caption{Effect of $L$. We plot the approximate values of $J_{opt}$, $J_{rel}$, and $\Delta$ under the isotropic distribution and the empirical performance.}
  \label{fig:simulation}
\end{figure*}

\begin{figure*}[!t]
  \centering
  \subfloat[DEEP10M-angular, $K=100$]{\includegraphics[width=.3\textwidth]{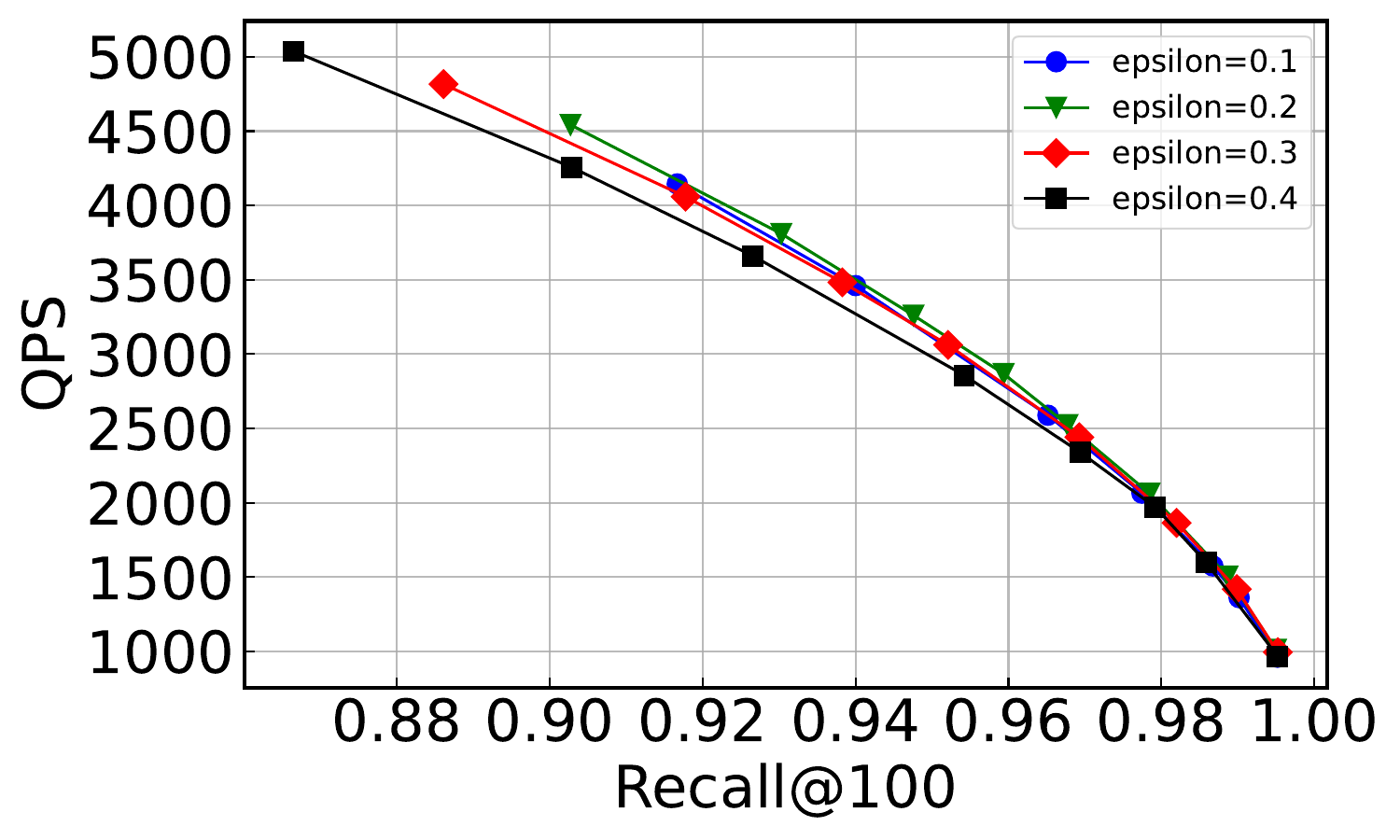}}
  \subfloat[GloVe200-angular, $K=100$]{\includegraphics[width=.3\textwidth]{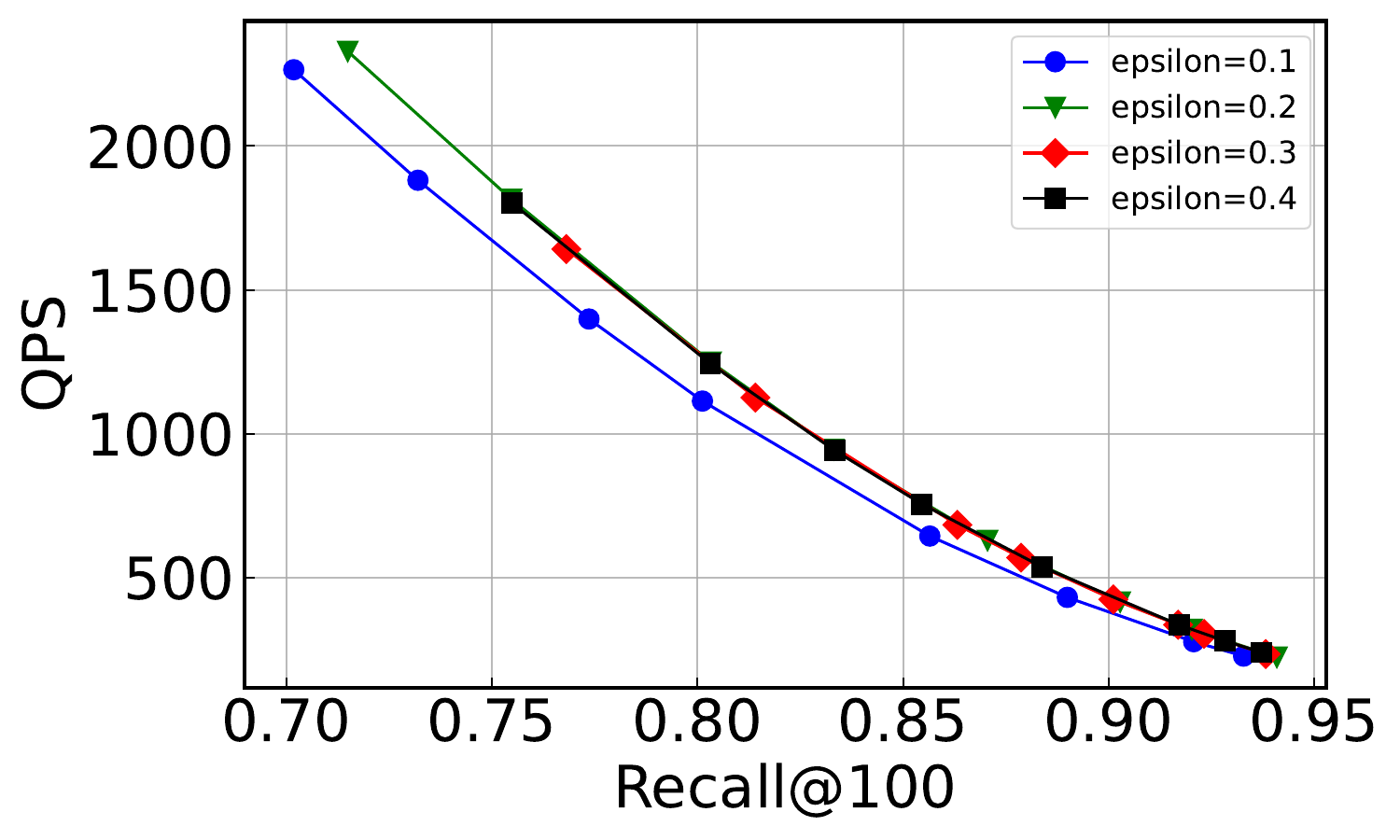}}
  \subfloat[GloVe300-$\ell_2$, $K=100$]{\includegraphics[width=.3\textwidth]{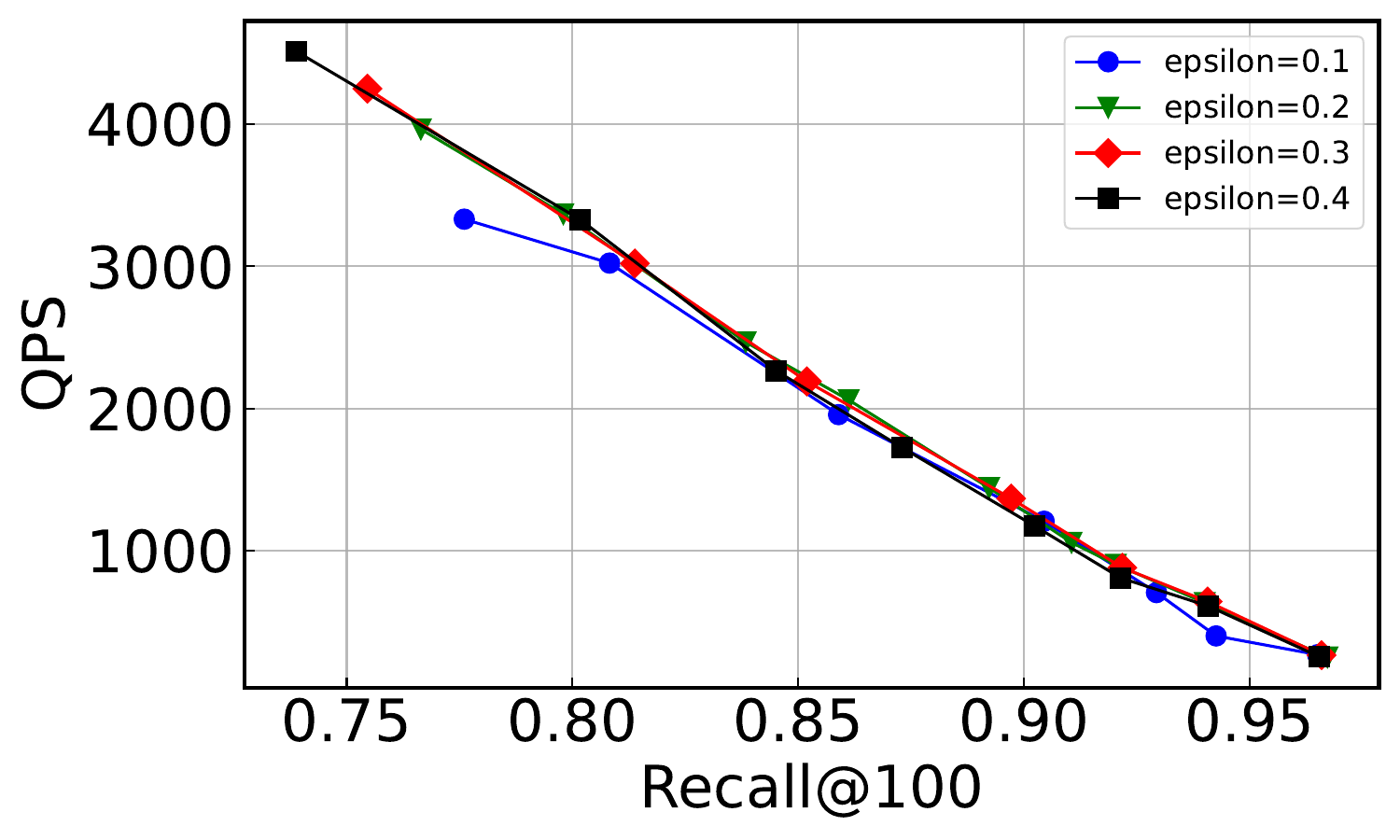}}
  \caption{Effect of $\epsilon$ on DEEP10M, GloVe200 and GloVe300.}
 \label{fig:different-eps}
\end{figure*}

\section{Experiments}
\label{sec:exp}

\subsection{Datasets and Methods}
We use six million-scale datasets and a 100M dataset for scalability test. The statistics can be found in Table~\ref{tab:datasets}. We implement PEOs on two widely used SOTA graph index: HNSW and NSSG. 
In the experiments, PEOs refers to HNSW+PEOs unless stated otherwise. We select seven competitors: (vanilla) HNSW, (vanilla) NSSG~\cite{nssg}, Glass~\cite{glass}, ScaNN~\cite{scann}, IVFPQFS (in Faiss)~\cite{douze2024faiss}, FINGER~\cite{FINGER}, and RCEOs. 
The default value of $K$ is 100.

\begin{table}[!t]
  \caption{Dataset statistics.}
  \label{tab:datasets}
  \vskip 0.05in
  \small
  \centering
  \begin{tabular}{lccccr}
    \toprule
    Dataset    & Size ($\size{\mathcal{O}}$) & Dim. ($d$) & Type & Metric   \\
    \midrule
    GloVe200   & 1,183,514 & 200 & Text  & angular  \\
    GloVe300   & 2,196,017 & 300 & Text  & $\ell_2$ \\    
    DEEP10M    & 9,990,000 & 96  & Image & angular  \\
    SIFT10M    & 10,000,000 & 128 & Image & $\ell_2$ \\
    Tiny5M     & 5,000,000 & 384 & Image & $\ell_2$ \\
    GIST       & 1,000,000 & 960 & Image & $\ell_2$ \\
    DEEP100M   & 100,000,000 & 96 & Image & angular \\
    \bottomrule
  \end{tabular}
\end{table}

\begin{table*}[t]
  \caption{Index size and indexing time.}
  \label{tab:indexing-time}
  \vskip 0.05in
  \small
  \centering
  \begin{tabular}{l|ccc|ccc}
    \hline
    \multicolumn{1}{c}{\multirow{2}{*}{Dataset}} & \multicolumn{3}{|c}{\multirow{1}{*}{Index Size (GB)}} & \multicolumn{3}{|c}{\multirow{1}{*}{Indexing Time (s)}}\\ \cline{2-7}
    & \multicolumn{1}{|c}{\multirow{1}{*}{HNSW}}
    & \multicolumn{1}{c}{\multirow{1}{*}{HNSW+FINGER}}
    & \multicolumn{1}{c}{\multirow{1}{*}{HNSW+PEOs}}
    & \multicolumn{1}{|c}{\multirow{1}{*}{HNSW}}
    & \multicolumn{1}{c}{\multirow{1}{*}{HNSW+FINGER}}
    & \multicolumn{1}{c}{\multirow{1}{*}{HNSW+PEOs}}\\
    \hline
    GloVe200 & 1.19 & 3.89 (+2.27x) & 2.27 (+0.91x) & 737 & 463+38 & 794+33\\
    GloVe300 & 3.02 & 8.04 (+1.66x) & 3.70 (+0.23x) & 1310 & 1408+178 & 1346+24\\    
    DEEP10M & 6.14 & 31.15 (+4.07x) & 12.13 (+0.98x) & 1245 & 1103+849 & 1296+208\\
    SIFT10M & 7.34 & 31.44 (+3.28x) & 14.39 (+0.96x) & 1490 & 1308+1025 & 1536+204\\
    Tiny5M & 8.44 & 20.49 (+1.43x) & 12.89 (+0.53x) & 2738 & 2959+1220 & 2880+158\\
    GIST & 3.83 & 6.12 (+0.60x) & 4.64 (+0.21x) & 738 & 790+706 & 793+40\\
    \hline
  \end{tabular}
\end{table*}

\begin{table}[t]
  \caption{Effect of compact implementation on index size (GB).}
  \label{tab:compact}
  \small
  \centering
  \begin{tabular}{lccc}
    \toprule
    Dataset    & HNSW & PEOs & PEOs (Compact)  \\
    \midrule   
    DEEP10M    & 6.14 & 12.13 (+0.98x) & 7.89 (+0.29x)  \\
    SIFT10M    & 7.34 & 14.39 (+0.96x) & 9.62 (+0.26x) \\
    \bottomrule
  \end{tabular}
\end{table}

\begin{figure}[t]
  \centering
  \subfloat[\scriptsize{DEEP10M-angular, $K=10$}]{\includegraphics[width=.48\linewidth]{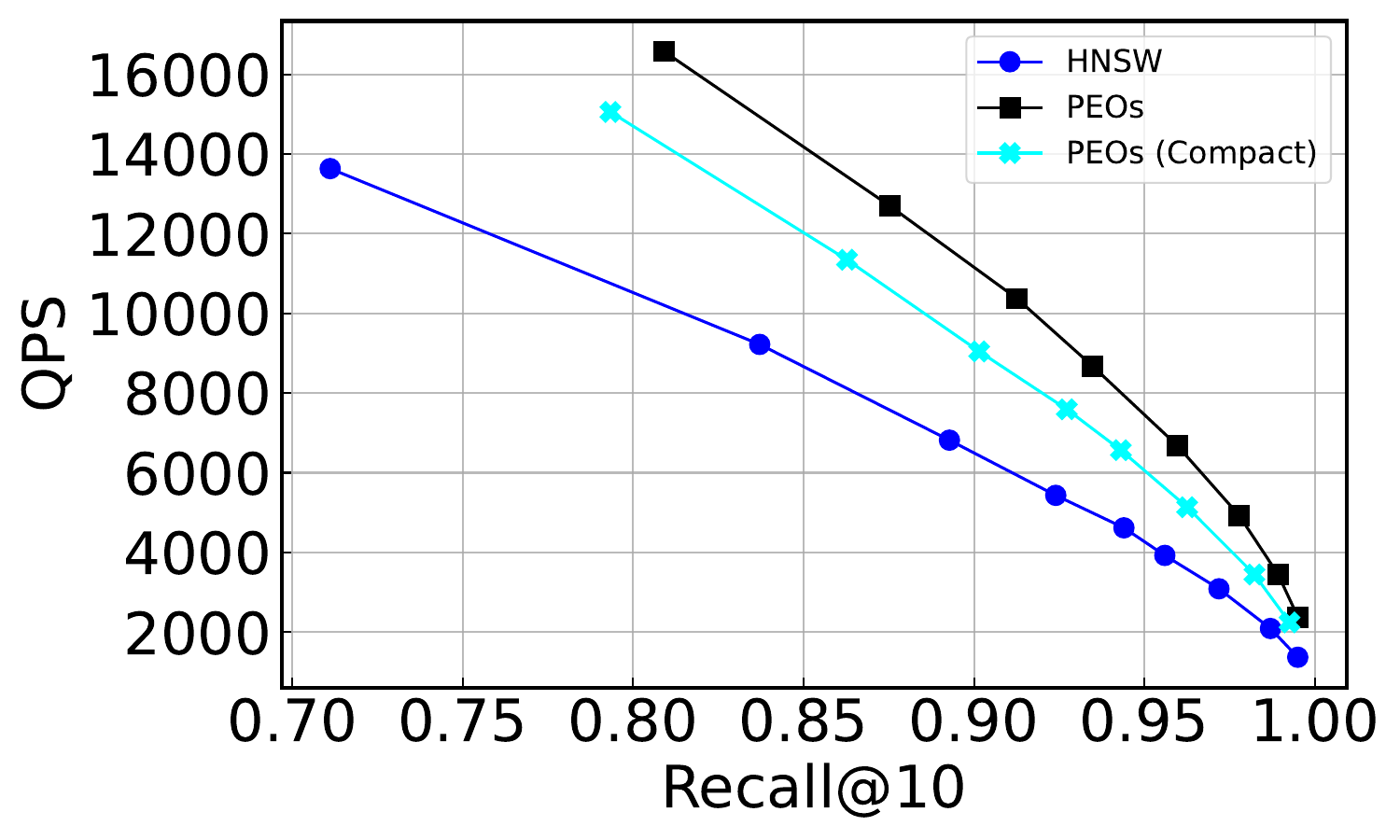}}
  \hspace{1ex}
  \subfloat[\scriptsize{SIFT10M-$\ell_2$, $K=10$}]{\includegraphics[width=.48\linewidth]{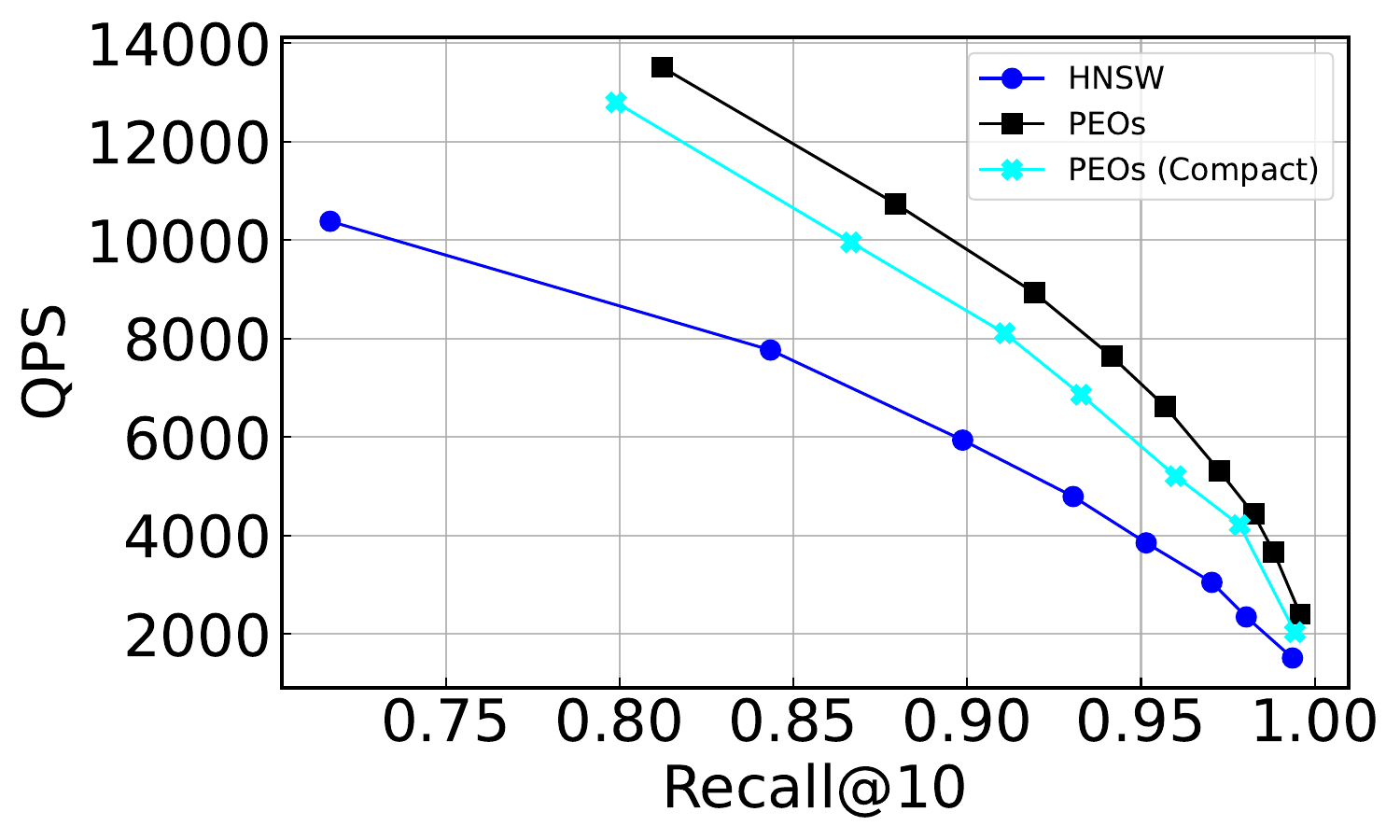}}
  \caption{Effect of compact implementation on search speed.}
  \label{fig:compact}
\end{figure}

\begin{figure*}[t]
  \centering
  \subfloat[DEEP100M-angular, $K=1$]{\includegraphics[width=.3\textwidth]{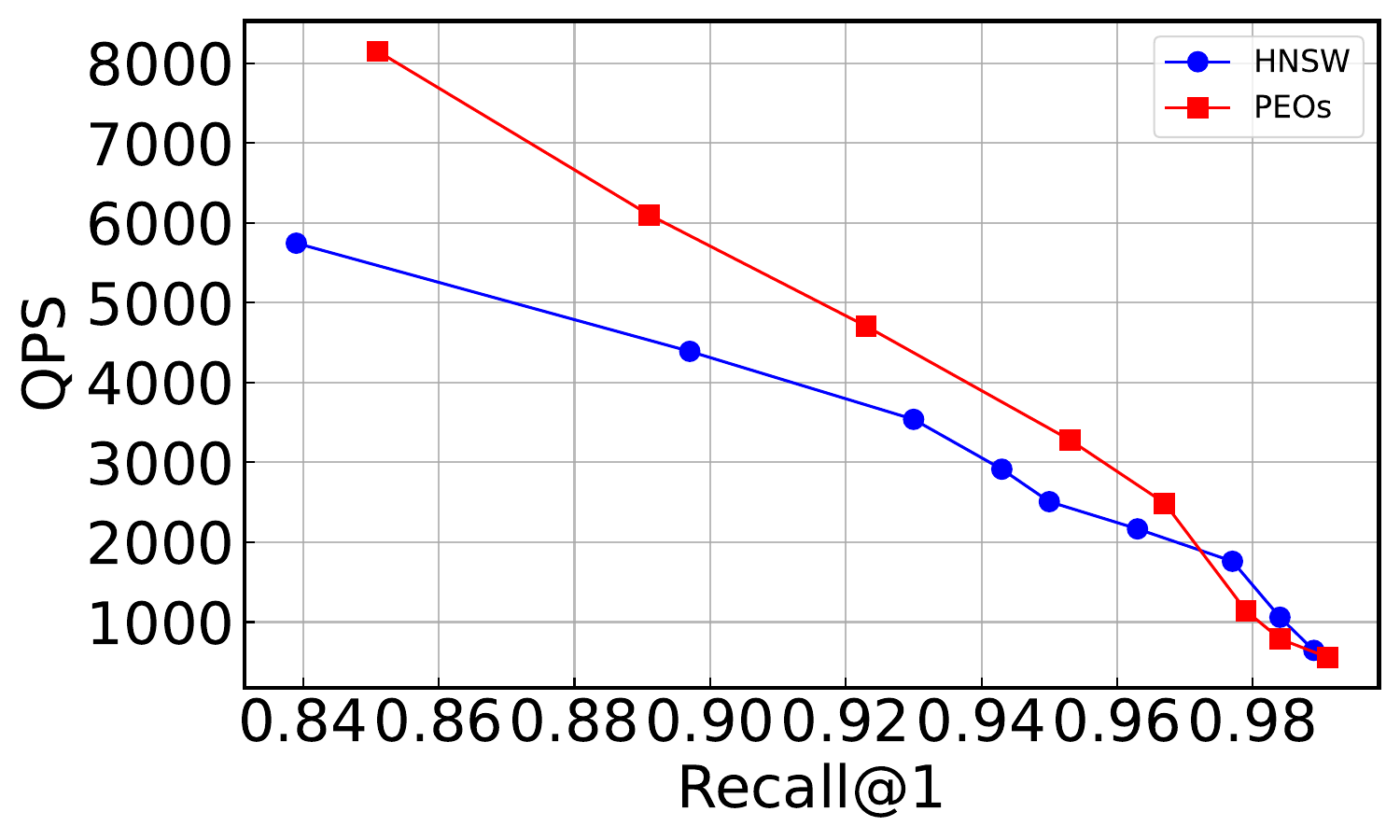}}
  \subfloat[DEEP100M-angular, $K=10$]{\includegraphics[width=.3\textwidth]{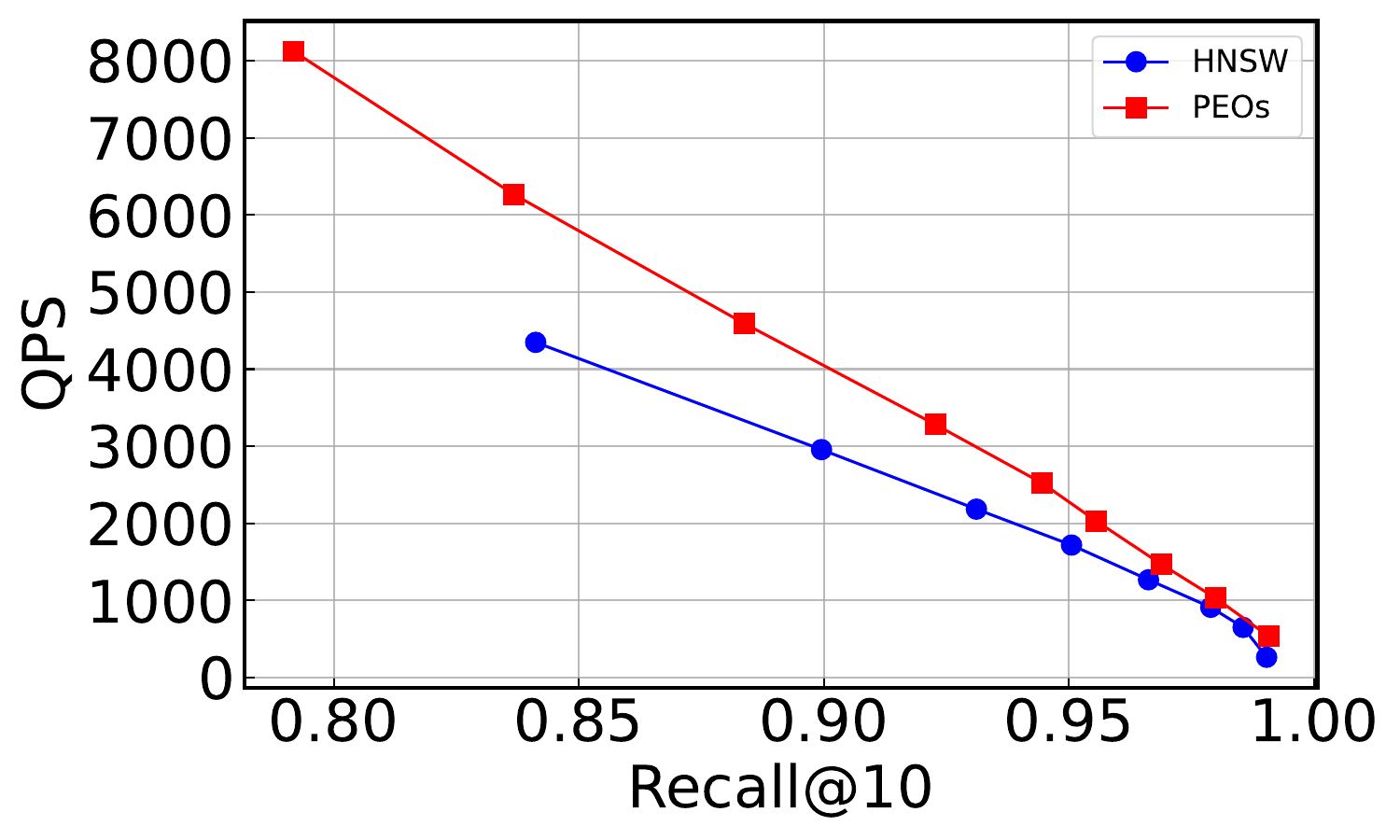}}
  \subfloat[DEEP100M-angular, $K=100$]{\includegraphics[width=.3\textwidth]{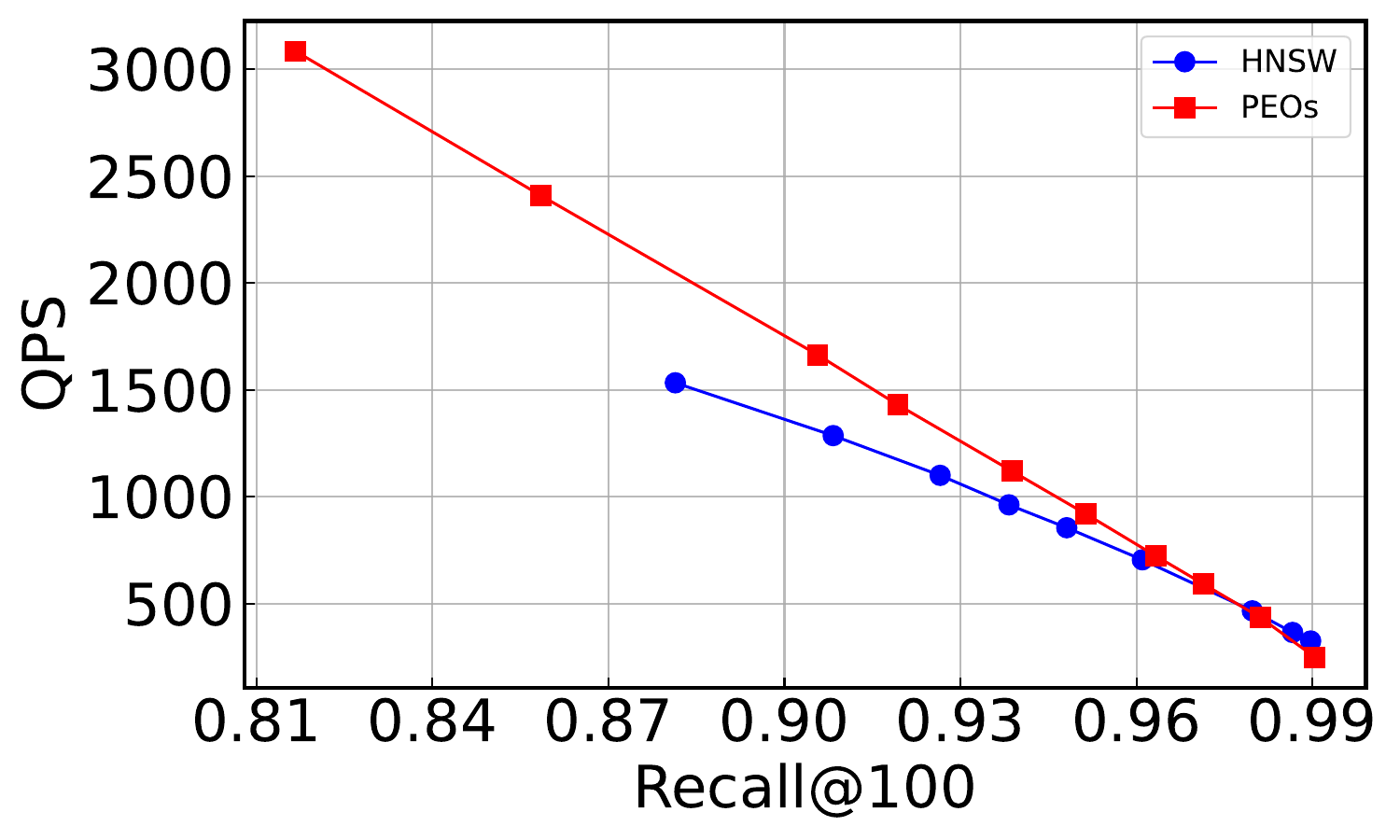}}
  \caption{Recall-QPS evaluation on DEEP100M. $L = 2$ for PEOs.}
  \label{fig:deep100M}
\end{figure*}

\subsection{Queries Per Second (QPS) Evaluation}
Figure~\ref{fig:performance-evaluation} reports the recall-QPS curves of all the competitors on the million-scale datasets. We have four observations. 

(1) On all the datasets, the winner is PEOs, trailed by FINGER in most cases. This demonstrate that our routing is effective. In particular, PEOs accelerates HNSW by 1.6 to 2.5 times and it is faster than FINGER by 1.1 to 1.4 times. 
(2) On GloVe300, Tiny5M and GIST, NSSG+PEOs obviously outperforms HNSW+PEOs, despite a smaller performance gap of the two vanilla versions. This is because, compared with HNSW, NSSG achieves a higher search accuracy at the cost of longer time for routing, making the improvement of PEOs over NSSG more significant.
(3) The improvement of PEOs is more significant on the datasets with more dimensions, because calculating the exact distance to the query is more costly on these datasets. 
(4) On GloVe200, FINGER and Glass report very low recall ($<$ 30\%) recall while the improvement of PEOs is still obvious under high recall settings. Specifically, FINGER is also based on routing, but incurs unbounded estimation errors and the errors might be very large on GloVe200, resulting in many false negatives of and rendering the graph under-explored. This result evidences the importance of the probability guarantee of routing. 


\subsection{Effect of Space Partition Size $L$}
\label{exp:effect-of-L}
We evaluate the effect of $L$ in HNSW+PEOs and report the results in Figure~\ref{fig:simulation} on the SIFT10M, Tiny5M, and GIST datasets (other datasets are evaluated in Appendix~\ref{sec:effect-of-L-others}). Figures~\ref{fig:simulation-sift-var} -- \ref{fig:simulation-gist-var} show theoretical results, and Figures~\ref{fig:simulation-sift-L} -- \ref{fig:simulation-gist-L} are their empirical counterparts. The empirical results are consistent with our analysis in Sec.~\ref{sec:impact-of-L}. That is, the smaller $J_{rel}$ is, the better the performance is. We also have the following observations. (1) The performance under $L > 1$ is obviously better than that under $L = 1$, showcasing the effectiveness of space partitioning. (2) An $L$ of 32 leads to a worse performance than an $L$ of 8 on the SIFT10M dataset, because the variance is larger when $L = 32$. (3) When $L > 16$ on the GIST dataset, the performance tends to be stable since the variance barely changes when $L$ exceeds 16.


\subsection{Effect of Error Bound $\epsilon$}
We vary the value of $\epsilon$ and report the results in Figure~\ref{fig:different-eps}. The performance of PEOs under $\epsilon = 0.1$ is slightly worse than those under other $\epsilon$ settings. On the other hand, $\epsilon = 0.2$ is consistently the best choice, leading to the best recall-QPS curve. Based on this observation, we suggest users choose $\epsilon = 0.2$ to seek best performance with a guaranteed routing.

\subsection{Indexing Cost}
\label{sec:index-size-and-time}
In Table~\ref{tab:indexing-time}, the index size of HNSW+PEOs is 1.2x -- 2.0x larger than that of HNSW. On the two datasets with lower dimensions, the additional space overheads are more obvious. Meanwhile, FINGER requires more space cost than PEOs due to storing the information of generated subspaces. 
The indexing time of PEOs is much shorter than the time of graph construction. On the other hand, FINGER spends more indexing time due to its additional subspace for each node. 

From the index size results, the amount of additional storage for PEOs is evident for relatively low-dimensional ($d < 200$) datasets (over 90\%). Therefore, we propose another implementation, dubbed PEOs (Compact), to alleviate this issue. In this implementation, we use $L=2$ (a smaller $d$ leads to a smaller $L$) and 8 bits to compress scalars (the standard one uses 16 bits). In Figure~\ref{fig:compact} and Table~\ref{tab:compact}, 
despite slightly sacrificing search speed, a significant reduction in index size can be achieved. Note that PEOs (Compact) is still faster than other graph-based methods compared.


\begin{table}[t]
  \caption{Index size and indexing time on DEEP100M.}
  \label{table:deep100M}
  \small
  \centering
  \begin{tabular}{lcc}
    \toprule
    Method  & Index Size (GB) & Indexing Time (s)    \\
    \midrule    
    HNSW & 66.0 & 8877 \\
    HNSW+PEOs & 67.9 (+0.029x) & 5832+622 \\
    \bottomrule
  \end{tabular}
\end{table}

\subsection{Scalability}
\label{sec:100M}
We discuss how to tackle the scalability issue on datasets larger than the million scale. There are three ways to reduce the space cost of PEOs: (1) decreasing $M$, (2) decreasing $L$, and (3) using only one byte for scalar quantization. When $L \le 4$, $w_{res}$ is generally very small, meaning that we can set $w_{reg}$ to 1 and $w_{res}$ to 0 for every $\bm{e}$. In this case, for each neighbor $u$, we only need $L$ bytes ($2 \le L \le 4$) for sub-vector IDs, one byte for the norm of $\bm{u}$ and one byte for the norm of $\bm{e}$. Although such setting is not optimal for speed, it can significantly reduce space cost. 
Following this analysis, we use $L = 2$ and $M = 16$ for PEOs on DEEP100M. 
Figure~\ref{fig:deep100M} and Table~\ref{table:deep100M} show that in most cases, PEOs has 30\% performance improvement on HNSW with a 3\% additional space cost. For datasets with higher dimensions, the percentage of additional space cost can be smaller.

%% file: concl.tex
\section{Conclusion}
\label{sec:concl}
We studied the problem of probabilistic routing in graph-based ANNS, which yields a probabilistic guarantee of estimating whether the distance between a node and the query should be calculated when exploring the graph index for ANNS, thereby preventing unnecessary distance calculation. We considered two baseline algorithms by adapting locality-sensitive approaches to routing in graph-based ANNS, and devised PEOs, a novel approach to this problem. We proved the probabilistic guarantee of PEOs and conducted an empirical evaluation using two popular graph indexes on seven datasets. The results showed that PEOs is effective in enhancing the performance of graph-based ANNS and consistently outperforms the SOTA routing technique by 1.1 to 1.4 times. 


%% file: ack.tex
\section*{Acknowledgement}
This work was supported by JST CREST Grant Number JPMJCR22M2, Japan and JSPS Kakenhi Grant Numbers 23H03406, 23K17456, 23K21726, 23K24850, 23K25157.

%% file: appendix.tex
\clearpage
\appendix

\section{Frequently Used Notations}
Table~\ref{table:notation_table} shows the notations used frequently in this paper.

\begin{table}[htbp]
  \caption{Frequently used notations.}
  \label{table:notation_table}
  \vskip 0.05in
  \small
  \centering
  \begin{tabular}{lc}
    \hline
    Notation & Explanation\\
    \hline \hline
    $d$ & Data dimension \\
    $\bm{q}$ & Query\\
    $\bm{e}$ & Edge\\
    $\epsilon$ & Error rate\\
    $w_{reg}$, $w_{res}$ & Two weights associated with edges \\
    $m$ & The number of projected vectors\\
    $\{a_j^i\}$, $\{b_j\}$ & Projected vectors\\
    $\delta$ & Threshold of priority queue\\
    $L$ & The number of space partitions\\
    $J_{rel}$ & Indicator of estimation error\\
    $\mathcal N^{\theta}_{opt}$ & Optimal normal distribution in PEOs test\\
    $\mathcal N_{H}$ & Real normal distribution in PEOs test\\
    \hline
  \end{tabular}
\end{table}

\section{Proofs}
\subsection{Proof of Lemma~\ref{lemma:comparison-of-distributions}}

\begin{proof}
  We first consider the variance. By $\norm{\bm{e}}=1$, $\norm{\bm{q}}=1$ and the Cauchy-Schwarz inequality, we have
  \begin{equation}
    |\bm{e}^\top \bm{q}| \le \sqrt{\sum\limits_i \norm{\bm{q}_i}^2 \cos^2 \theta_i} 
  \end{equation}
  which means that ${\rm{Var}}{[\mathcal N^{\theta}_{\bm{e}, \bm{q}}]} \le 1 - \cos^2 {\theta}$.
  Next, we consider $\mathbb E {[\mathcal N^{\theta}_{\bm{e}, \bm{q}}]}$. To get a lower bound of $\sum\limits_i \norm{\bm{q}_i} \cos \theta_i$, we only need to solve the following linear programming problem, that is, to calculate $S$.
  \begin{equation} \label{eq:linear-programming1}
    \begin{split}
      S  = \min{\sum\limits_i{s_i}} \\
      s.t. \quad - \norm{\bm{q}_i} \le s_i \le \norm{\bm{q}_i} \qquad (1 \le i \le L) \\
      \sum\limits_i{s_i \norm{\bm{e}_i}} = \cos \theta.
    \end{split}
  \end{equation}
  By replacing $s_i$ by $t_i = s_i + \norm{\bm{q}_i}$, we only need to solve the following problem. Noting that all $ \norm{\bm{q}_i}$'s can be viewed as constants since $\bm{q}$ is fixed in the query phase.
  \begin{equation} \label{eq:linear-programming2}
    \begin{split}
      S'  = \min{ \sum\limits_i{t_i} } \\
      s.t. \quad 0 \le t_i \le 2 \norm{\bm{q}_i} \quad (1 \le i \le L) \\
      \sum\limits_i{t_i \norm{\bm{e}_i}} = \cos \theta + \sum\limits_i{\norm{\bm{q}_i} \norm{\bm{e}_i}}.
    \end{split}
  \end{equation}
  Let $\norm{\bm{q}_{\alpha(1)}} \ge \norm{\bm{q}_{\alpha(2)}} \cdots \ge \norm{\bm{q}_{\alpha(L)}}$ and $\norm{\bm{e}_{\beta(1)}} \ge \norm{\bm{e}_{\beta(2)}} \cdots \ge \norm{\bm{e}_{\beta(L)}}$, where $\alpha(\cdot)$ and $\beta(\cdot)$ are two permutations of $1, \cdots, L$. Let $U = \cos \theta + \sum\limits_i{\norm{\bm{q}_i} \norm{\bm{e}_i}}$. Suppose that $1 \le L' \le L-1$ is an integer such that the following relationship holds (noting that $U > 0$ and the leftest term is set to 0 when $L' = 0$).
  \begin{equation}
  \label{eq:relations-of-G}
    \sum\limits_{1 \le i \le L'}{\norm{\bm{q}_{\alpha(i)}} \norm{\bm{e}_{\beta(i)}}} \le \frac{U}{2} < \sum\limits_{1 \le i \le L'+1}{\norm{\bm{q}_{\alpha(i)}} \norm{\bm{e}_{\beta(i)}}}.
  \end{equation}
  Note that we do not need to consider the case $U/2 \ge \sum\limits_{1 \le i \le L}{\norm{\bm{q}_{\alpha(i)}} \norm{\bm{e}_{\beta(i)}}}$ since it does not occur for any adequate $\theta$. If it occurs, we can get the following inequality by the rearrangement inequality:
  \begin{equation}
    \cos \theta \ge \sum\limits_i{\norm{\bm{q}_i} \norm{\bm{e}_i}}.
  \end{equation}
  In this case, we definitely know the neighbor $u$ cannot be added into the priority queue. 
  
  By~\eqref{eq:relations-of-G}, we can easily calculate $S'$ in~\eqref{eq:linear-programming2} as follows: 
  \begin{equation}
    S' = 2\sum\limits_{1 \le i \le L'}{\norm{\bm{q}_{\alpha(i)}}} + \frac{U-U'}{\norm{\bm{e}_{\beta(L'+1)}}}
  \end{equation}
  where $U' = 2\sum\limits_{1 \le i \le L'}{\norm{\bm{q}_{\alpha(i)}} \norm{\bm{e}_{\beta(i)}}}$. Then we have the following inequalities:
  \begin{equation}
    \label{eq:inequalities-of-T}
    \begin{split}
      S' &\ge 2\sum\limits_{1 \le i \le L'}{\norm{\bm{q}_{\alpha(i)}}} +\frac{\cos \theta + \sum\limits_{i}{\norm{\bm{q}_{i}} e_{\min}} - U'}{e_{\max}}\\
      &\ge  \frac{\cos \theta + \sum\limits_{i}{\norm{\bm{q}_{i}} e_{\min}}}{e_{\max}}.
    \end{split}
  \end{equation} 
  Thus, we have the following result on the expected value: 
  \begin{equation}
    \frac{\cos \theta + (e_{\min} - e_{\max})\sum\limits_i{\norm{\bm{q}_i}}}{e_{\max}} \le S \le \mathbb E {[\mathcal N^{\theta}_{\bm{e}, \bm{q}}]} / \eta.
  \end{equation}
  For the case $\norm{\bm{e}_1} = \cdots = \norm{\bm{e}_L}$, because $\sum\limits_i \norm{\bm{q}_i} \cos \theta_i = \sqrt{L} \cos \theta$ always holds, we can prove the lemma. 
\end{proof}

\subsection{Proof of Theorem~\ref{theorem}}
\begin{proof}
  (1) In Sec.~\ref{sec:process-of-peos}, from the formulation of $A_r({\bm{e}})$, we can see that $A_r({\bm{e}})$ yields a threshold of the angle between $\bm{e}$ and $\bm{q}$. Therefore, we only consider the angle between them by assuming $\norm{\bm{e}} = \norm{\bm{q}} = 1$ to simplify the proof. 

  For the case when $A_r({\bm{e}}) \ge 1$, $dist(\bm{u}, \bm{q}) \ge \delta$. Hence $\bm{u}$ can be safely pruned and the PEOs test returns false. For the case when $A_r({\bm{e}}) \le 0$, $\bm{u}$ is likely to be have a distance less than $\delta$ from $\bm{q}$, and the PEOs test always returns true. Therefore, the statement can be proved by showing that the case $0 < A_r({\bm{e}}) < 1$ is $(\delta, 1-\epsilon)$-routing. 
  
  By the result in Lemma~\ref{lemma:comparison-of-distributions}, we have $w_{reg}H_1(\bm{e}) \sim \mathcal N_{reg}$ and $\sqrt{L}w_{res}H_2(\bm{e}) \sim \mathcal N_{res}$, where $\mathcal N_{reg}$ and $\mathcal N_{res}$ have the following properties:
  \begin{equation}
    \label{eq:n_reg}
    \mathcal N(w_{reg} \cos \theta_1 \sqrt{2L \ln{m}}, w_{reg}^2(1-\cos^2 \theta_1)) \preceq  \mathcal N_{reg},
  \end{equation}
  \begin{equation}
    \label{eq:n_res}
    \mathcal N_{res} = \mathcal N(w_{res} \cos \theta_2 \sqrt{2L \ln{m}}, w_{res}^2L(1-\cos^2 \theta_2))
  \end{equation}
  where $\theta_1$ is the angle between $\bm{e}_{reg}$ and $\bm{q}$, and $\theta_2$ is the angle between $\bm{e}_{res}$ and $\bm{q}$. Note that~\eqref{eq:n_reg} holds since $e[i] = e_{reg}[i]$ ($1 \le i \le L$). Suppose that $dist(\bm{u}, \bm{q}) < \delta$. Then, by the definition of $r$, the condition that $u$ can pass the PEOs test is equivalent to the following inequality:
  \begin{equation}
    \cos \theta \ge F_{\bm{e},\epsilon}(T_{r}(\bm{e})) = A_r(\bm{e}) = \cos \tilde \theta > 0.
  \end{equation}
  That is, $\tilde \theta$ is the threshold angle. By $\norm{\bm{e}} = \norm{\bm{q}} = 1$, we have 
  \begin{equation}
    \cos \theta = \bm{e}^\top \bm{q} = w_{reg} \cos \theta_1 + w_{res} \cos \theta_2.
  \end{equation}
  Thus, we have the following relationship:
  \begin{equation}
     \mathcal N_{rel} = \mathcal N(\cos \theta \sqrt{2L\ln{m}}, V) \preceq \mathcal N_{reg} + \mathcal N_{res}
  \end{equation}
  where $V = w_{reg}^2(1-\cos^2 \theta_1)+w_{res}^2L(1-\cos^2 \theta_2)$.
  Then, we have the following claim.
  
  \textbf{Claim:} if the following inequality holds:
  \begin{equation}
    \label{eq:V}
    V \le w_{reg}^2 + Lw_{res}^2 - \frac{L \cos^2 \theta}{L+1},
  \end{equation}
  then, $\Pr[H(\bm{e}) \ge T_{r}(\bm{e})] \ge 1 - \epsilon$.

  \textbf{Proof of Claim:} Clearly, $H(\bm{e})$ can be viewed as a random variable and $H(\bm{e}) \sim \mathcal N_{reg} + \mathcal N_{res}$. Then, we have
  \begin{equation}
    \Pr[H(\bm{e}) \ge Q_{\mathcal N_{reg} + \mathcal N_{res}}(\epsilon)] = 1- \epsilon 
  \end{equation}
  where $Q_{\mathcal N}$ denotes the quantile function with respect to the normal distribution $\mathcal N$. By~\eqref{eq:V}, we have $\mathcal N^{\bm{e}, \cos \theta}_{\min} \prec \mathcal N_{rel} $. Then, it can be seen that $Q_{\mathcal N_{rel}}(\epsilon) \ge Q_{N^{\bm{e}, \cos \theta}_{\min}}(\epsilon)$, where $\epsilon \le 0.5$. For two normal distributions $\mathcal N(\mu_1,\sigma_1)$ and $\mathcal N(\mu_2,\sigma_2)$, when $\epsilon \le 0.5$, $\mu_1 \le \mu_2$ and $\sigma_1 \ge \sigma_2$, i.e., $\mathcal N(\mu_1,\sigma_1) \prec \mathcal N(\mu_2,\sigma_2)$, we have the following relationship:
  \begin{equation}
    \label{eq:quatile-function-of-normal-distribution}
    \mu_1 + \sigma_1\sqrt{2}\erf^{-1}(2\epsilon - 1) \le \mu_2 + \sigma_2\sqrt{2}\erf^{-1}(2\epsilon - 1).
  \end{equation}
  That is, $Q_{\mathcal N(\mu_1,\sigma_1)} \le Q_{\mathcal N(\mu_2,\sigma_2)}$. On the other hand, because $\mathcal N^{\bm{e}, \cos \tilde \theta}_{\min} \preceq \mathcal N^{\bm{e}, \cos \theta}_{\min}$, we have
  \begin{equation}
    \begin{split}
      \Pr[H(\bm{e}) \ge T_{r}(e)] &= \Pr[H(\bm{e}) \ge Q_{\mathcal N^{\bm{e}, \cos \tilde \theta}_{\min}}(\epsilon)]\\
      &\ge \Pr[H(\bm{e}) \ge Q_{\mathcal N^{\bm{e}, \cos \theta}_{\min}}(\epsilon)]\\      
      &\ge  \Pr[H(\bm{e}) \ge Q_{\mathcal N_{rel}}(\epsilon)]\\
      &\ge  \Pr[H(\bm{e}) \ge Q_{\mathcal N_{reg}+ \mathcal N_{res}}(\epsilon)]\\
      &= 1- \epsilon.
    \end{split}
  \end{equation}
  Thus, the claim is proved. Note that the result in inequality~\eqref{eq:quatile-function-of-normal-distribution} is used three times.

  Now, the remaining work is to prove the inequality in~\eqref{eq:V}. First, we introduce $W$ such that
  \begin{equation}
  \label{eq:expression-of-W}
    W = w_{reg}^2 \cos^2 \theta_1 + Lw_{res}^2 \cos^2 \theta_2.
  \end{equation}
  To get a lower bound of $W$, we only need to solve the following linear programming problem:
  \begin{equation} \label{eq:linear-programming3}
    \begin{split}
      W_{\min} = \min\{s_1^2 + Ls_2^2\} \\
      s.t. \quad s_1 + s_2  = \cos \theta \\
      -w_{reg} \le s_1 \le w_{reg}\\
      -w_{res} \le s_2 \le w_{res}.
    \end{split}
  \end{equation}
  Because 
  \begin{equation}
    \begin{split}
      s_1^2 + Ls_2^2 &= (L+1)(s_2 - \frac{\cos \theta}{L+1})^2 + \frac{L\cos^2 \theta}{L+1}\\ 
      &\ge \frac{L\cos^2 \theta}{L+1}, 
    \end{split}
  \end{equation}
  we prove the first statement. 

  (2) For the second statement, we first note that 
  \begin{equation}
    \Phi_{\lambda \mathcal N_1}(Q_{\lambda \mathcal N_2}(\epsilon)) = \Phi_{\mathcal N_1}(Q_{\mathcal N_2}(\epsilon)) 
  \end{equation}
  where $\lambda > 0$, $\mathcal N_1$ and $\mathcal N_2$ are two normal distributions, and $\Phi$ is the CDF of normal distribution. Suppose that $dist(\bm{u}, \bm{q}) \ge \delta$ and $H(\bm{e}) / \sqrt{2 L \ln m}\sim \mathcal N_H$. By the analysis for the first statement, we have
  \begin{equation}
   \mathbb E[\mathcal N_H] = \cos \theta.
  \end{equation}
  \begin{equation}
    {\rm{Var}}[\mathcal N_H] \le \frac{w_{reg}^2 + Lw_{res}^2}{2L\ln m} - \frac{\cos ^2 \theta}{2(L+1)\ln m}. 
  \end{equation}
  Under the condition that $\cos \theta \le \tilde F^{-1}_{\tilde \theta}(\epsilon)$, we can immediately prove the second statement by using a similar result in~\eqref{eq:quatile-function-of-normal-distribution} again.

  (3) For the third statement, let $\Delta' = {\rm{Var}}[\mathcal N_H] - {\rm{Var}}[\mathcal N^{\theta}_{\opt}]$ and $\tau = 2L\ln m$, On one hand, we have 
  \begin{equation}
    \tau \Delta' \le (L-1)w^2_{res} + \frac{\cos^2 \theta}{L+1} \le \frac{2L}{(L+1)^2}.
  \end{equation}
  On the other hand, we have 
  \begin{equation}
    \begin{split}
      \tau \Delta' &\ge (L-1)w^2_{res} + \frac{\cos^2 \theta}{L+1}-(L+1)(w_{res}+\frac{\cos \theta}{L+1})^2\\
      &\ge -2w^2_{res}-2w_{res}\\
      &\ge - \frac{2L+4}{(L+1)^2}.
    \end{split}
  \end{equation}
  Therefore, the third statement is proved.
\end{proof}

\subsection{Proof of Lemma~\ref{lemma:bounds-of-wreg}}
\begin{proof}
  Since $\bm{e} \sim U(\mathbb R^d)$, we can generate $\tilde{\bm{e}} \sim \mathcal N(0, I)$. It can be seen that $\tilde{\bm{e}} / \norm{ \tilde{\bm{e}} } \sim U(\mathbb S^{d-1})$. Then we have $\norm{\bm{e}_1}^2 \sim Y/(Y +Z) = Beta(d'/2,(d-d')/2)$, where $Y \sim \mathcal X^2(d')$ and $Z \sim \mathcal X^2(d-d')$. By the summation of expectations, we have the following result.
  \begin{equation}
    \mathbb E[w_{reg}(L,d)] = \mathbb E_{X}[\sqrt{LX}]
  \end{equation}
  where $X \sim Beta(d'/2,(d-d')/2)]$. Let $\alpha = d'/2$ and $\beta = (d-d')/2$. Then we have 
  \begin{equation}
    \label{eq:expectation-of-w_reg}
    \begin{split}
      \mathbb E[w_{reg}(L,d)] &=  \sqrt{L} \int_{0}^{1} \frac{x^{\alpha-\frac{1}{2}}(1-x)^{\beta-1}}{B(\alpha,\beta)} \,dx\\
      &= \sqrt{L} \mathbb E_{X \sim Beta(\alpha-\frac{1}{2},\beta)}[X] \cdot \frac{B(\alpha-\frac{1}{2}, \beta)}{B(\alpha,\beta)}\\
      &= \sqrt{L} \cdot \frac{\alpha-\frac{1}{2}}{\alpha-\frac{1}{2} + \beta} \cdot \frac{B(\alpha-\frac{1}{2}, \beta)}{B(\alpha,\beta)}\\
      &= \frac{\sqrt{L}\Gamma(\frac{d'-1}{2})\Gamma(\frac{d}{2})(d'-1)}{\Gamma(\frac{d'}{2})\Gamma(\frac{d-1}{2})(d-1)}\\
      &\ge \frac{(d'-1)\sqrt{2Ld-3L}}{(d-1) \sqrt{2d'+2\sqrt{3}-6}}.
    \end{split}
  \end{equation}
  We use the following Gautschi's inequality to get the last inequality in~\eqref{eq:expectation-of-w_reg}. 
  \begin{equation}
    \sqrt{x+\frac{1}{4}} \le \frac{\Gamma(x+1)}{\Gamma(x+\frac{1}{2})}\le \sqrt{x + \frac{\sqrt{3}}{2} - \frac{1}{2}}.
  \end{equation}
\end{proof}

\section{Comparison of Routing Tests}
\label{sec:different-tests}
To compare the three routing tests in this paper (SimHash, RCEOs, and PEOs), we first introduce the following definition.
\begin{definition}
  \label{def:comparison-of-ceos-and-simhash}
  Let $\theta$ be the angle of $\bm{e}$ and $\bm{q}$ to be estimated, and let $X(\theta)$ and $Y(\theta)$ be two random variables regarding $\theta$, such that (1) $\mathbb E[X(\theta)] \in C^1$ and $\mathbb E[Y(\theta)] \in C^1$ are two strictly monotone functions of $\theta$, and (2) ${\rm{Var}}[X(\theta)]$ and ${\rm{Var}}[Y(\theta)]$ are continuous. Let $\gamma(\theta) = |(\mathbb E[X(\theta)])'| / |(\mathbb E[Y(\theta)])'|$, where $|(\mathbb E[Y(\theta)])'| > 0$. We say that the estimator associated with $Y$ is better than that associated with $X$ for $\theta$, if ${\rm{Var}}[X(\theta)] > {\rm{Var}}[\gamma(\theta) Y(\theta)]$.
\end{definition}
Roughly speaking, for a given $\theta$, $Y$ is considered to be better than $X$ if the distributions associated with $X$ can distinguish the angles around $\theta$ more easily than those associated with $Y$. Based on Definition~\ref{def:comparison-of-ceos-and-simhash}, we have the following result.

\begin{lemma}
  \label{lemma:varaince-comparison}
  Suppose that SimHash uses $n$ hash functions and RCEOs uses $m$ projected vectors. If $m > \exp(n/[2\theta(\pi-\theta)])$, RCEOs is better than SimHash for $\theta$ ($0 < \theta < \pi$).
\end{lemma}

Before the proof of Lemma~\ref{lemma:varaince-comparison}, to explain why we use Definition~\ref{def:comparison-of-ceos-and-simhash} to compare the estimators associated with $X$ and $Y$, we prove the following claim.

\textbf{Claim:} Suppose that $0 < \tilde \theta_1  < \tilde \theta_2 < \pi$ and $Y$ is better than $X$ for all $\theta$'s, where $\tilde \theta_1 < \theta < \tilde \theta_2$. For an arbitrary $\theta_1$, where $\tilde \theta_1 < \theta_1 < \tilde \theta_2$, we can find a $\epsilon > 0$ such that for any $\theta_2$, where $\theta_1-\epsilon \le \theta_2 \le \theta_1 + \epsilon$, ${\rm{Var}}[X(\theta_1)] > {\rm{Var}}[\gamma Y(\theta_1)]$ and ${\rm{Var}}[X(\theta_2)] > {\rm{Var}}[\gamma Y(\theta_2)]$, where $\gamma = |\mathbb E[X(\theta_1)] - \mathbb E[X(\theta_2)]| / |\mathbb E[Y(\theta_1)] - \mathbb E[Y(\theta_2)]|$.

\begin{table*}[hbtp]
  \caption{Comparison of routing tests when $d = 384$ and $\theta = \pi/2$. The baseline is SimHash ($n=64$). We show under which conditions, RCEOs, PEOs (opt), and PEOs can approximately outperform SimHash.}
  \label{table:test_comparison}
  \vskip 0.05in
  \small
  \centering
  \begin{tabular}{lccc}
    \hline
    Routing Test  & \#Projected Vectors & $L$ & Code Length (Byte)  \\
    \hline \hline
    SimHash & $n = 64$ & NA  &  8 \\
    RCEOs & $m > 428957$  & $L = 1$ & 4 \\
    PEOs (opt) & $m = 128$  & $L > 2.67$  & 4 (3+1) \\
    PEOs  &  $m = 128$  & $ L > 2.69$  &  4 (3+1)\\
    \hline
  \end{tabular}
\end{table*}

Proof of the claim: this result can be immediately obtained by the continuity of ${\rm{Var}}[X(\theta)]$, ${\rm{Var}}[Y(\theta)]$, $\gamma(\theta) = |(\mathbb E[X(\theta)])'| / |(\mathbb E[Y(\theta)])'|$ and the mean value theorem.

This claim shows that, under the conditions in the claim, for two close angles $\theta_1$ and $\theta_2$, the variance associated with $Y$ is smaller than that associated with $X$ when the difference of expected values are the same (after scaling).

\begin{proof}
  By symmetry, we only need to consider the case $0 < \theta \le \pi/2$. To make the ranges of $\theta$ and $\cos \theta$ the same, we consider $2 (n-\#Col)/n$ in SimHash and $f(m) \bm{q}^{\top}\bm{a}_1$ in RCEOs, where $\#Col$ denotes the collision number of the pair $(\bm{e}, \bm{q})$ in SimHash and $f(m) = 1/ \sqrt{2 \ln m}$. By the lemma of SimHash (Lemma~\ref{lem:simhash}), we have 
  \begin{equation}
    X(\theta) = 2 (n-\#Col)/n \sim \frac{2}{n} \times B(n, \frac{{ \theta }}{{\pi}}).
  \end{equation}
  By the lemma of RCEOs (Lemma~\ref{lem:rceos}), we have
  \begin{equation}
    Y(\theta) = f(m) \bm{q}^{\top}\bm{a}_1 \sim N(\cos{\theta}, f^2(m) \sin^2{\theta})
  \end{equation}
  where $X(\theta)$ and $Y(\theta)$ are two random variables depending on $\theta$. It is easy to verify that all the regularity conditions in Definition~\ref{def:comparison-of-ceos-and-simhash} are satisfied. Since both of the ranges of $2 \theta / \pi $ and $\cos{\theta}$ are $[0,1]$, and they are two monotone functions of $\theta$, it is feasible to compare them by the criterion in Definition~\ref{def:comparison-of-ceos-and-simhash}. First, it is easy to see that $\gamma(\theta) = 2/(\pi sin\theta)$. On the other hand, we have
  \begin{equation}
    {\rm{Var}}[X(\theta)] = 4 \theta(\pi - \theta) / (n \pi ^ 2).
  \end{equation}
  \begin{equation}
    {\rm{Var}}[\gamma(\theta) Y(\theta)] = 2 / ( \pi ^ 2 \ln m).
  \end{equation}
  Then, we can see that 
  \begin{equation}
    m > \exp(n/[2\theta(\pi-\theta)]) \Leftrightarrow \rm{Var}[X(\theta)] / \rm{Var}[\gamma(\theta)Y(\theta)] > 1.
  \end{equation}
\end{proof}
Next, we consider PEOs (opt) and PEOs, where $w_{res}$ is assumed to be 0 in PEOs (opt). For PEOs (opt), the distribution associated with $\cos \theta$ can be approximated by $\mathcal N^{\theta}_{opt}$, and for PEOs, the distribution associated with $\cos \theta$ can be approximated by $\mathcal N^{\theta}_{act}$, which is defined as follows:
\begin{equation}
  \mathcal N^{\theta} _{\opt}\sim \mathcal N(\cos{\theta}, \frac{\sin^2{\theta}}{2L\ln m}),
\end{equation}
\begin{equation}
  \mathcal N^{\theta}_{act} \sim \mathcal N(\cos{\theta}, \frac{\sin^2{\theta}(w^2_{reg} + Lw^2_{res})}{2L\ln m}).
\end{equation}
That is, $W$ in~\eqref{eq:expression-of-W} is approximated by $(w^2_{reg} + Lw^2_{res})\cos^2 \theta$. According to the previous analysis, we can see that, to make them outperform SimHash, $m$ and $L$ should satisfy the following requirements, where $m$ is sufficiently large.
\begin{equation}
  {\rm{\textbf{PEOs(opt):}}} \quad L\ln m > n/[2\theta(\pi-\theta)].
\end{equation}
\begin{equation}
  {\rm{\textbf{PEOs:}}} \quad L\ln m > (w^2_{reg} + Lw^2_{res}))n/[2\theta(\pi-\theta)].
\end{equation}

Based on the results above, we can approximately compare different tests by a concrete example. Let $d = 384$ and $\theta = \pi/2$. Suppose that SimHash ($n = 64$) is regarded as the baseline. For RCEOs, PEOs (opt) and PEOs, we check how many projected vectors are needed to make them outperform SimHash, as shown in Table~\ref{table:test_comparison}. Here, $w_{res}$ is taken as the expected value under the isotropic distribution.

\section{Technical Extensions}
\subsection{Dimension Permutation}
\label{sec:dimension-permutation}
We describe the details of the optional dimension permutation in PEOs. Only in this section, we use $e_j$ to denote the $j$-th coordinate of $\bm{e}$, instead of the projection on the $j$-th subspace. We introduce the following notation.
\begin{equation}
  {\rm{Avg}}(e_j)=\sum\limits_{e \in E}(e_j)^2/\size{E} 
\end{equation}
where $E$ denotes the edge set of the graph index. For the dimension permutation in PEOs, we use the following greedy algorithm to permute the coordinates of data vectors, divided into four steps: 

(1) We generate $L$ empty sets, denoted by $S_1, S_2, \cdots, S_L$, each representing a subspace. 

(2) We calculate ${\rm{Avg}}(e_j)$ for every dimension $j$ and sort the dimensions in the ascending order of ${\rm{Avg}}(e_j)$. 

(3) We execute the allocation procedure by $d' = d/L$ rounds. In the $l$-th round, for each $1 \le j \le L$, we allocate dimension $(l - 1)L + j$, to set $S_i$, where $S_i$ has the greatest $\sum\limits_{k \in S}{\rm{Avg}}(e_k)$ among all the sets that have not been added any dimension in this round. 

(4) After the allocation in Step 3, we permute the coordinates of all $\bm{e}$'s such that coordinate $e_j$ appears in the $i$-th sub-vector of $\bm{e}$ if $j$ is in $S_i$.

When dimension permutation is finished, we store the permutation, in order to permute the coordinates of the query for search. 

\subsection{Extension to MIPS}
\label{extension-to-mips}
To support MIPS, we only need to use the following $A_r(\bm{e})$ to replace the one in~\eqref{eq:A_r(e)}:
\begin{equation}
  A_r(\bm{e}) = \frac{\bm{p}^{\top}\bm{q} -\bm{v}^{\top}\bm{q}}{\norm{\bm{q}} \norm{\bm{e}}}
\end{equation}
where $\bm{p}$ denotes the element in the temporary result list having the smallest inner product with $\bm{q}$. The remaining procedure is completely same as that of $\ell_2$ metric. 

\begin{figure*}[t]
  \centering
  \subfloat[GloVe200-angular, $K=10$]{\includegraphics[width=.7\columnwidth]{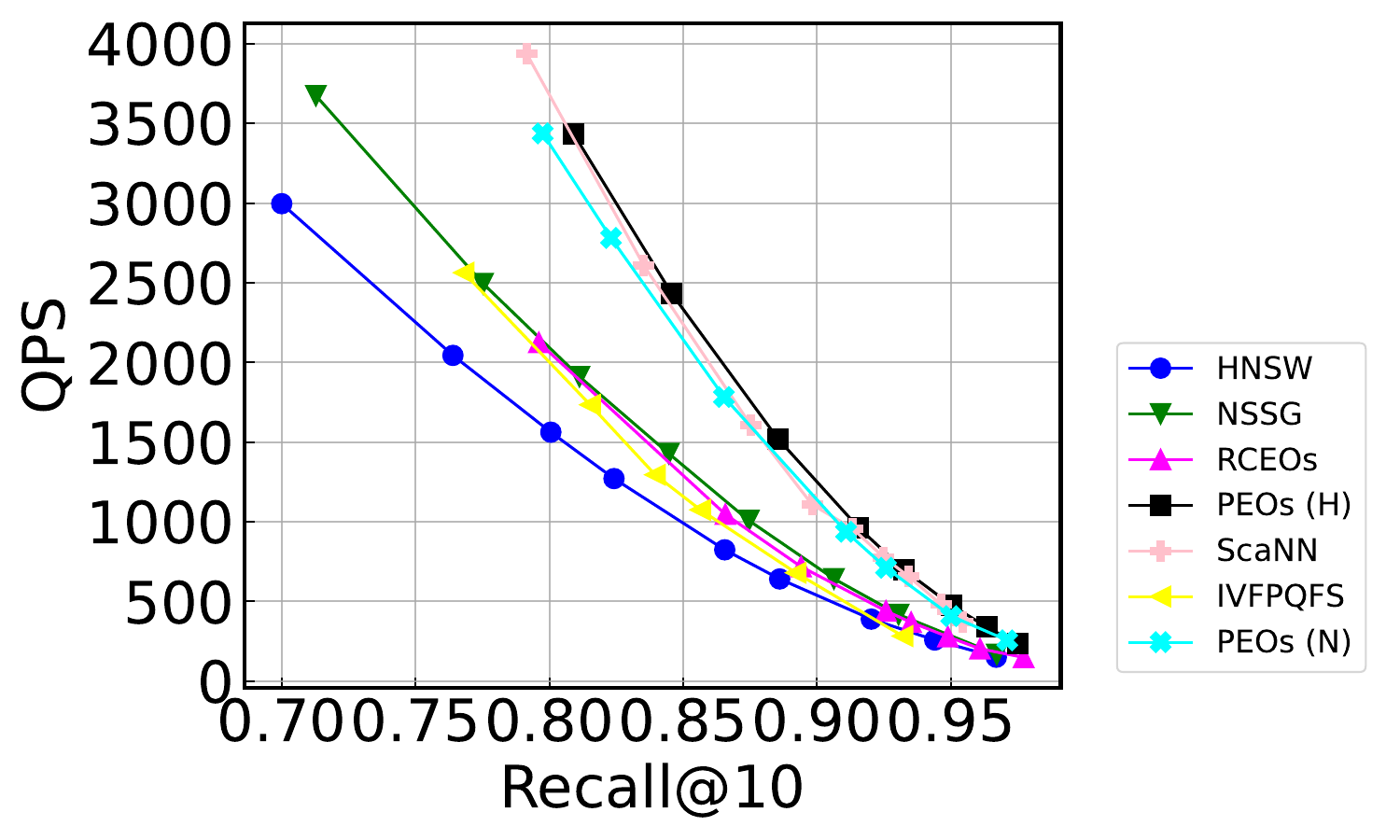}}
  \subfloat[GloVe300-$\ell_2$, $K=10$]{\includegraphics[width=.7\columnwidth]{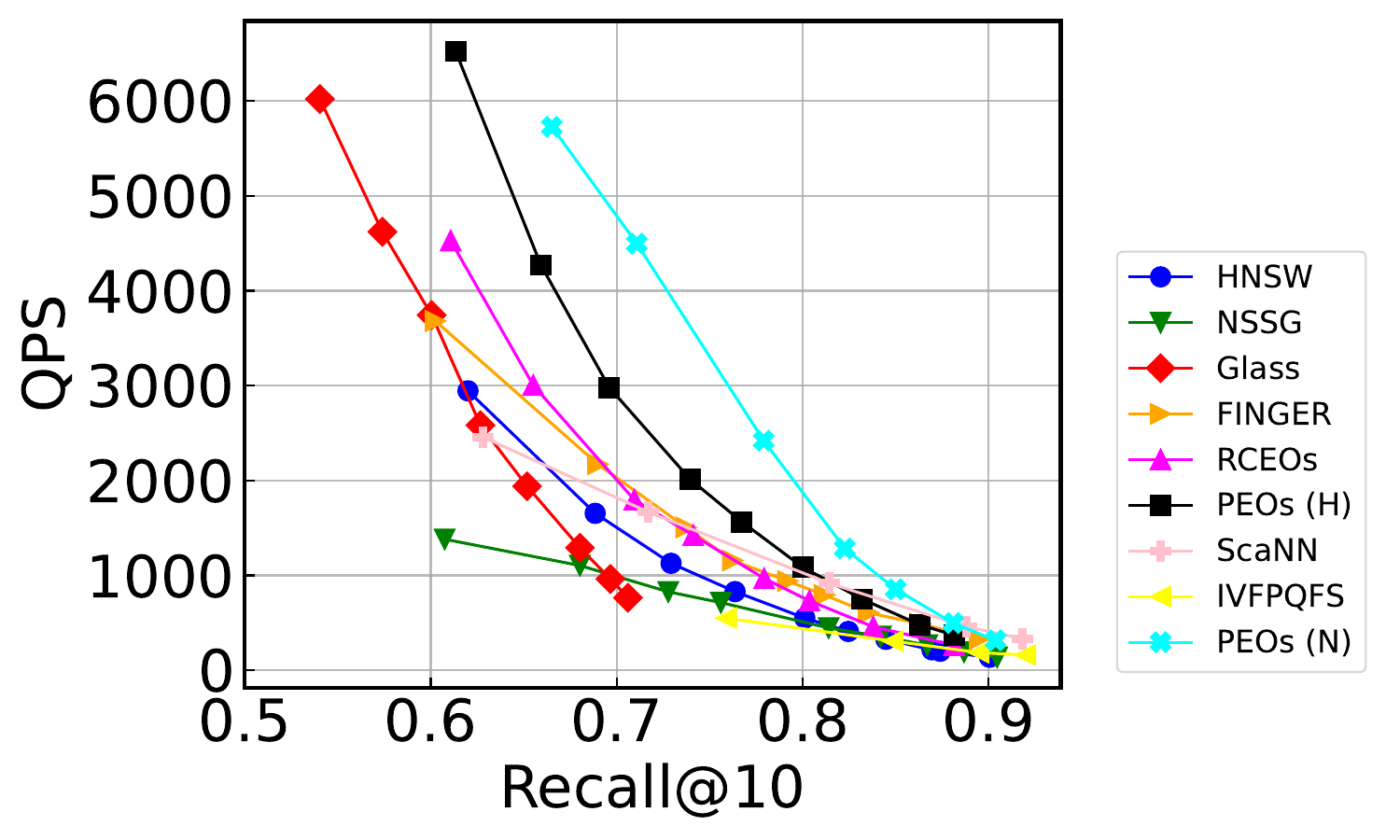}}
  \subfloat[DEEP10M-angular, $K=10$]{\includegraphics[width=.7\columnwidth]{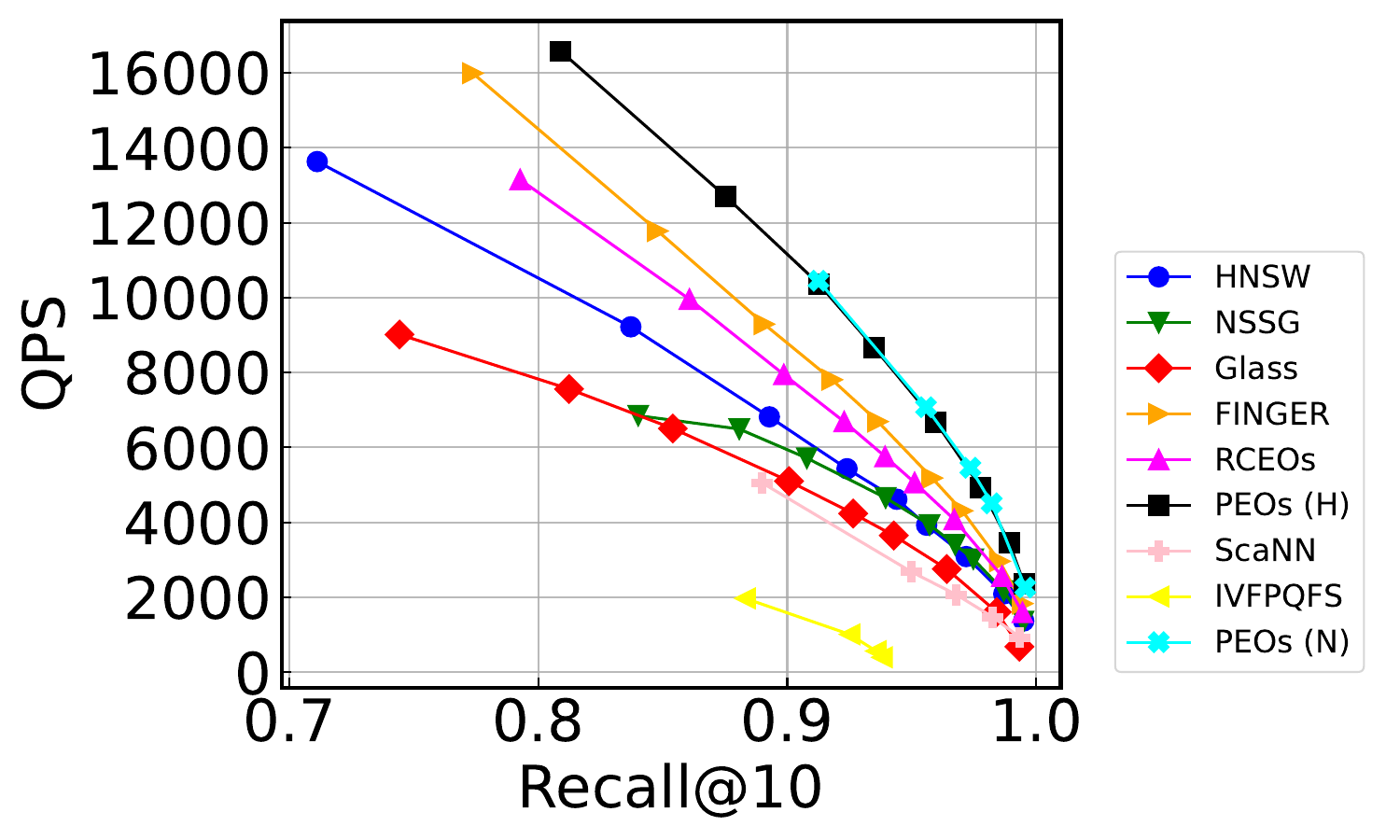}}
  
  \subfloat[SIFT10M-$\ell_2$, $K=10$]{\includegraphics[width=.7\columnwidth]{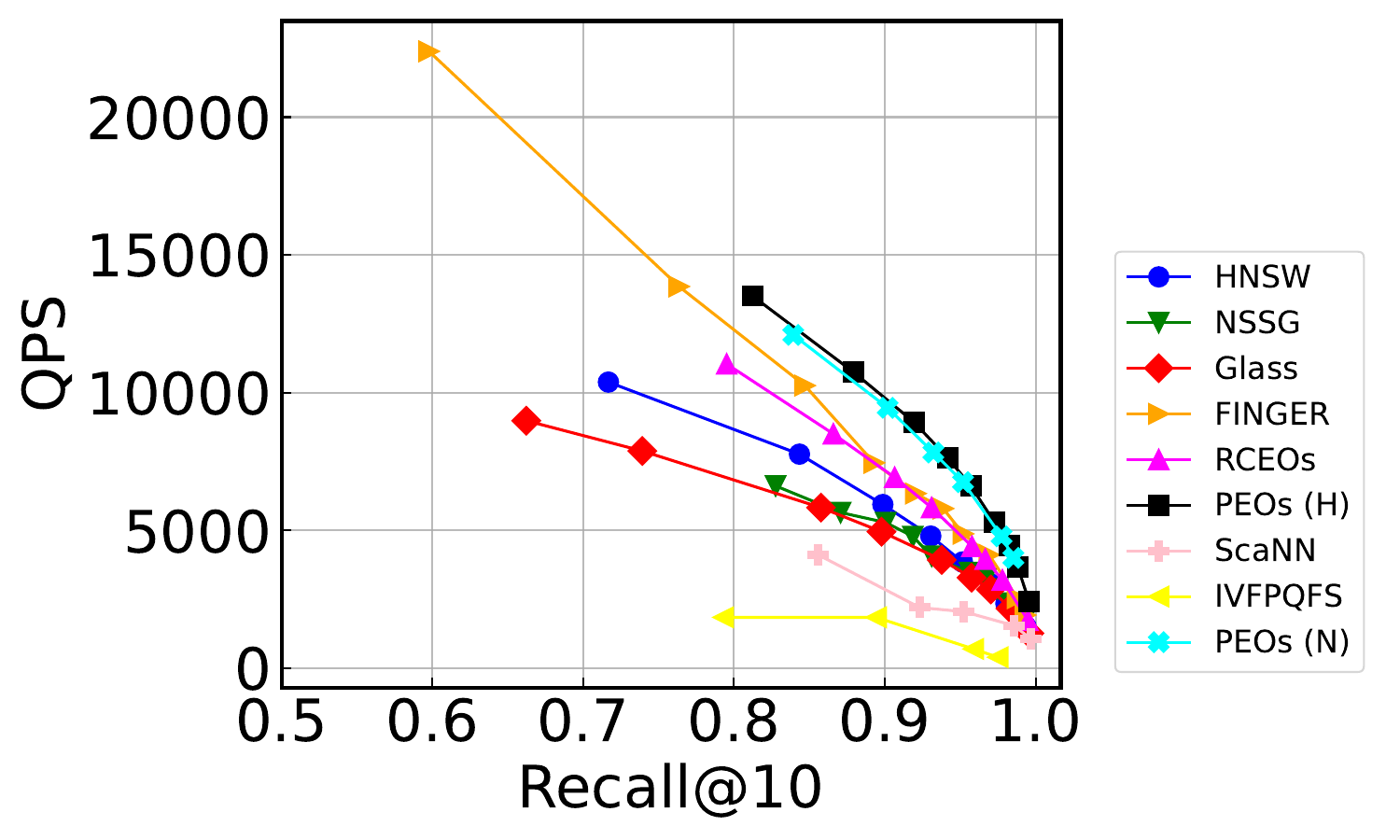}}
  \subfloat[Tiny5M-$\ell_2$, $K=10$]{\includegraphics[width=.7\columnwidth]{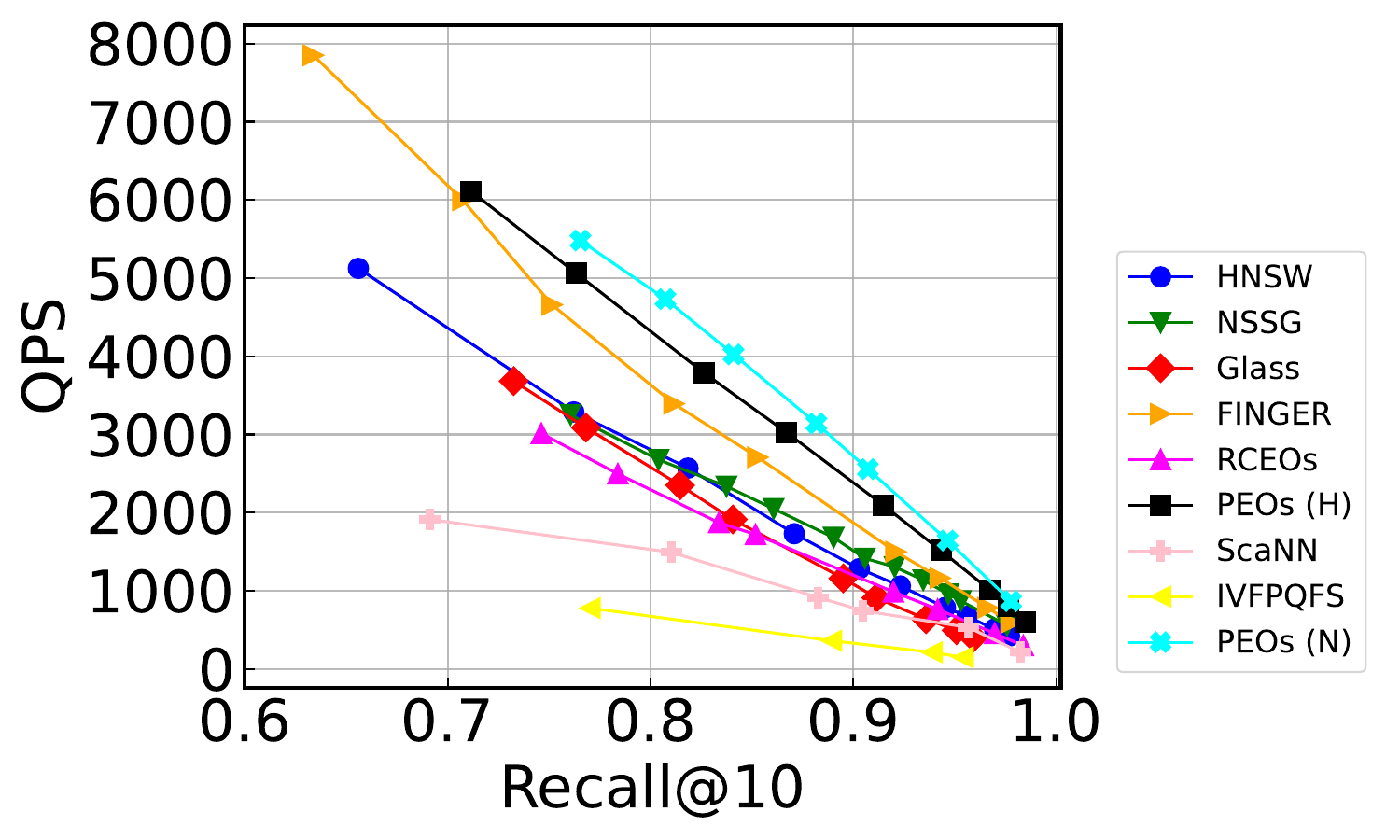}}
  \subfloat[GIST-$\ell_2$, $K=10$]{\includegraphics[width=.7\columnwidth]{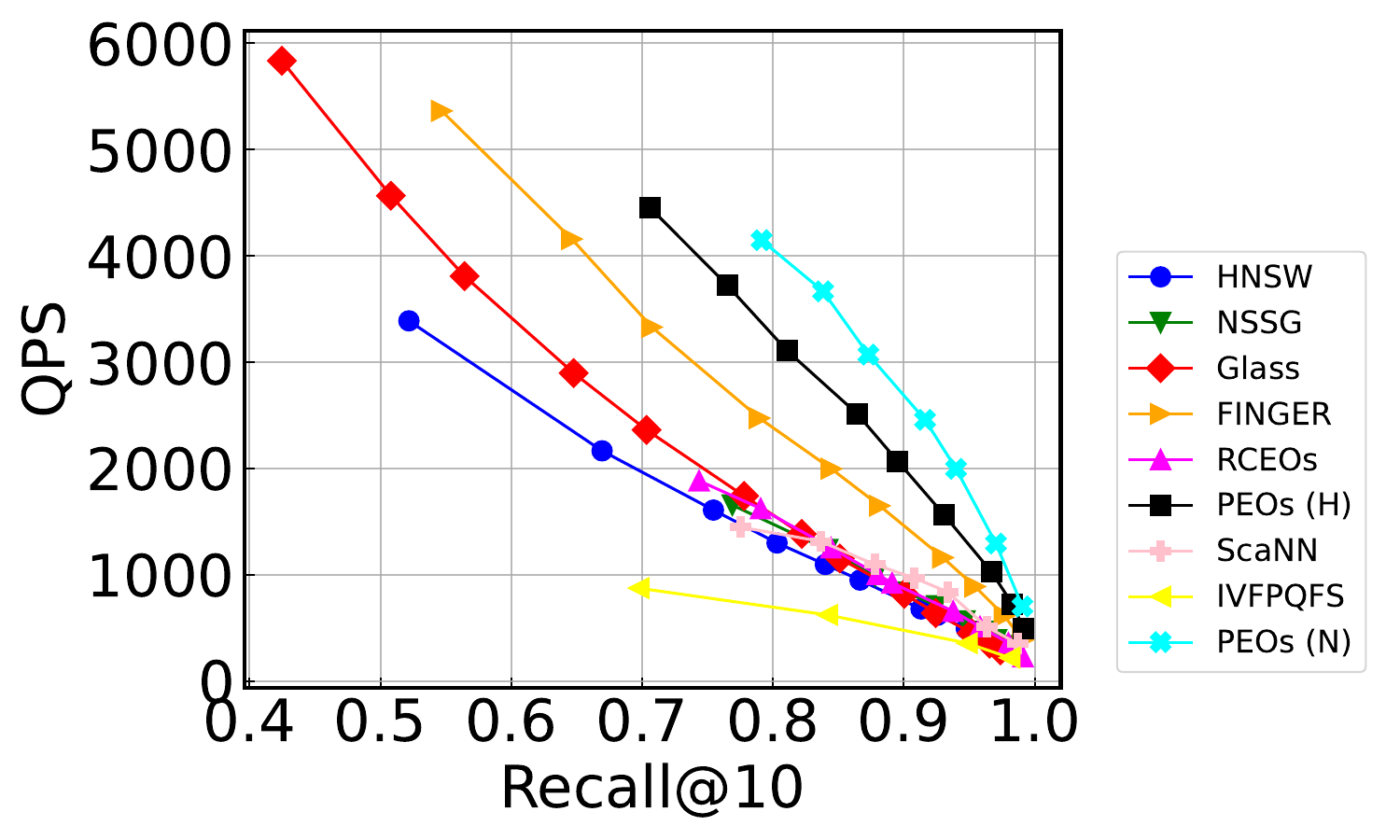}}

  \subfloat[GloVe200-angular, $K=1$]{\includegraphics[width=.7\columnwidth]{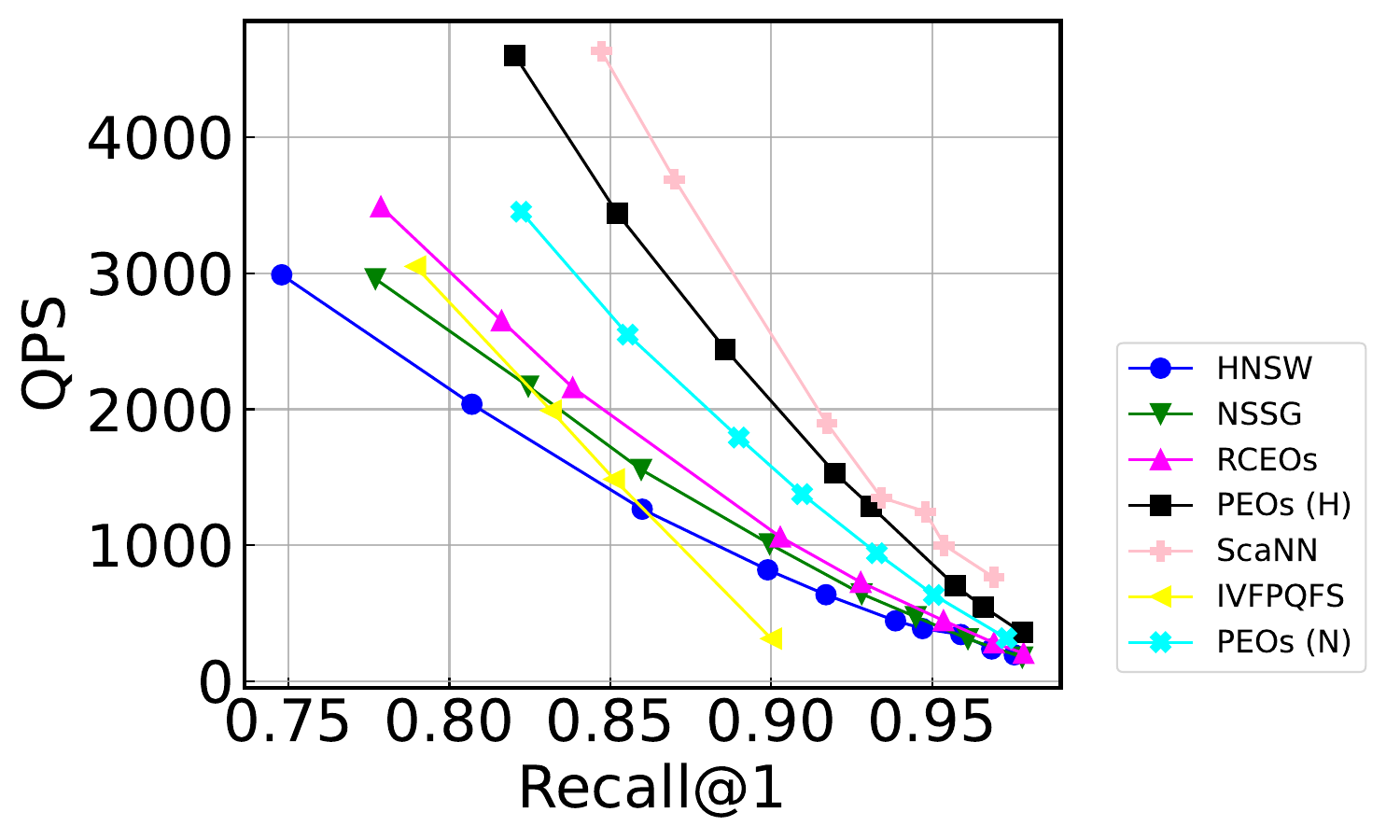}}
  \subfloat[GloVe300-$\ell_2$, $K=1$]{\includegraphics[width=.7\columnwidth]{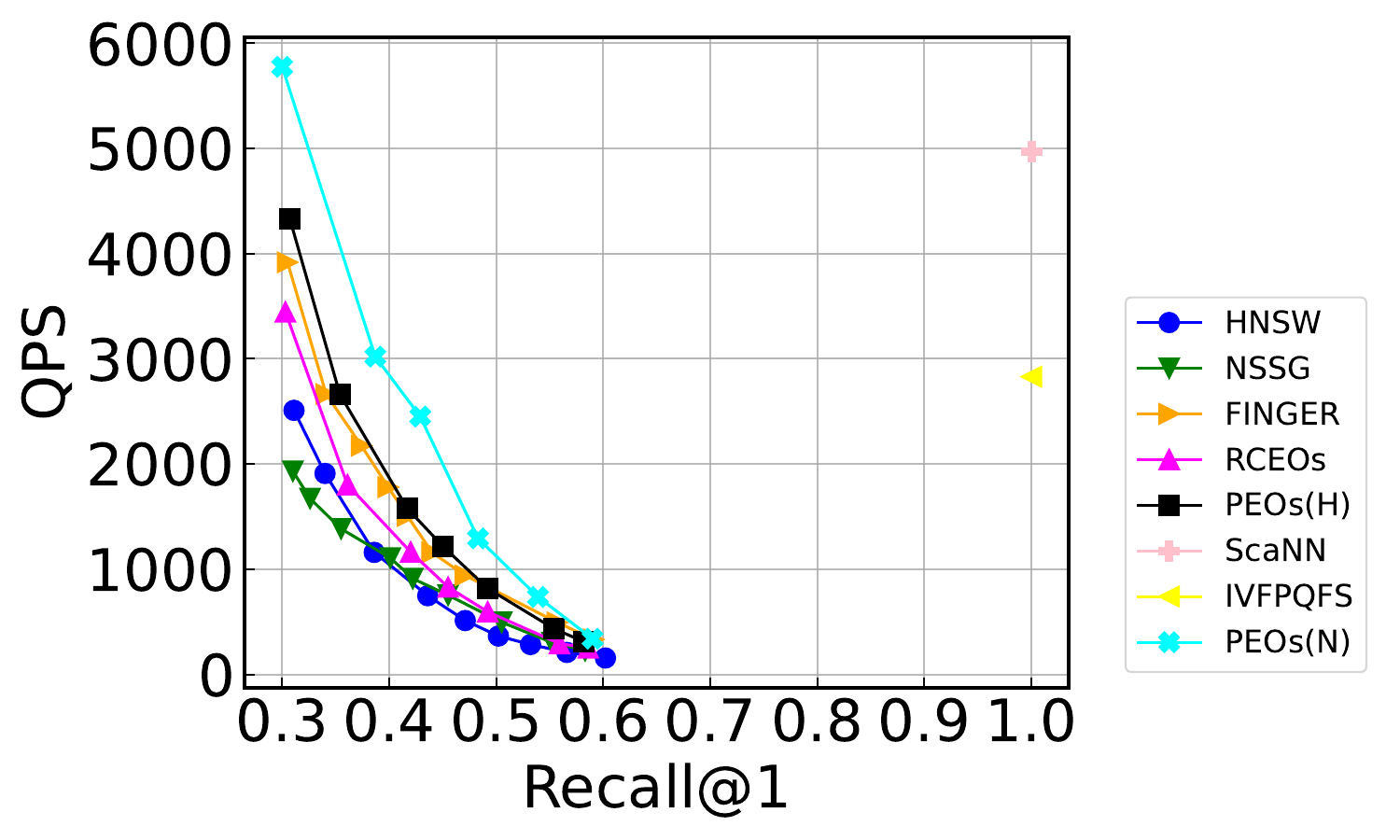}}
  \subfloat[DEEP10M-angular, $K=1$]{\includegraphics[width=.7\columnwidth]{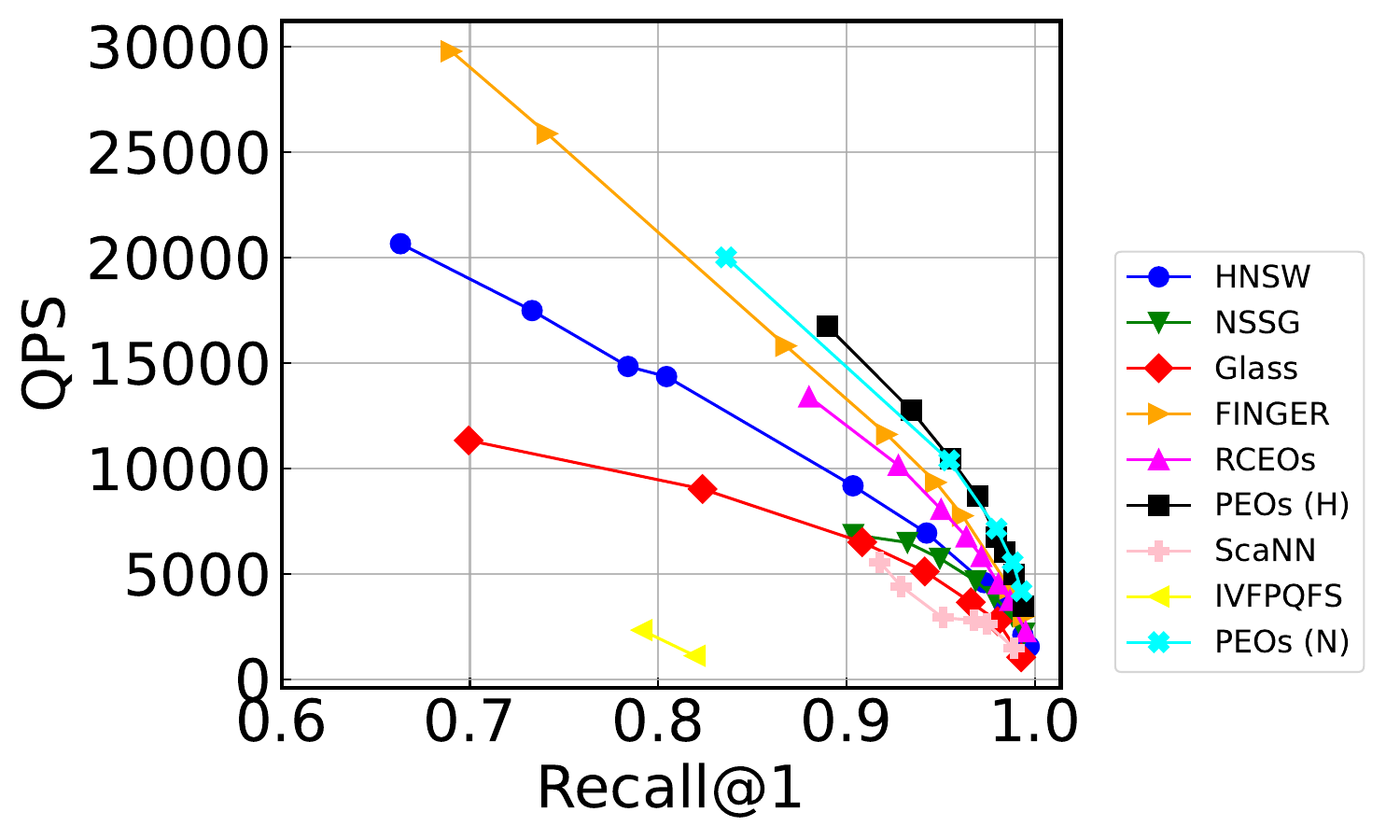}}
  
  \subfloat[SIFT10M-$\ell_2$, $K=1$]{\includegraphics[width=.7\columnwidth]{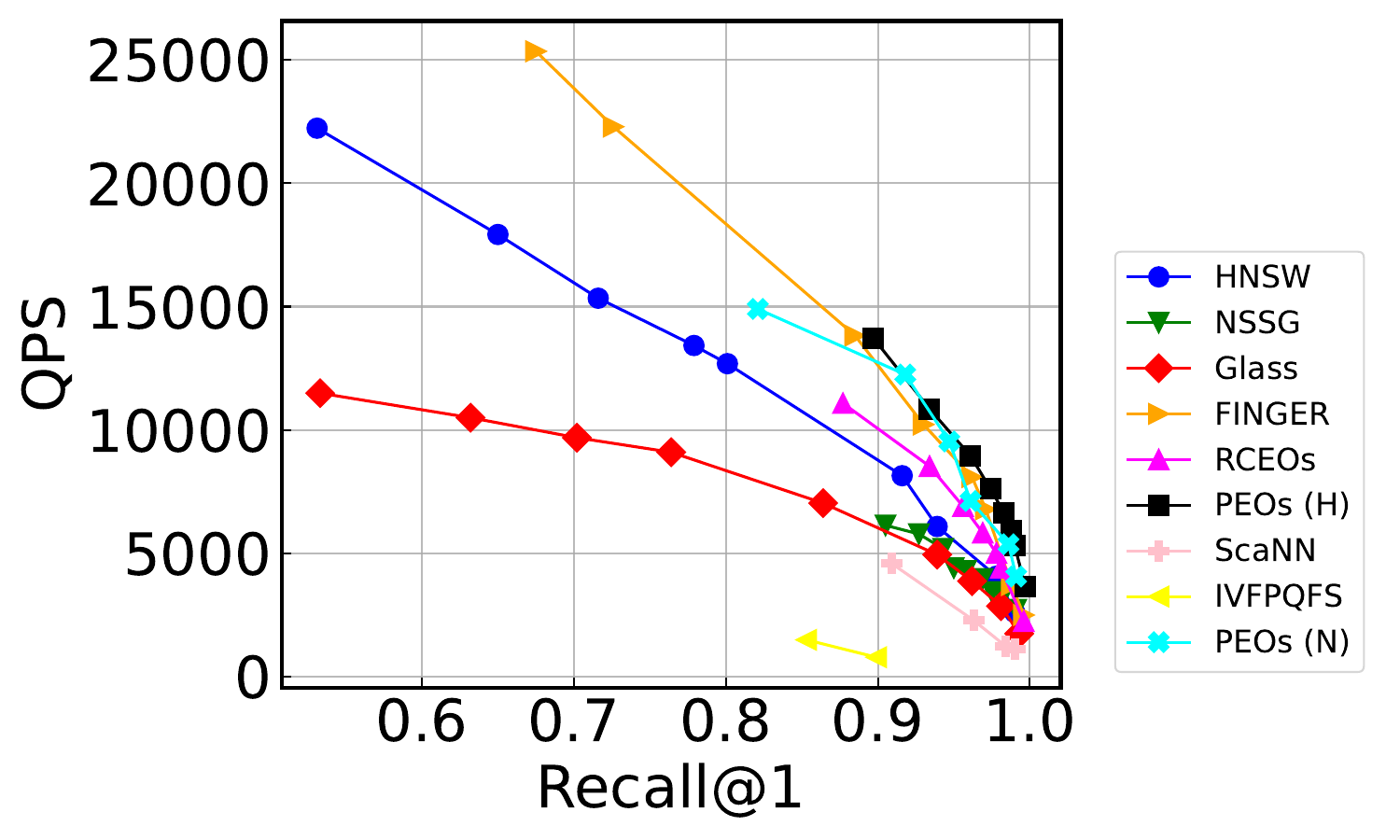}}
  \subfloat[Tiny5M-$\ell_2$, $K=1$]{\includegraphics[width=.7\columnwidth]{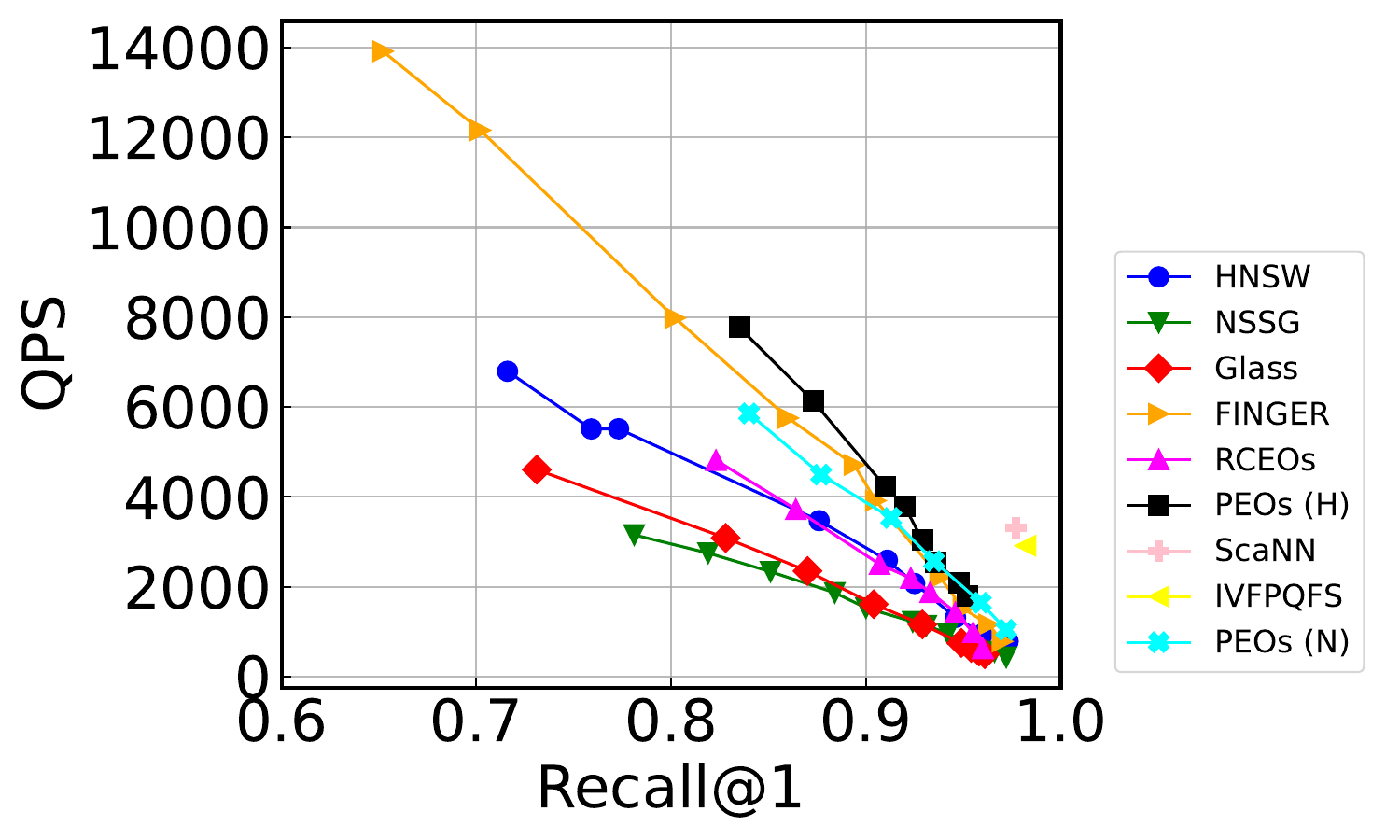}}
  \subfloat[GIST-$\ell_2$, $K=1$]{\includegraphics[width=.7\columnwidth]{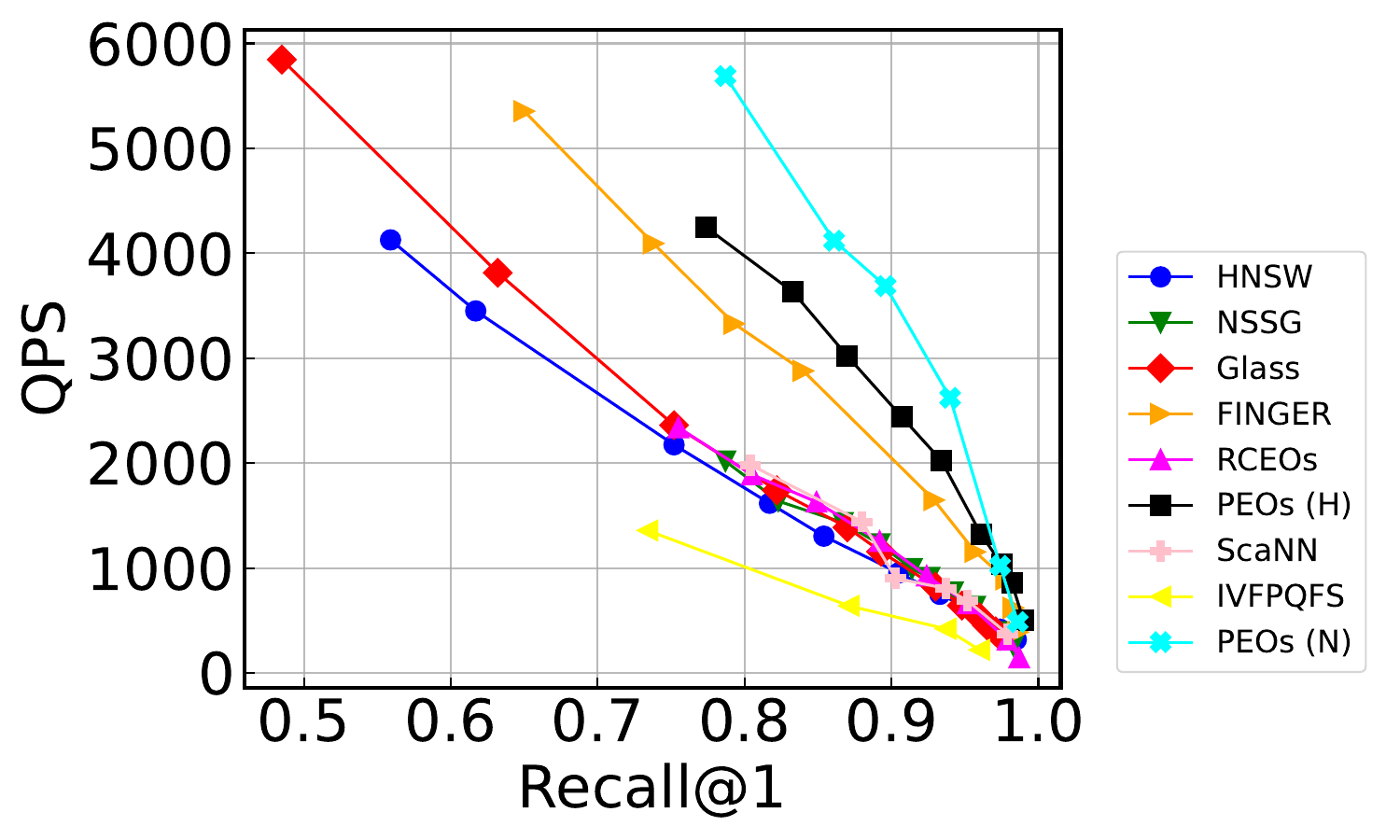}}
 
  \caption{Recall-QPS evaluation, $K = 10$ and $K = 1$. PEOs (H) denotes HNSW+PEOs and PEOs (N) denotes NSSG+PEOs.}
  \label{fig:performance-different-k}
\end{figure*}



\begin{figure*}[t]
  \centering
  \subfloat[GloVe200-angular, \#Dist. Calc.]{\includegraphics[width=.7\columnwidth]{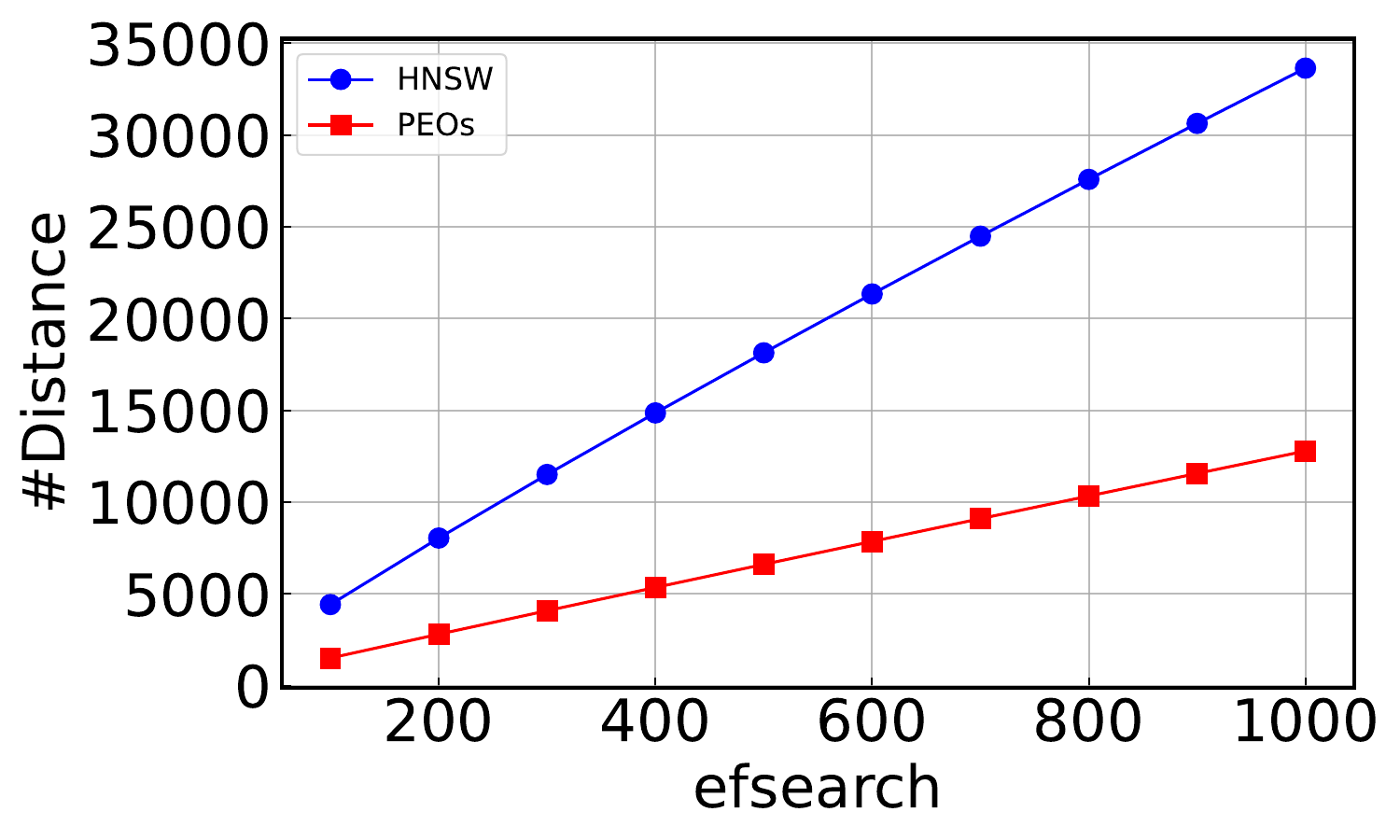}} 
  \subfloat[GloVe300-$\ell_2$, \#Dist. Calc.]{\includegraphics[width=.7\columnwidth]{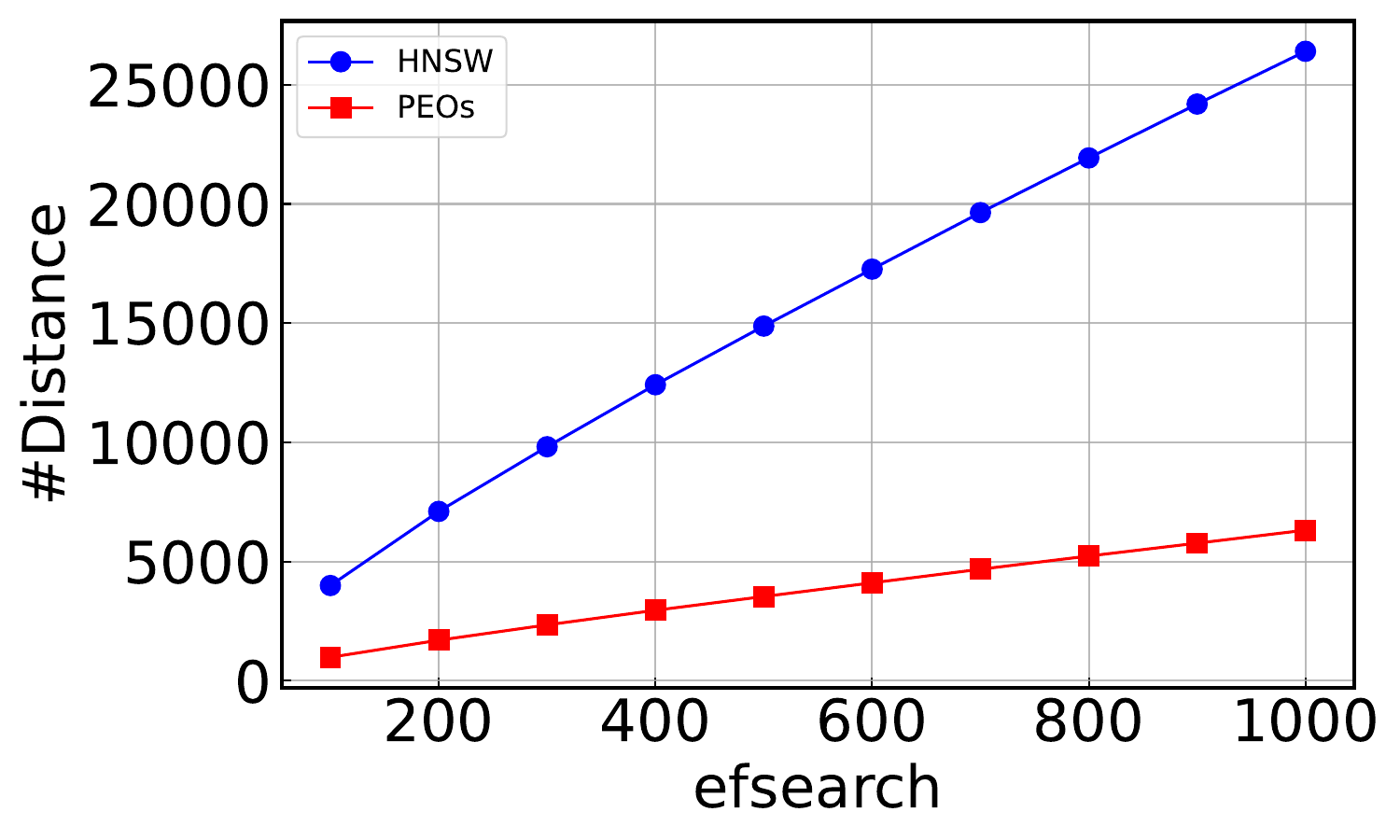}}
  \subfloat[DEEP10M-angular, \#Dist. Calc.]{\includegraphics[width=.7\columnwidth]{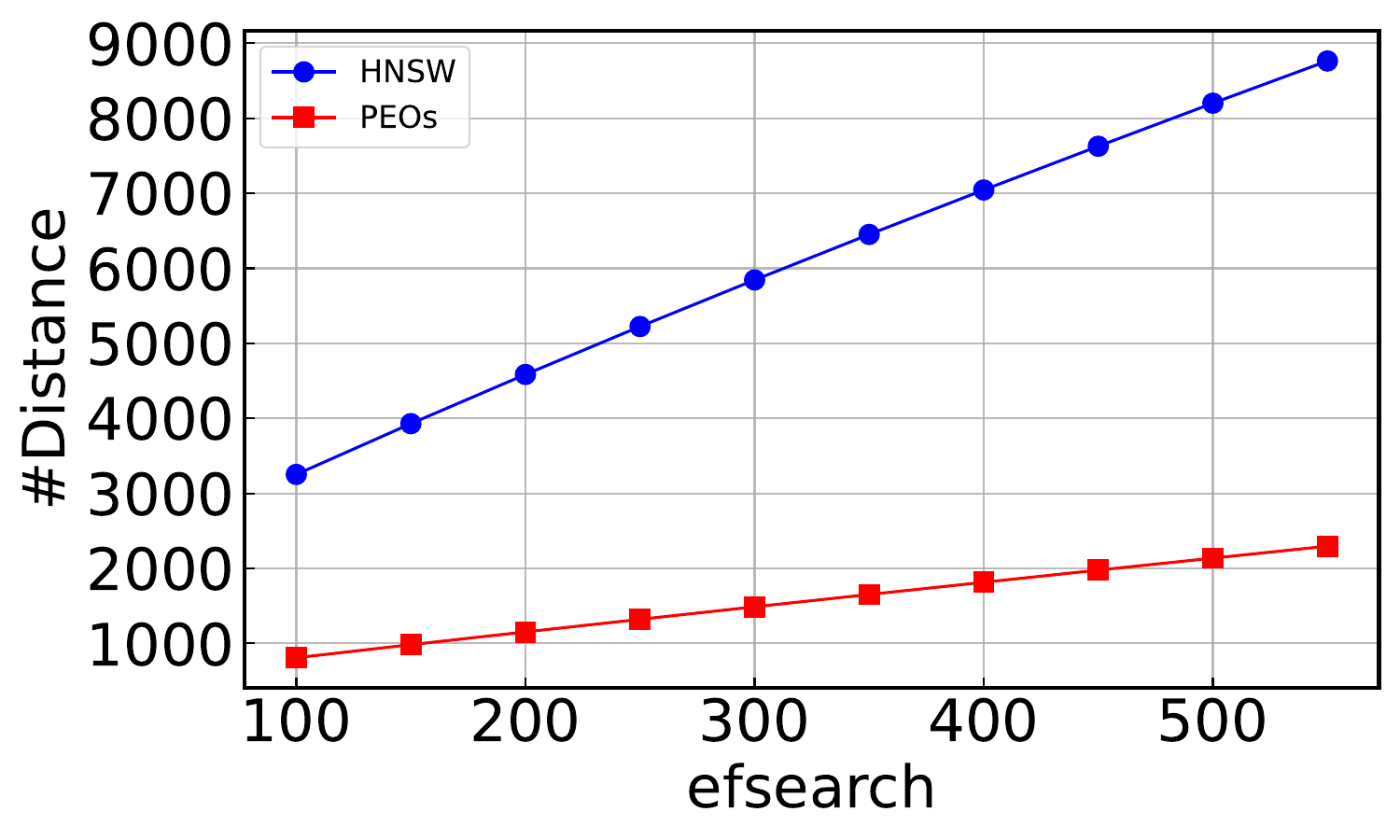}}

  \subfloat[GloVe200-angular, Recall]{\includegraphics[width=.7\columnwidth]{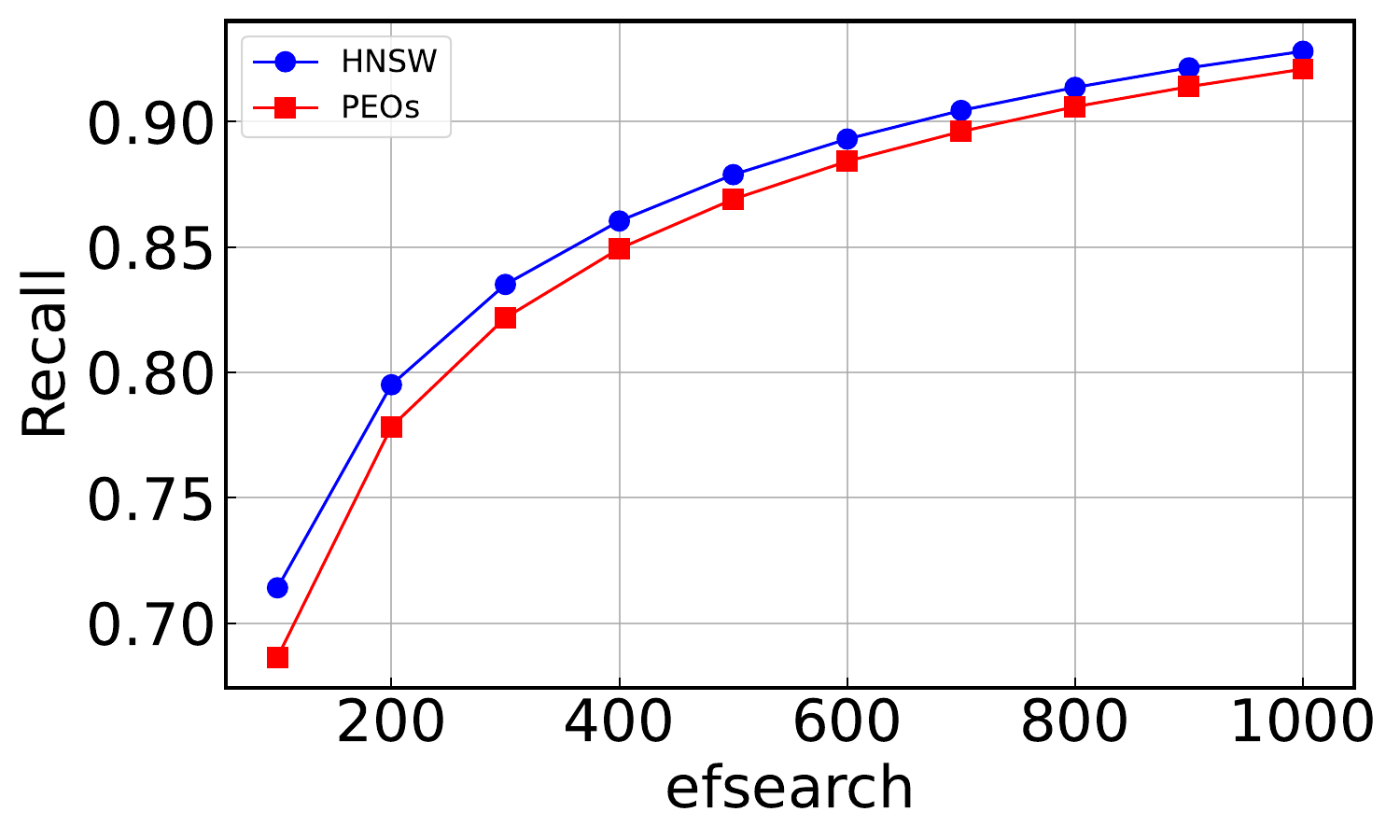}}
  \subfloat[GloVe300-$\ell_2$, Recall]{\includegraphics[width=.7\columnwidth]{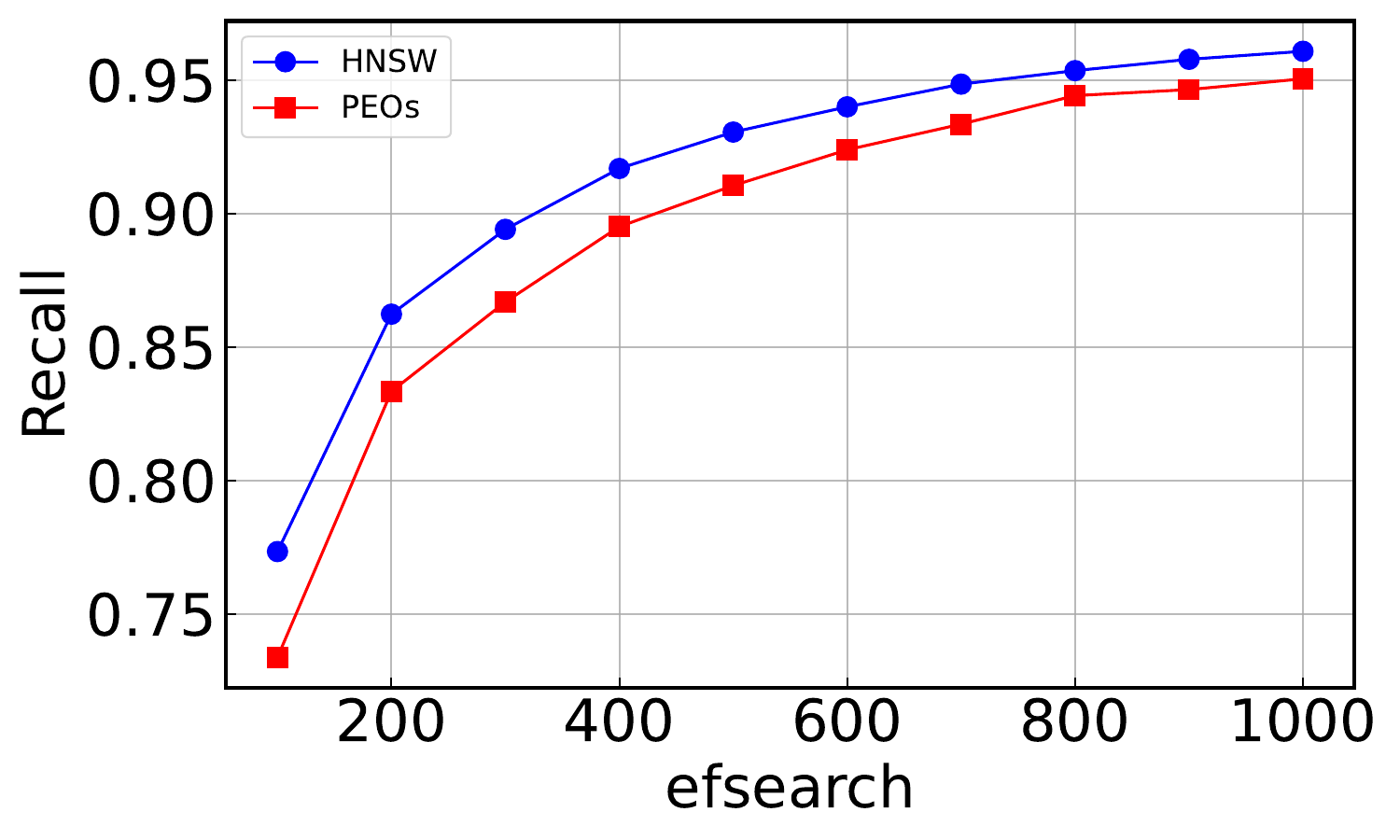}}
  \subfloat[DEEP10M-angular, Recall]{\includegraphics[width=.7\columnwidth]{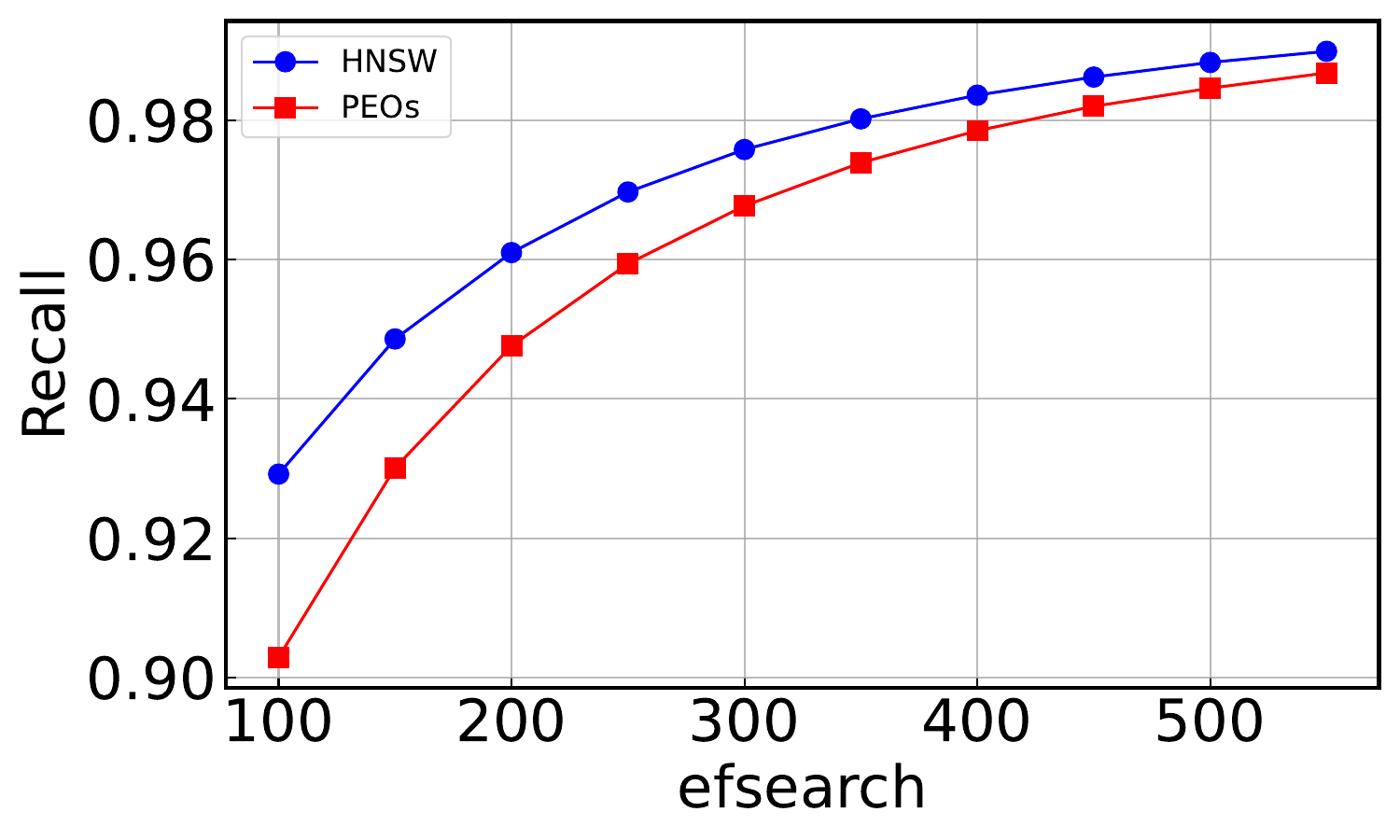}}

  \subfloat[SIFT10M-$\ell_2$, \#Dist. Calc.]{\includegraphics[width=.7\columnwidth]{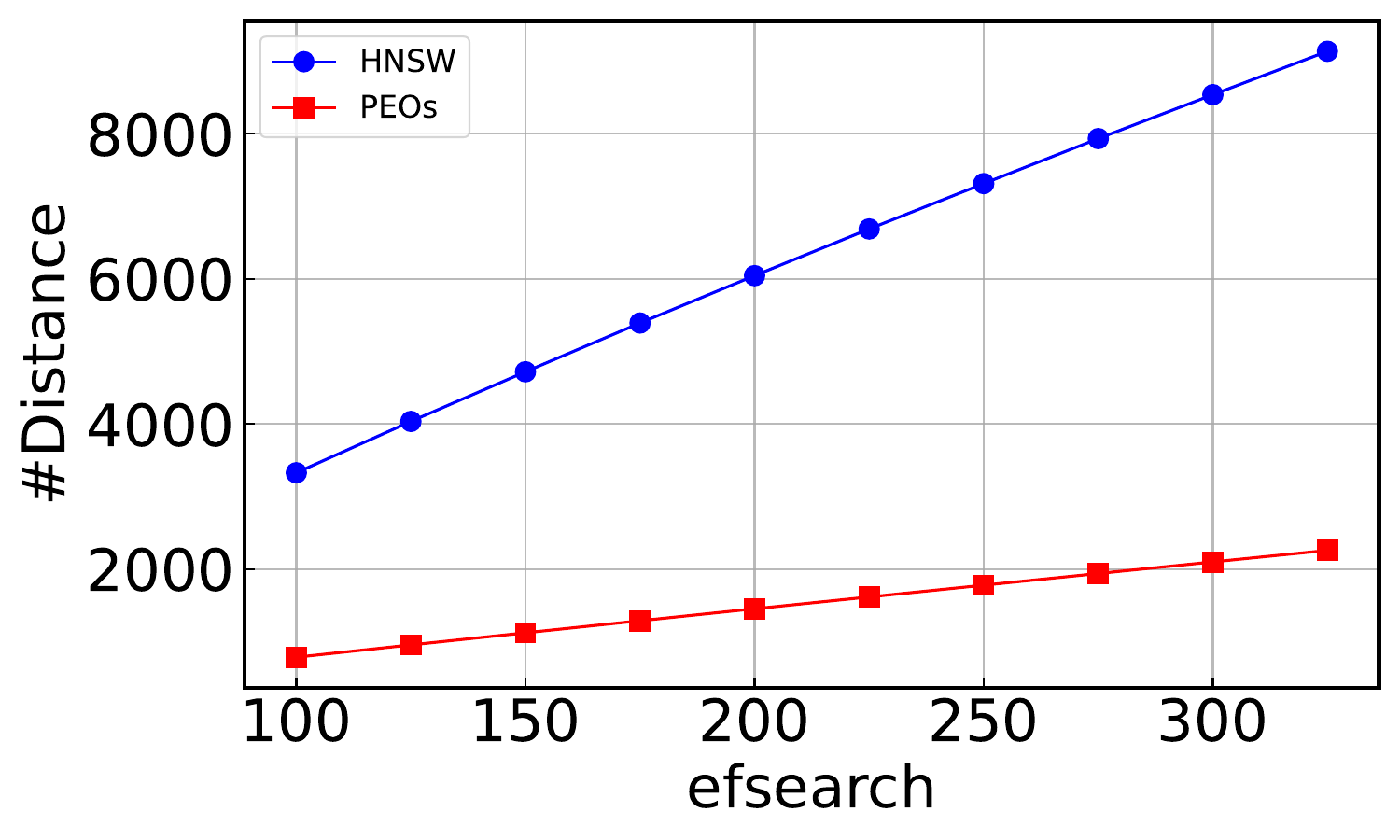}}
  \subfloat[Tiny5M-$\ell_2$, \#Dist. Calc.]{\includegraphics[width=.7\columnwidth]{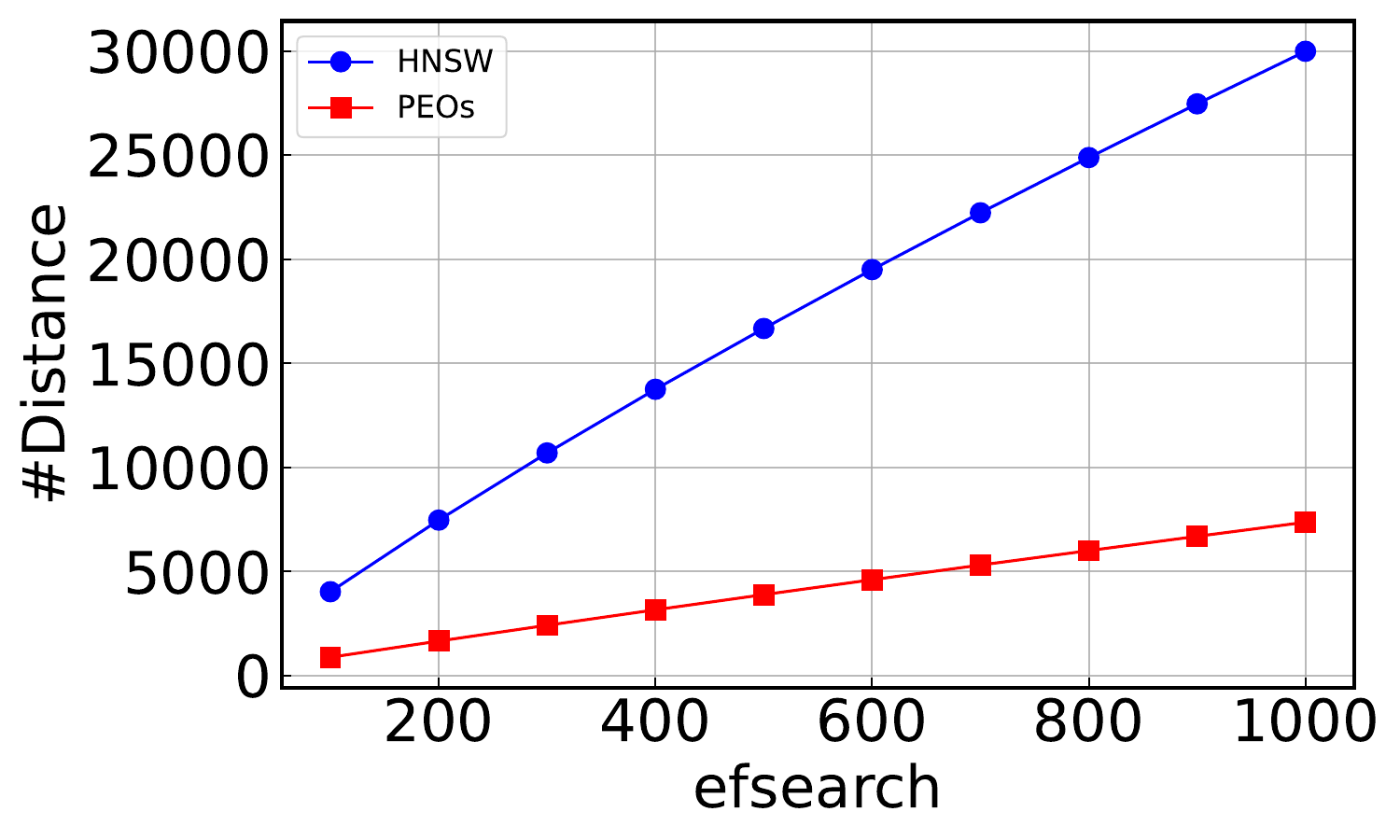}}
  \subfloat[GIST-$\ell_2$, \#Dist. Calc.]{\includegraphics[width=.7\columnwidth]{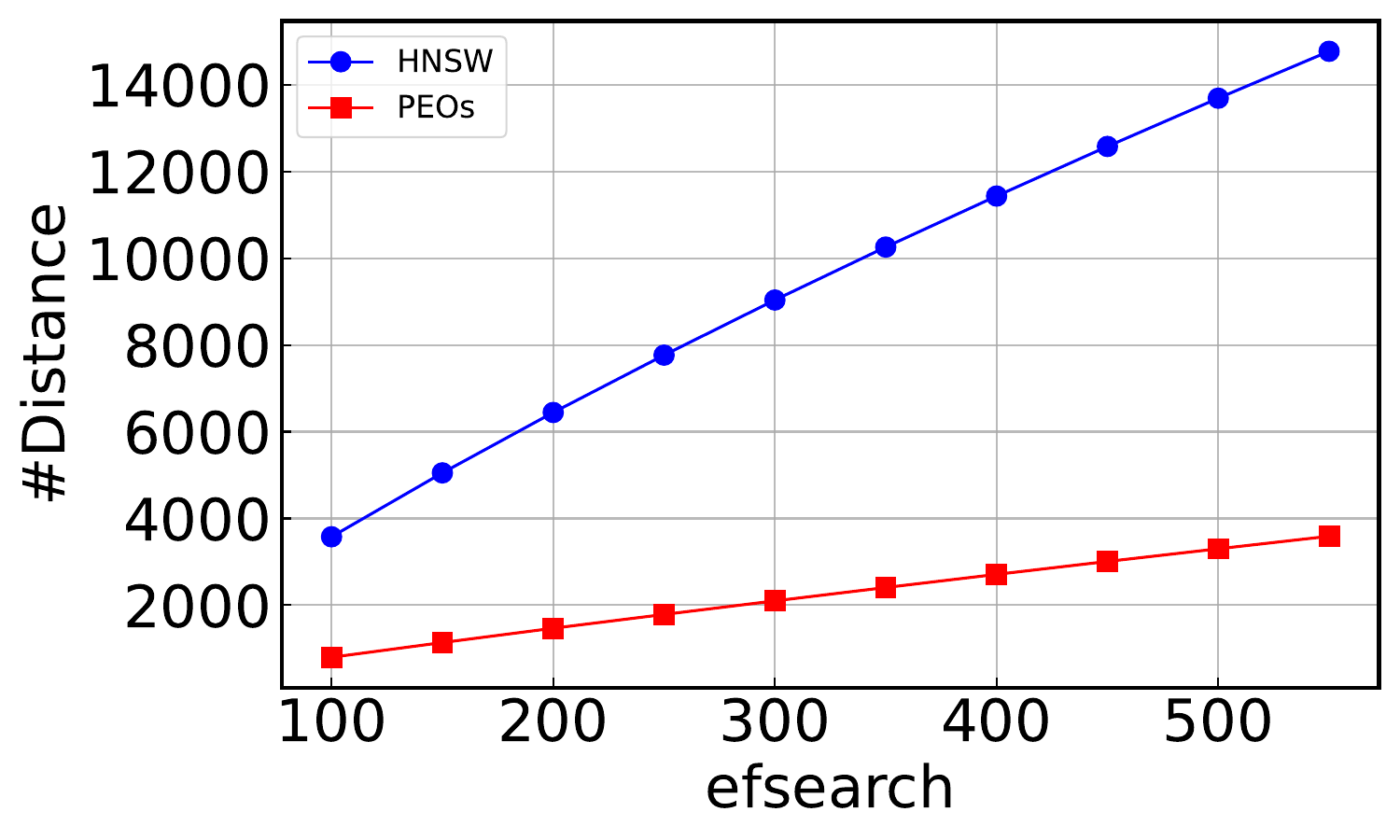}}

  \subfloat[SIFT10M-$\ell_2$, Recall]{\includegraphics[width=.7\columnwidth]{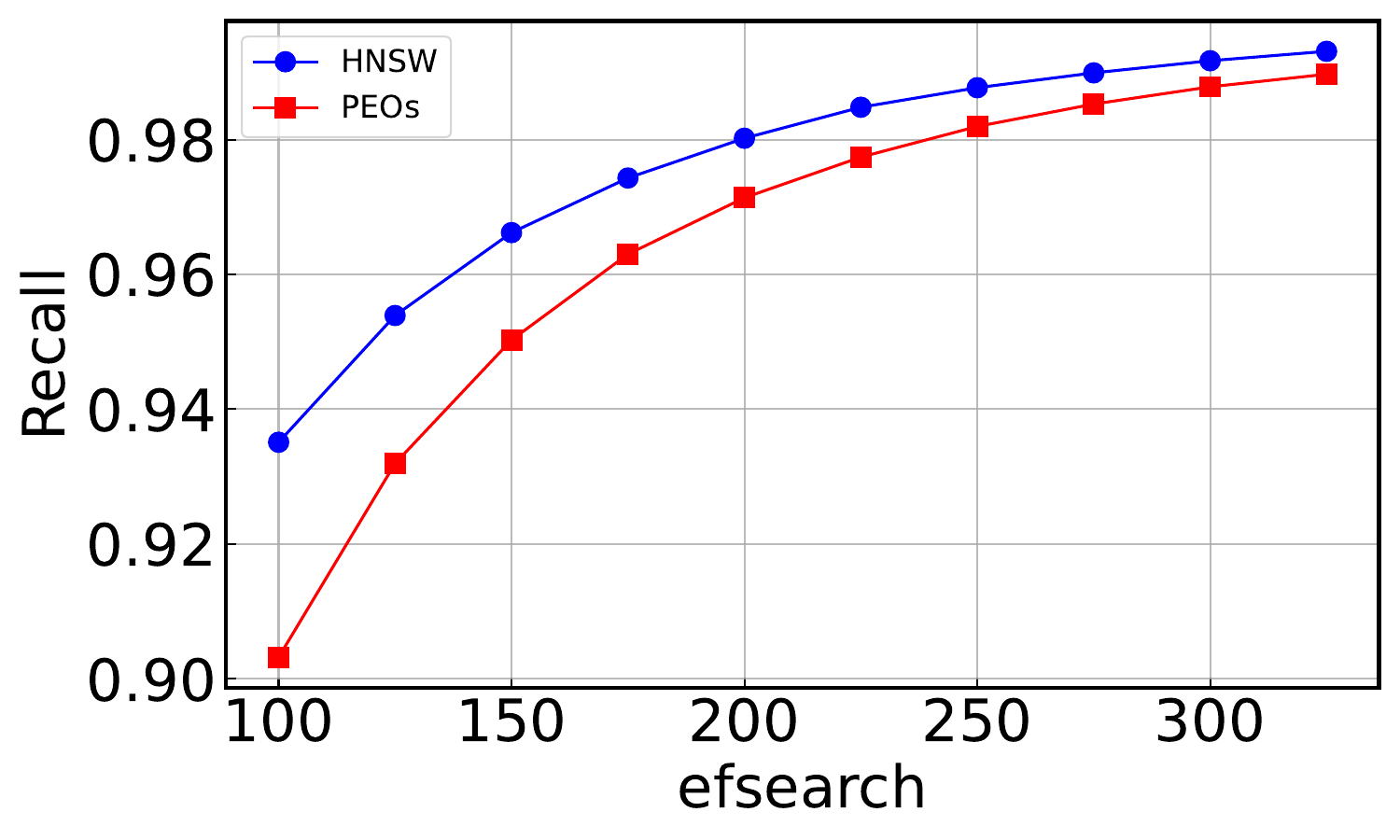}}
  \subfloat[Tiny5M-$\ell_2$, Recall]{\includegraphics[width=.7\columnwidth]{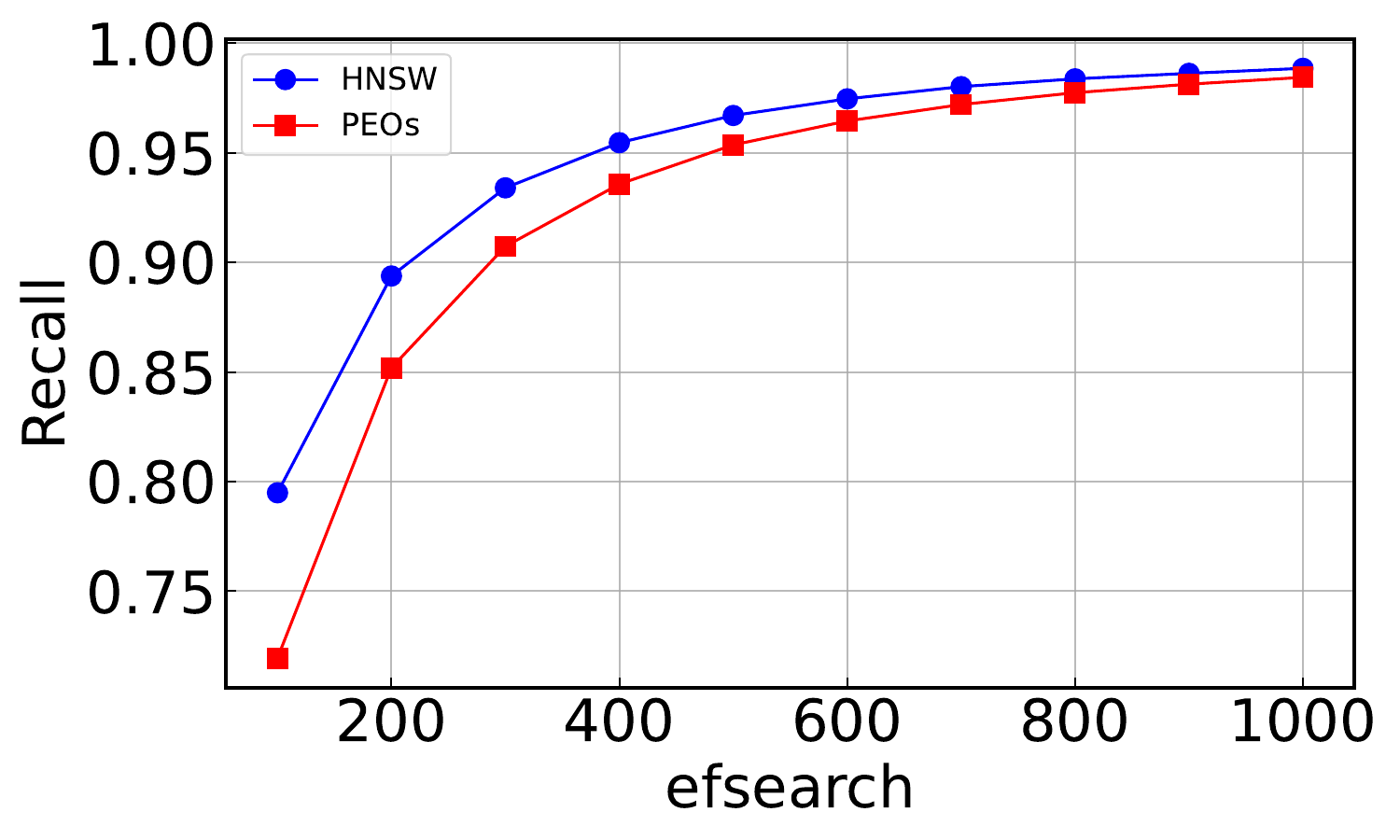}}
  \subfloat[GIST-$\ell_2$, Recall]{\includegraphics[width=.7\columnwidth]{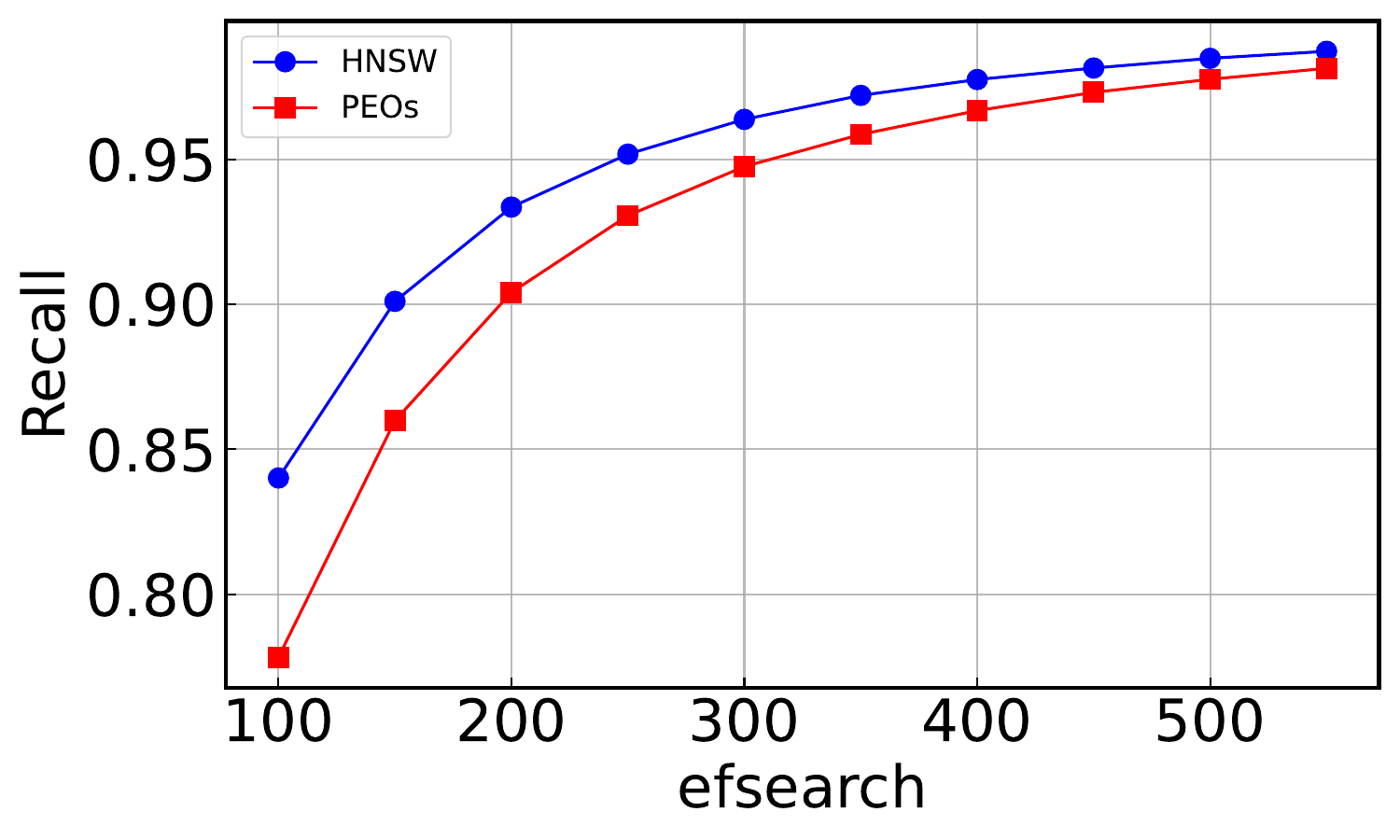}}
 
  \caption{Evaluation of $efs$-number of distance calculations and $efs$-recall, $K = 100$.}
  \label{fig:distance-curves}
\end{figure*}




\begin{figure*}[t]
  \centering
  \subfloat[Dimension: 200 (GloVe200)]{\includegraphics[width=.7\columnwidth]{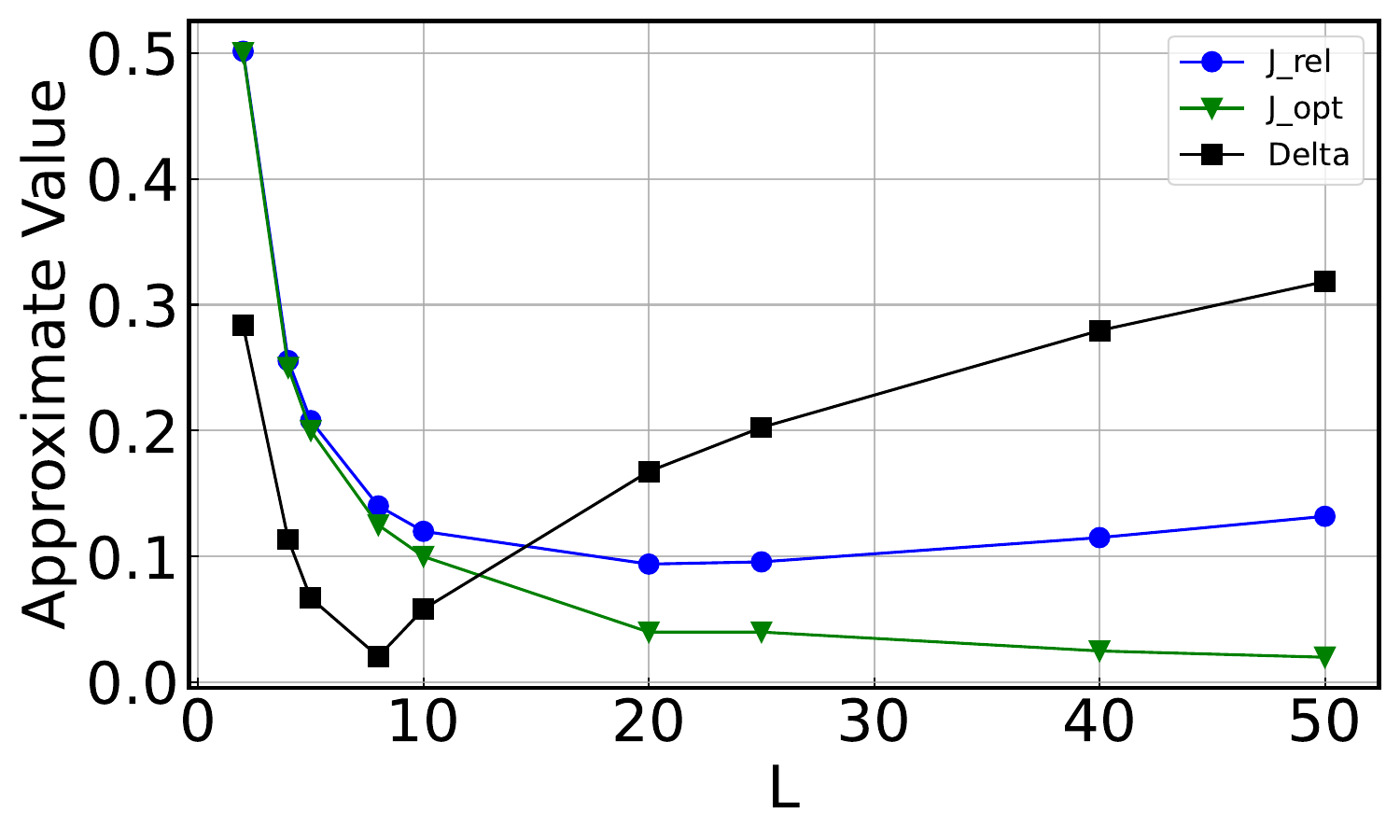}}
  \subfloat[Dimension: 300 (GloVe300)]{\includegraphics[width=.7\columnwidth]{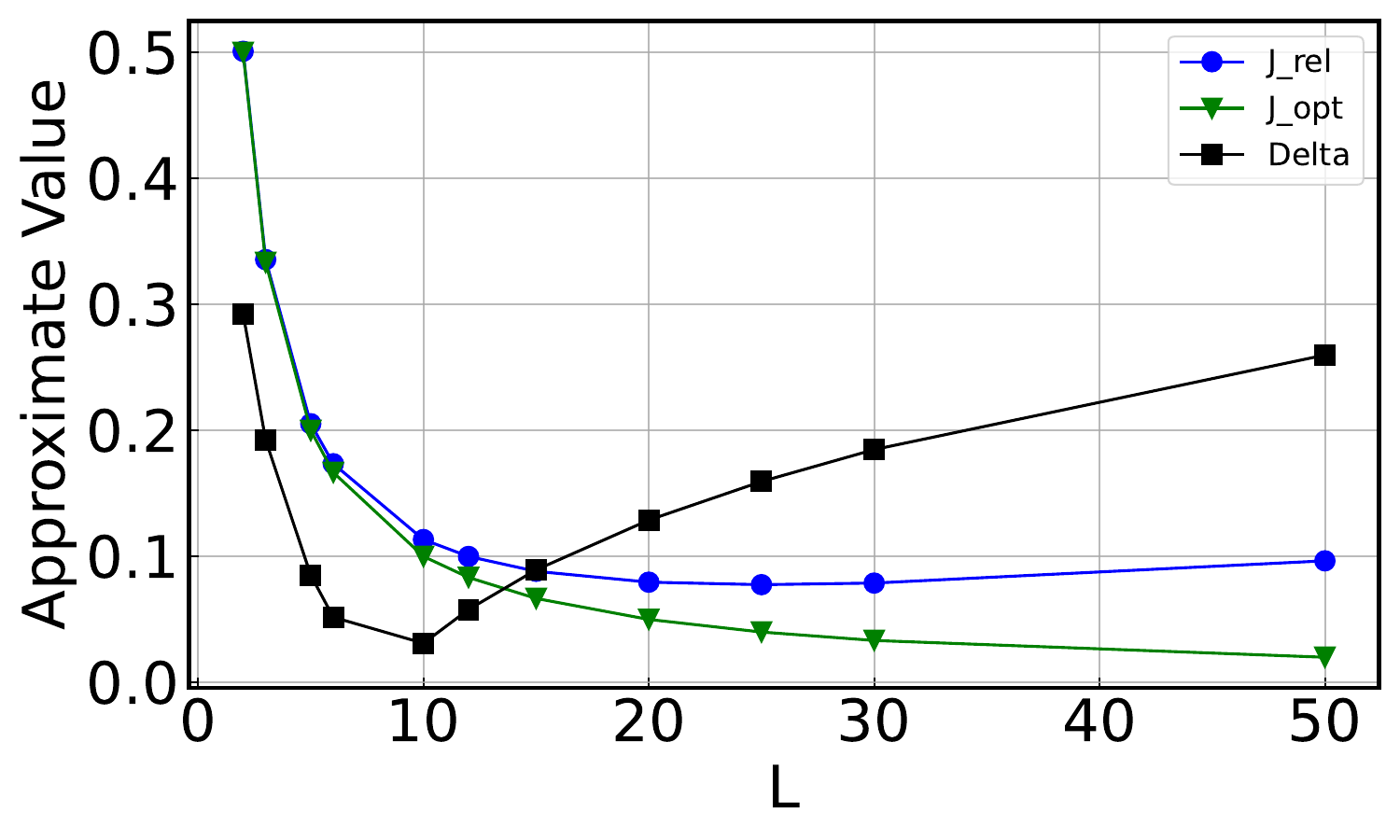}}
  \subfloat[Dimension: 96 (DEEP10M)]{\includegraphics[width=.7\columnwidth]{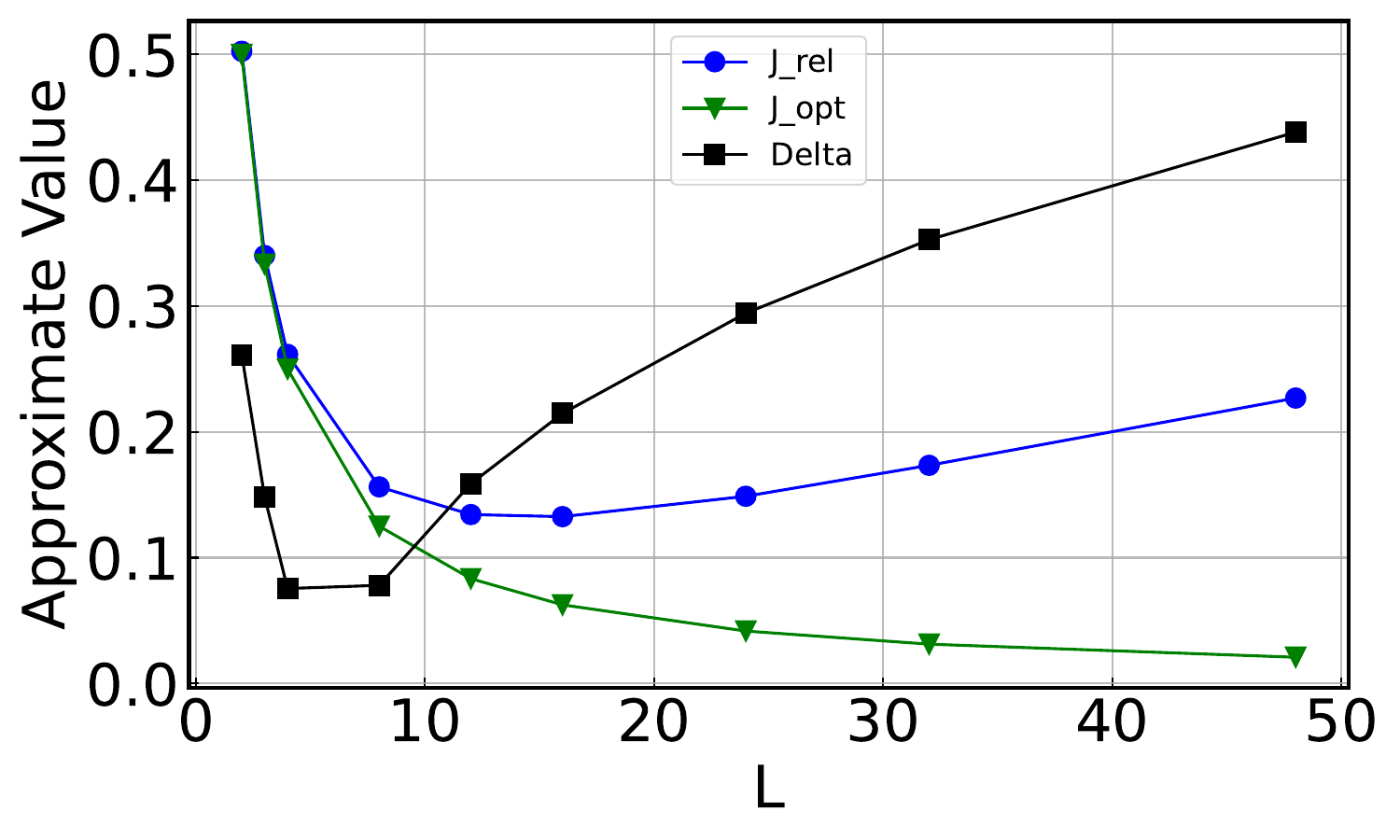}}

  \subfloat[GloVe200-angular, $d=200$]{\includegraphics[width=.7\columnwidth]{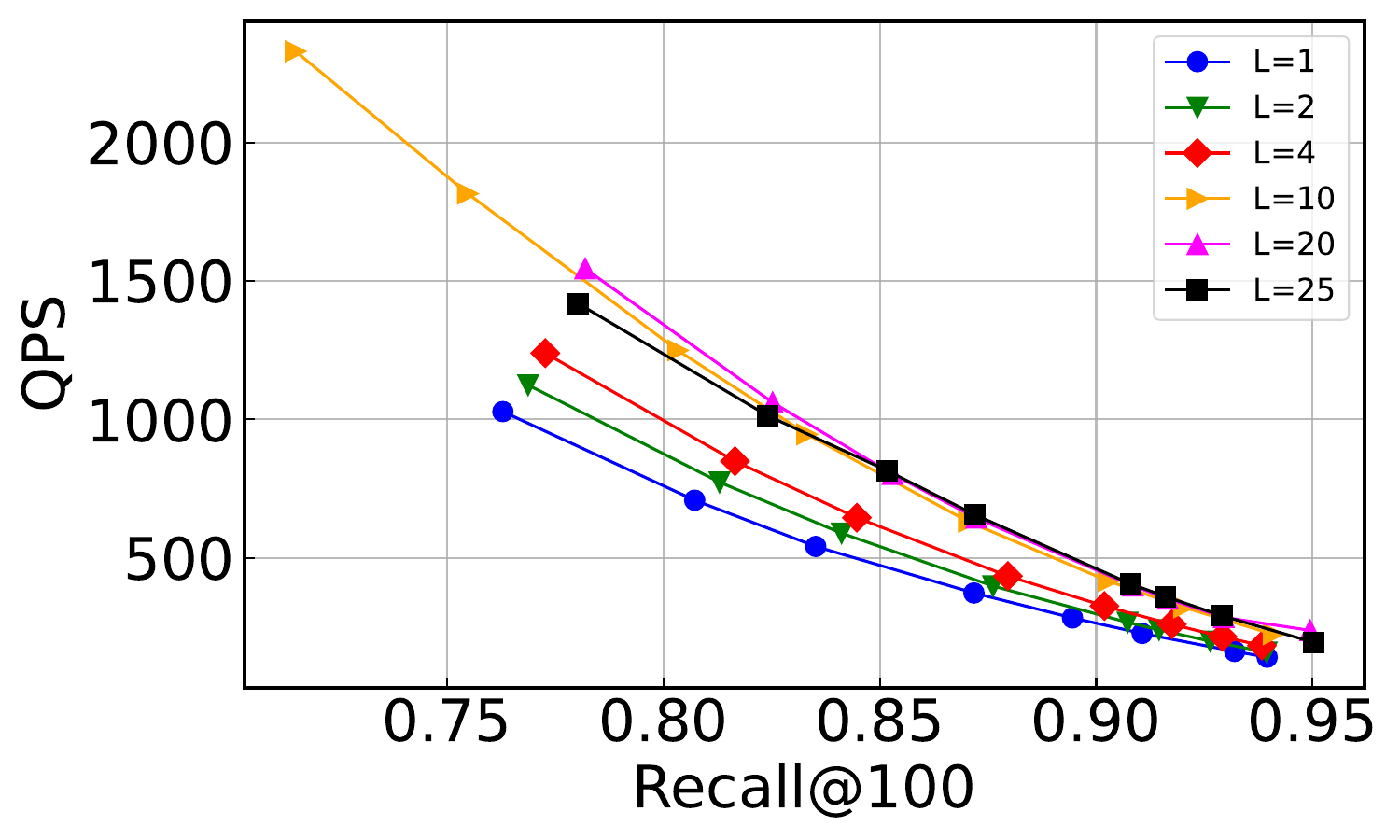}}
  \subfloat[GloVe300-$\ell_2$, $d=300$]{\includegraphics[width=.7\columnwidth]{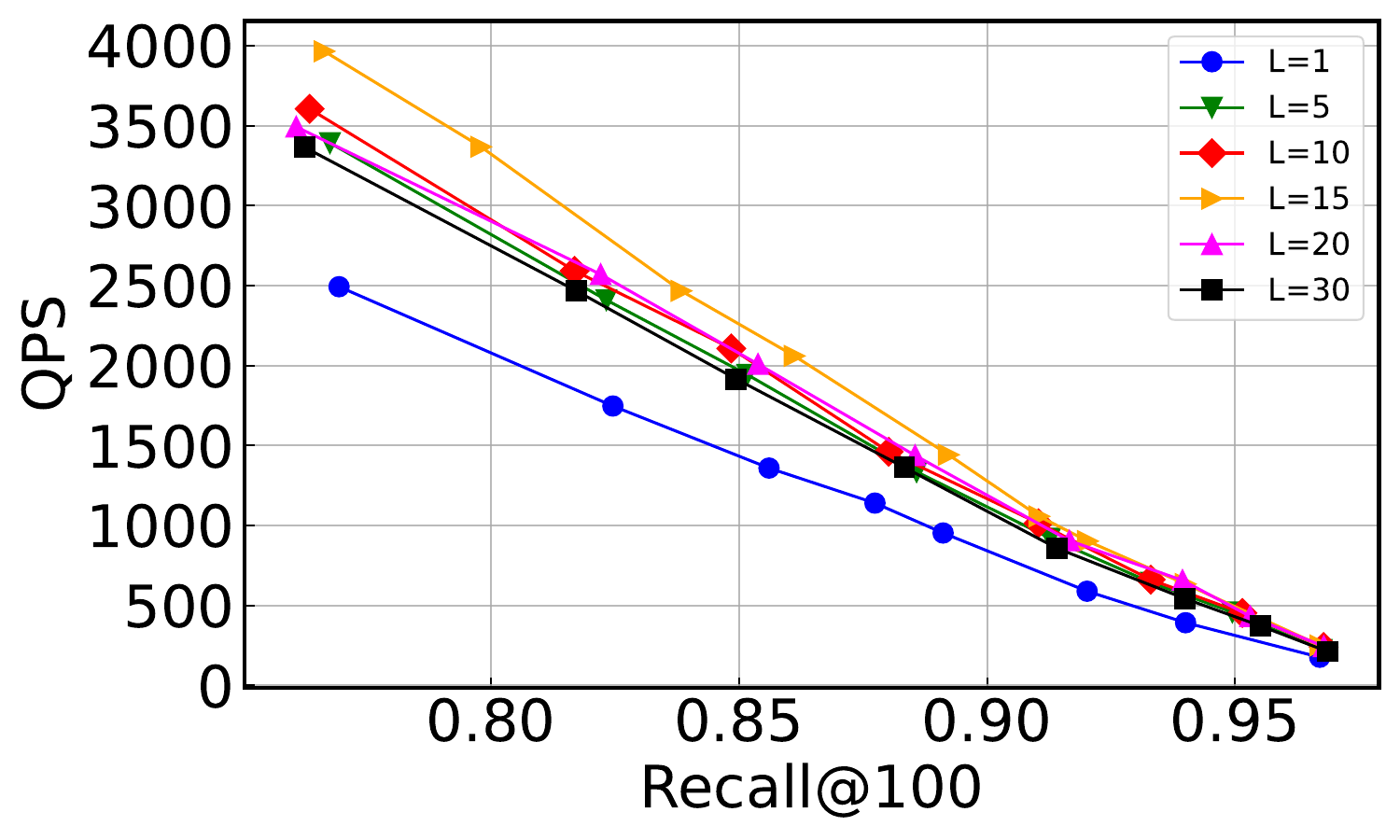}}
  \subfloat[DEEP10M-angular, $d=96$]{\includegraphics[width=.7\columnwidth]{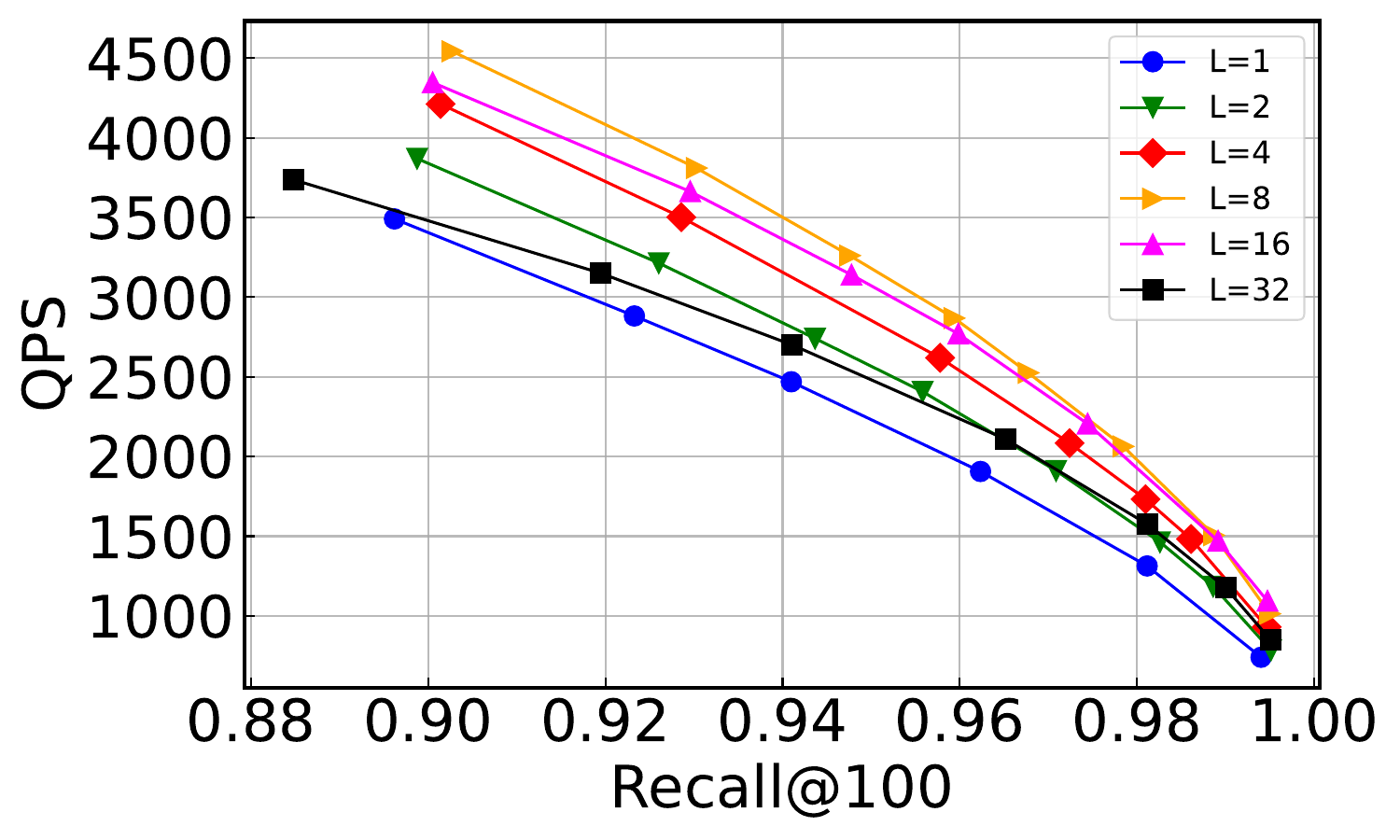}}
  
  \caption{Effect of $L$ on GloVe200, GloVe300, and DEEP10M.}
  \label{fig:L_other_datasets}
\end{figure*}

\begin{figure*}[t]
  \centering
  \subfloat[GloVe200-angular, $K=100$]{\includegraphics[width=.7\columnwidth]{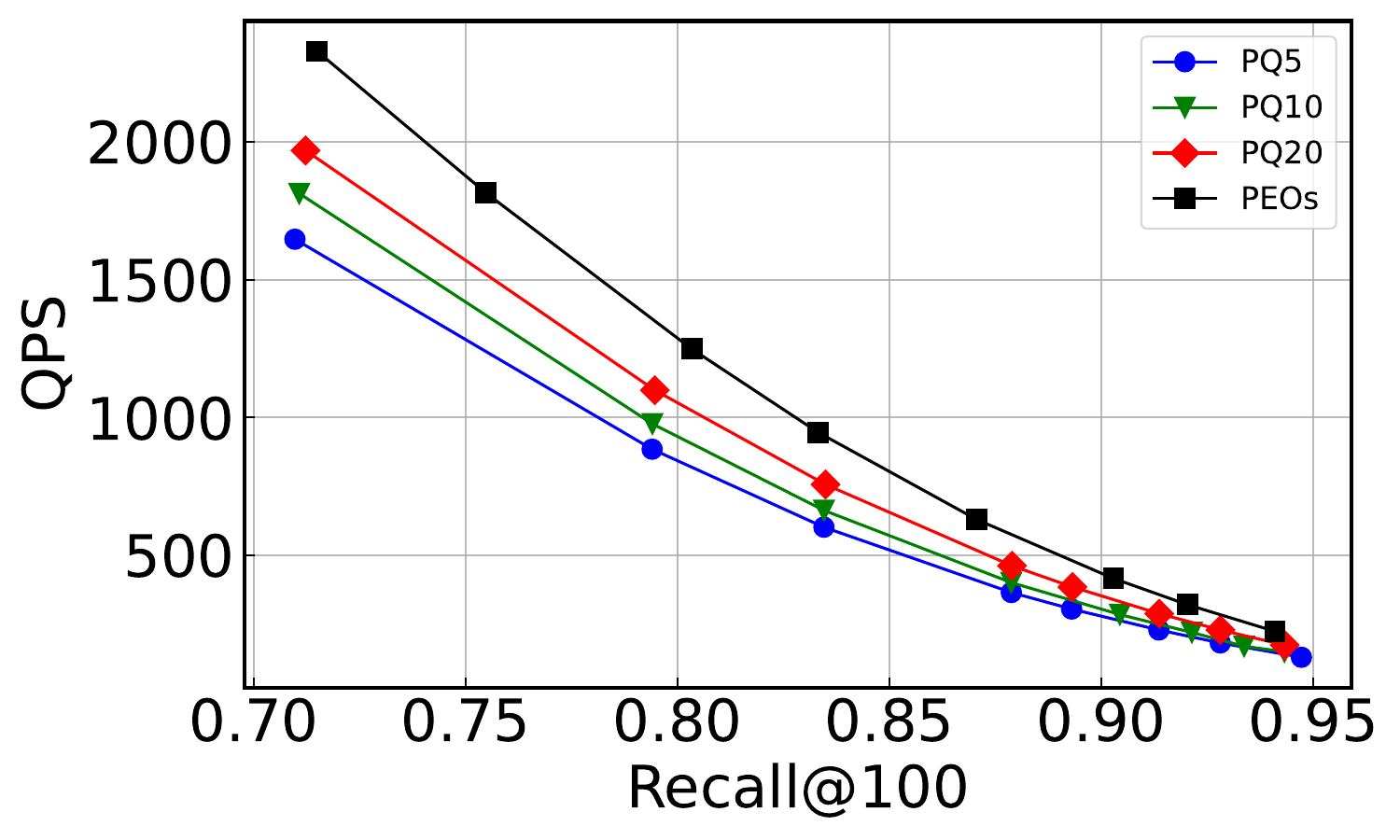}}
  \subfloat[GloVe300-$\ell_2$, $K=100$]{\includegraphics[width=.7\columnwidth]{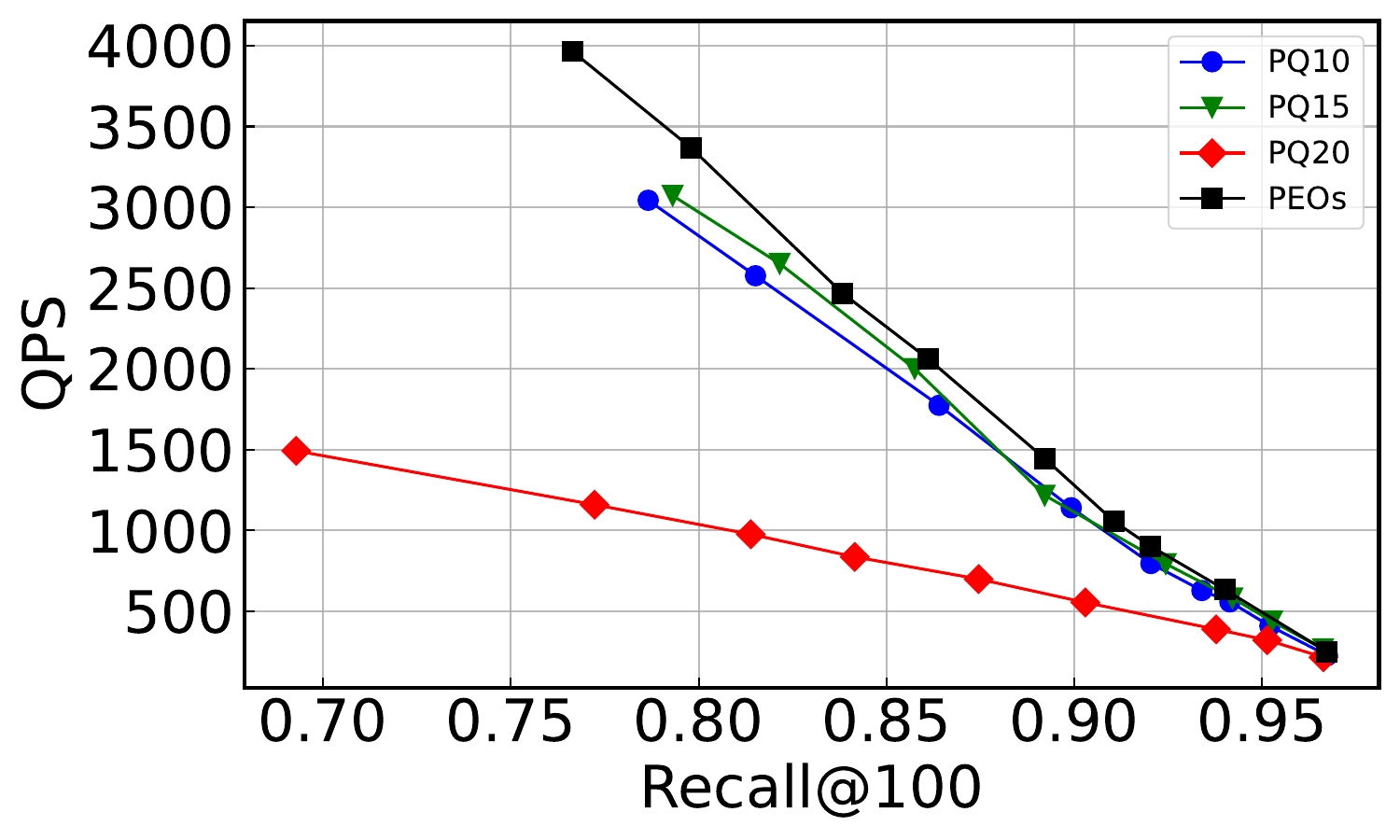}}
  \subfloat[DEEP10M-angular, $K=100$]{\includegraphics[width=.7\columnwidth]{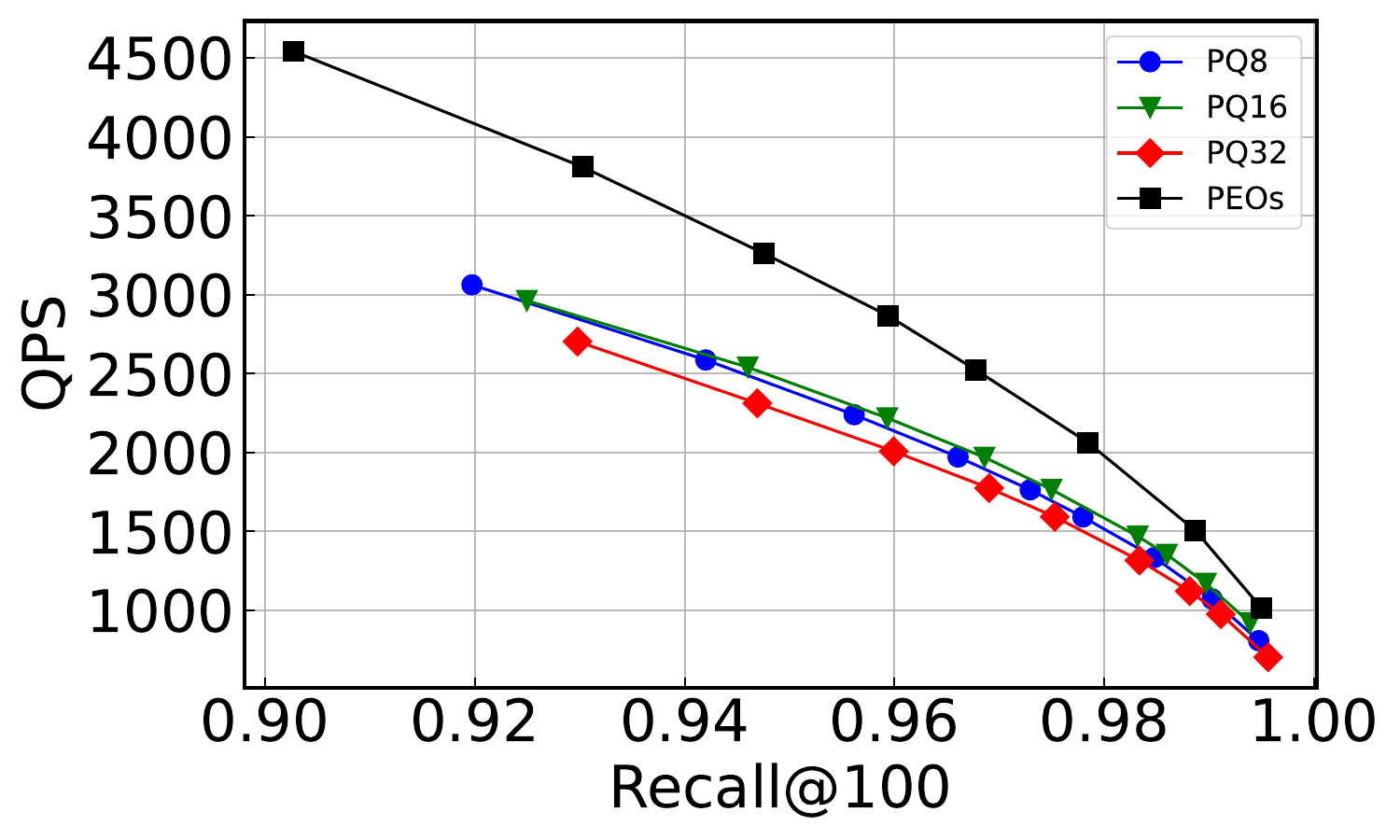}}
  
  \subfloat[SIFT10M-$\ell_2$, $K=100$]{\includegraphics[width=.7\columnwidth]{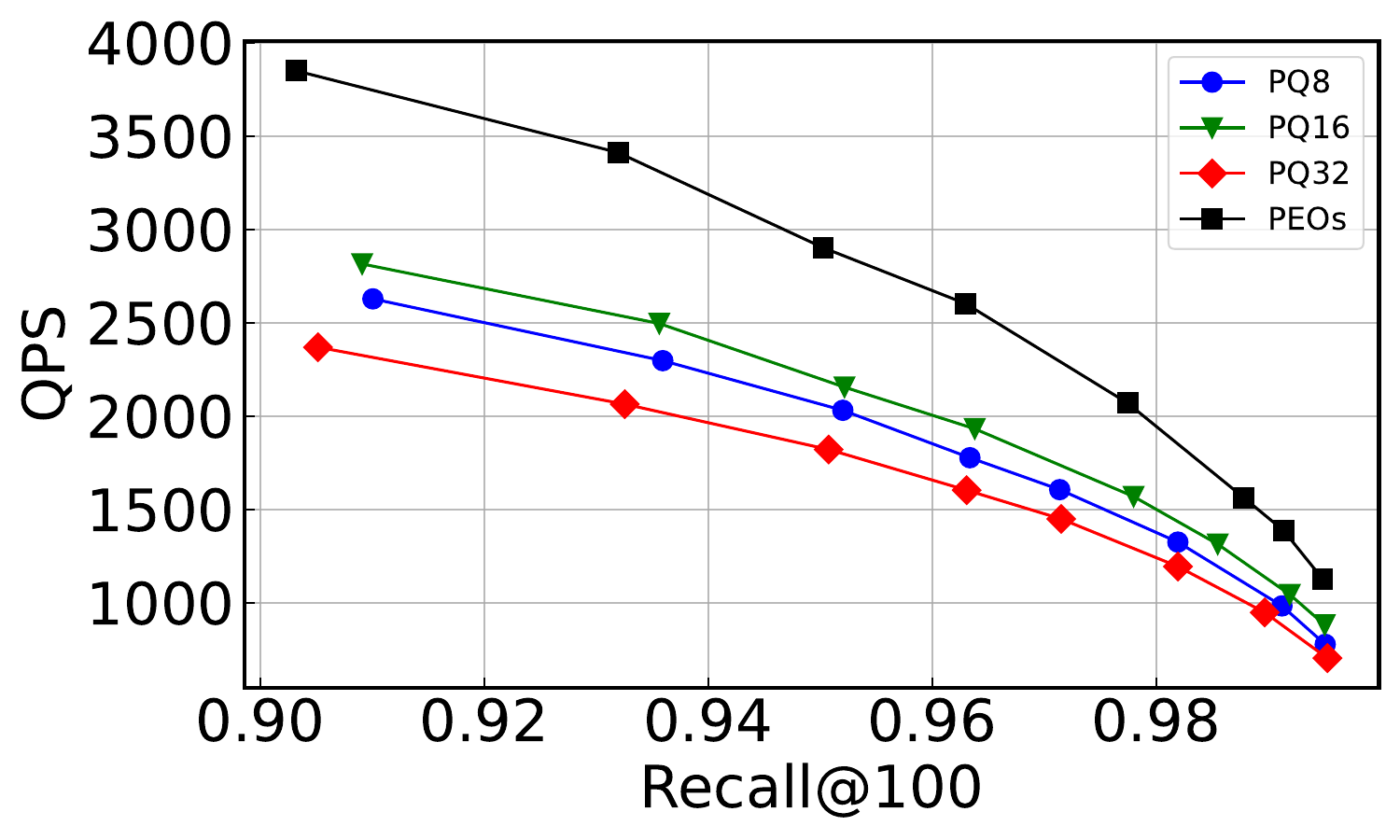}}
  \subfloat[Tiny5M-$\ell_2$, $K=100$]{\includegraphics[width=.7\columnwidth]{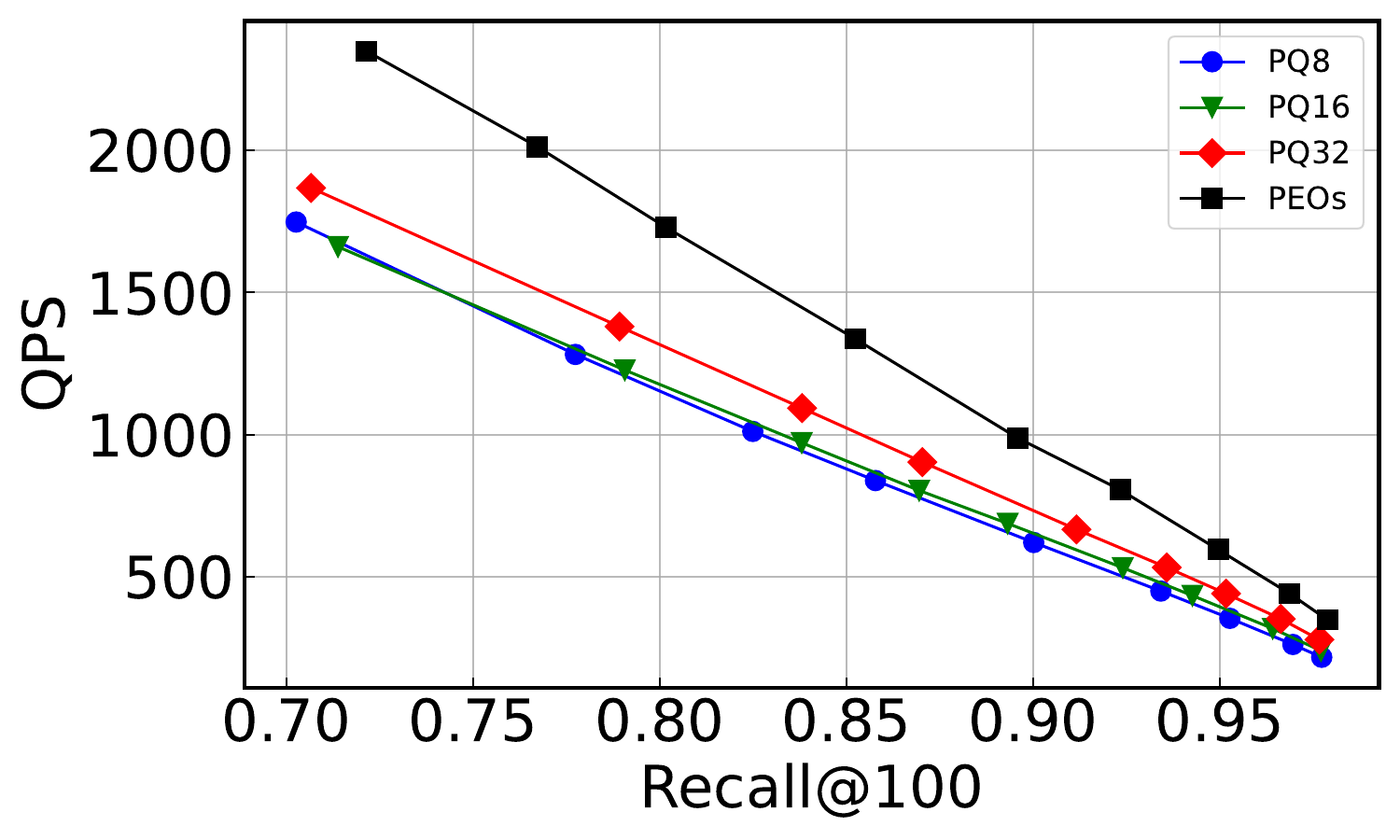}}
  \subfloat[GIST-$\ell_2$, $K=100$]{\includegraphics[width=.7\columnwidth]{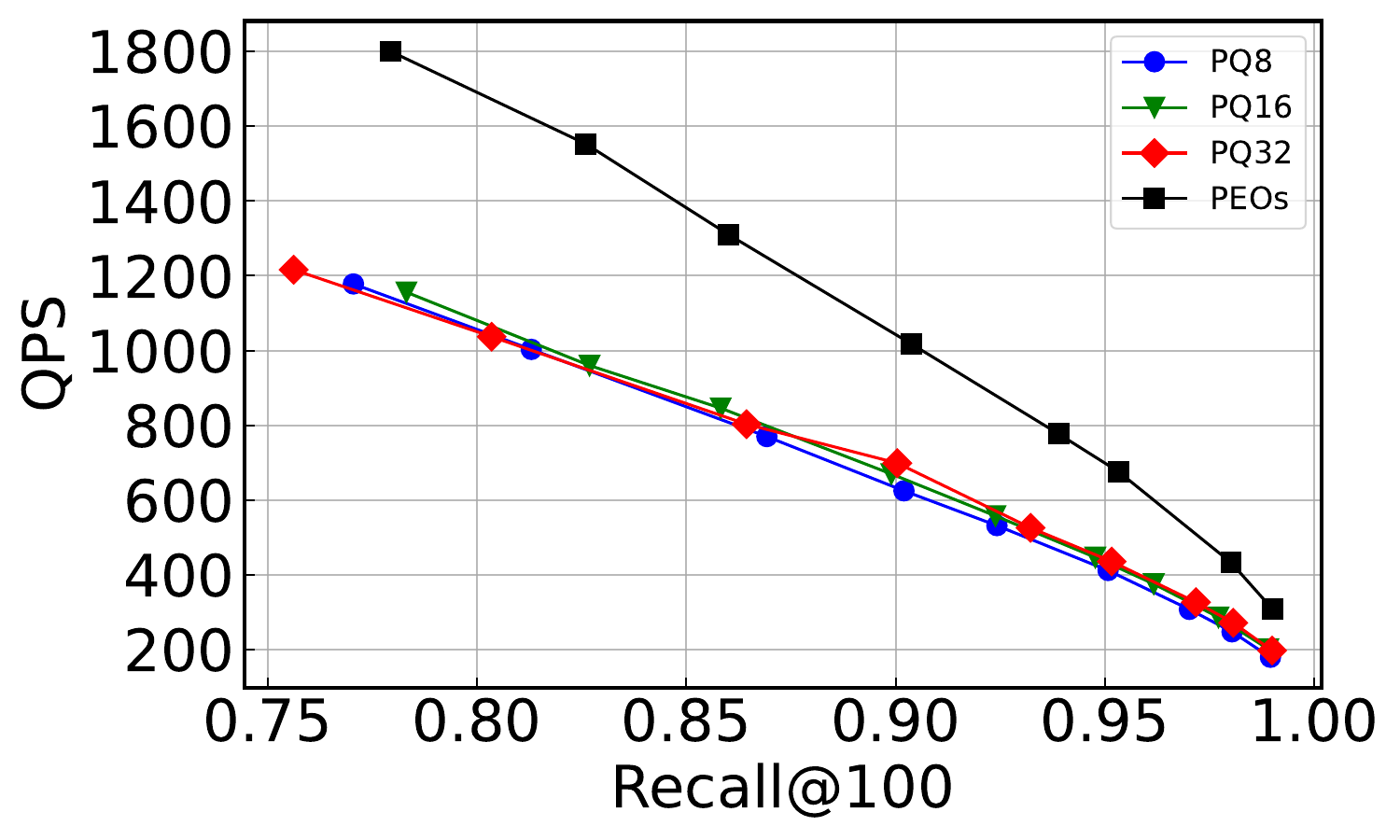}}
  \caption{Comparison with PQ-based routing. X in PQX denotes the number of sub-codebooks.}
  \label{fig:comparison-pq}
\end{figure*}

\section{Experimental Setup}
\label{sec:exp-setup}
All the experiments were performed on a PC with Intel(R) Xeon(R) Gold 6258R CPU @ 2.70GHz. All the compared methods were implemented in C++, with 64 threads for indexing and a single CPU for searching, following the standard setup in ANN-Benchmarks~\cite{ann-benchmarks}. 

We use the following parameter setup for the competitors: 

(1) \textbf{HNSW.} $M=32$, $efc=2000$ for the two GloVe datasets and $efc=1000$ for the other datasets. We also use this setting for the HNSW index of FINGER, RCEOs and PEOs.

(2) \textbf{NSSG.} ($L$, $R$, $C$) is set to (100, 50, 60) on SIFT10M, (500, 70, 60) on GIST, and (500, 60, 60) on the other datasets, The parameter $K$ in the prepared KNN-graph is 400.

(3) \textbf{Glass.} For DEEP10M, we use Glass (+HNSW) since Glass (+NSG) failed to finish the index construction. For the other datasets, we use Glass (+NSG) since it works better than Glass (+HNSW) especially for high recall rates. $R$ is 32, $L$ is set to an experimentally optimal value in $[200, 2000]$.

(4) \textbf{FINGER (+HNSW).} All the parameters are set to the recommended values in its source code. In particular, the dimension of subspace is set to 64.

(5) \textbf{RCEOs (+HNSW).} The only difference with PEOs is $L = 1$ in RCEOs. 

(6) \textbf{PEOs (+HNSW).} Based on the analysis in Sec.~\ref{sec:impact-of-L}, $L$ is set to 8, 8, 10, 15, 16, 20 on the six datasets sorted by ascending order of dimension. $\epsilon$ is set to 0.2 and $m = 128$ such that every vector ID can be encoded by one byte. 

(7) \textbf{PEOs (+NSSG).} The settings of $L$ are the same as those in PEOs (+HNSW). The NSSG parameters are set to the same values as in the vanilla NSSG.

(8) \textbf{ScaNN.} Dimensions\_per\_block is set to 2 on SIFT10M, DEEP10M, and GloVe200, and 4 on the other ones. The other user-specified parameters are adjusted to achieve the trade-off curves. 



\section{Additional Experiments}
\label{sec:additional-experiments}

\subsection{Effect of Result Number $K$}
In Figure~\ref{fig:performance-different-k}, we show the recall-QPS comparison under $K=10$ and $K=1$. We have the following observations.

(1) PEOs still performs the best on all the datasets, showcasing the robustness of PEOs for different values of $K$. In particular, the performance improvement of PEOs over HNSW under a small $K$ is almost consistent with that under $K=100$. 
(2) The improvement of PEOs over FINGER is marginal when $K=1$. This is because the search under $K=1$ is much easier than the search under a large $K$ value. When $K$ grows, we have to accordingly increase the size of the result list, under which situation, the routing becomes harder and a more accurate estimation is important for the performance improvement.

\subsection{Effect of List Size $efs$}
\label{sec:distance-reall-curves}
To evaluate the effectiveness of PEOs in reducing exact distance calculations, we also plot the $efs$-number of distance calculations and $efs$-recall curves, where $efs$ is the size of the temporary result list, adjusted to achieve a different trade-off between efficiency and accuracy. From the results in Figure~\ref{fig:distance-curves}, we have the following observations. 

(1) PEOs saves around 75\% exact distance calculations on each dataset, and this is the main reason for the improvement on QPS. On the other hand, such improvement is not sensitive to the value of $efs$. 
(2) Due to the existence of an estimation error, for the same $efs$, the recall of PEOs is smaller than that of HNSW. Nonetheless, the difference is quite small, especially when $efs > 100$ (note that $efs \ge 100$ when $K=100$), thanks to the theoretical guarantee of PEOs.
(3) As $efs$ grows, the difference between the recalls of HNSW and PEOs are very small while the saved distance calculations by PEOs are still large. This explains why PEOs works better for larger $efs$ values, which generally corresponds to larger $K$.




\subsection{Effect of Space Partition Size $L$ on Other Datasets}
\label{sec:effect-of-L-others}
In addition to the evaluation in Sec.~\ref{exp:effect-of-L}, we show the effect of $L$ on the other three real datasets in Figure~\ref{fig:L_other_datasets}. We can see that the performances on these three real datasets are still consistent with our theoretical analysis and the advantage of $L > 1$ over $L = 1$ remains. 

\subsection{Comparison with Product Quantization (PQ)}
\label{sec:quantization}

We can see that PEOs has similarities with PQ~\cite{PQ}, both partitioning the original space into orthogonal subspaces and combining the information in different subspaces. Thus, an interesting question is if the quantization techniques can be used for the acceleration of routing. First, we note that, since $\size{E}$ is much larger than the data size $\size{\mathcal{O}}$, and the local intrinsic dimensionality (LID) of $\bm{e}$'s is also very large due to the edge-selection strategy, the quantization of $\bm{e}$'s is not as effective as the quantization of raw vectors. On the other hand, apart from the probability guarantee, PEOs has the following two advantages. (1) The impact of $w_{res}$ is fully considered in PEOs while the impact of individual quantization error is hard to be measured. (2) PEOs applies a non-linear transformation, i.e., $F^{-1}_{\bm{e}, \epsilon}$, to the threshold $\cos \theta$. After such transformation, as threshold $\theta$ decreases from $\pi/2$ to $0$, $T_{r}(\bm{e})$ will grow rapidly such that passing the PEOs test becomes much harder, because the possibility that a small angle $\theta$ exists between two high-dimensional vectors is very small. That is, PEOs takes the potential impact of threshold into consideration thanks to the probability guarantee. On the other hand, quantization-based techniques focus on the minimization of quantization errors and do not consider the impact of the threshold. 

For an empirical evaluation, first, we need to adapt the existing quantization-based technique for routing. Specifically, we use norm-explicit product quantization~\cite{norm-explicit-quantization} -- which has shown competitive performance for the estimation of inner product -- to quantize $\bm{e}$'s and calculate the approximate inner product between $\bm{q}$ and each $\bm{e}$. Since the standard quantization generally does not consider the effect of quantization error in the search phase, to obtain better search performance, we also maintain the norm of the residual part of $\bm{e}$ after quantization, denoted by $\norm{\bm{e}}_{res}$. Then we need to introduce a coefficient $c$ and write the test of quantization as follows:
\begin{equation}
  c \norm{\bm{e}}_{res} \norm{\bm{q}} \ge I
\end{equation}
where $I$ denotes the threshold of inner product for being added into the temporary result list, which is calculated in a similar way to $A_{r}(\bm{e})$. In practice, although we do not know the optimal value of $c$ for each dataset, we experimentally adjust it in $(0, 1)$ to get a near-optimal value. By the above setting, we can compare the performances of HNSW+PEOs and HNSW+PQ. From the results in Figure~\ref{fig:comparison-pq}, we have the following three observations. (1) Expect for GloVe300, PEOs obviously performs better than PQ, as analyzed in the previous discussion. (2) Due to the hardness of residual quantization, the quantization error may be quite large especially for the high-dimensional datasets, such as GIST, which makes the estimation inaccurate. (3) The impact of the number of sub-codebooks is hard to be predicted, as shown on GloVe300.

%% file: main.bbl
\begin{thebibliography}{36}
\providecommand{\natexlab}[1]{#1}
\providecommand{\url}[1]{\texttt{#1}}
\expandafter\ifx\csname urlstyle\endcsname\relax
  \providecommand{\doi}[1]{doi: #1}\else
  \providecommand{\doi}{doi: \begingroup \urlstyle{rm}\Url}\fi

\bibitem[Andoni \& Indyk(2008)Andoni and Indyk]{AndoniI08:near-optimal-hashing}
Andoni, A. and Indyk, P.
\newblock Near-optimal hashing algorithms for approximate nearest neighbor in high dimensions.
\newblock \emph{Commun. ACM}, 51\penalty0 (1):\penalty0 117--122, 2008.

\bibitem[Andoni et~al.(2015)Andoni, Indyk, Laarhoven, Razenshteyn, and Schmidt]{falconn}
Andoni, A., Indyk, P., Laarhoven, T., Razenshteyn, I.~P., and Schmidt, L.
\newblock Practical and optimal {LSH} for angular distance.
\newblock In \emph{NeurIPS}, pp.\  1225--1233, 2015.

\bibitem[Babenko \& Lempitsky(2016)Babenko and Lempitsky]{imi}
Babenko, A. and Lempitsky, V.~S.
\newblock Efficient indexing of billion-scale datasets of deep descriptors.
\newblock In \emph{{CVPR}}, pp.\  2055--2063, 2016.

\bibitem[Baranchuk et~al.(2019)Baranchuk, Persiyanov, Sinitsin, and Babenko]{learn-to-route}
Baranchuk, D., Persiyanov, D., Sinitsin, A., and Babenko, A.
\newblock Learning to route in similarity graphs.
\newblock In \emph{{ICML}}, pp.\  475--484, 2019.

\bibitem[Bernhardsson(2024)]{ann-benchmarks}
Bernhardsson, E.
\newblock Ann benchmarks.
\newblock \url{https://github.com/erikbern/ann-benchmarks/}, 2024.

\bibitem[Charikar(2002)]{simhash}
Charikar, M.
\newblock Similarity estimation techniques from rounding algorithms.
\newblock In Reif, J.~H. (ed.), \emph{{STOC}}, pp.\  380--388. {ACM}, 2002.

\bibitem[Chen et~al.(2023)Chen, Chang, Jiang, Yu, Dhillon, and Hsieh]{FINGER}
Chen, P.~H., Chang, W., Jiang, J., Yu, H., Dhillon, I.~S., and Hsieh, C.
\newblock {FINGER:} fast inference for graph-based approximate nearest neighbor search.
\newblock In \emph{{WWW}}, pp.\  3225--3235. {ACM}, 2023.

\bibitem[Curtin et~al.(2014)Curtin, Ram, and Gray]{cover_tree}
Curtin, R.~R., Ram, P., and Gray, A.~G.
\newblock Fast exact max-kernel search.
\newblock \emph{Statistical Analysis and Data Mining}, 7\penalty0 (1):\penalty0 1--9, February--December 2014.

\bibitem[Dai et~al.(2020)Dai, Yan, Ng, Liu, and Cheng]{norm-explicit-quantization}
Dai, X., Yan, X., Ng, K. K.~W., Liu, J., and Cheng, J.
\newblock Norm-explicit quantization: Improving vector quantization for maximum inner product search.
\newblock In \emph{{AAAI}}, pp.\  51--58, 2020.

\bibitem[Datar et~al.(2004)Datar, Immorlica, Indyk, and Mirrokni]{DatarIIM04:p-stable}
Datar, M., Immorlica, N., Indyk, P., and Mirrokni, V.~S.
\newblock Locality-sensitive hashing scheme based on p-stable distributions.
\newblock In \emph{SoCG}, pp.\  253--262, 2004.

\bibitem[Dong et~al.(2011)Dong, Moses, and Li]{dong2011efficient}
Dong, W., Moses, C., and Li, K.
\newblock Efficient k-nearest neighbor graph construction for generic similarity measures.
\newblock In \emph{{WWW}}, pp.\  577--586, 2011.

\bibitem[Douze et~al.(2024)Douze, Guzhva, Deng, Johnson, Szilvasy, Mazar{\'e}, Lomeli, Hosseini, and J{\'e}gou]{douze2024faiss}
Douze, M., Guzhva, A., Deng, C., Johnson, J., Szilvasy, G., Mazar{\'e}, P.-E., Lomeli, M., Hosseini, L., and J{\'e}gou, H.
\newblock The faiss library.
\newblock \emph{arXiv preprint arXiv:2401.08281}, 2024.

\bibitem[Fu et~al.(2019)Fu, Xiang, Wang, and Cai]{nsg}
Fu, C., Xiang, C., Wang, C., and Cai, D.
\newblock Fast approximate nearest neighbor search with the navigating spreading-out graph.
\newblock \emph{{PVLDB}}, 12\penalty0 (5):\penalty0 461--474, 2019.

\bibitem[Fu et~al.(2022)Fu, Wang, and Cai]{nssg}
Fu, C., Wang, C., and Cai, D.
\newblock High dimensional similarity search with satellite system graph: Efficiency, scalability, and unindexed query compatibility.
\newblock \emph{{IEEE} Trans. Pattern Anal. Mach. Intell.}, 44\penalty0 (8):\penalty0 4139--4150, 2022.

\bibitem[Ge et~al.(2014)Ge, He, Ke, and Sun]{OPQ}
Ge, T., He, K., Ke, Q., and Sun, J.
\newblock Optimized product quantization.
\newblock \emph{IEEE Trans. Pattern Anal. Mach. Intell}, 36\penalty0 (4):\penalty0 744--755, 2014.

\bibitem[Guo et~al.(2020)Guo, Sun, Lindgren, Geng, Simcha, Chern, and Kumar]{scann}
Guo, R., Sun, P., Lindgren, E., Geng, Q., Simcha, D., Chern, F., and Kumar, S.
\newblock Accelerating large-scale inference with anisotropic vector quantization.
\newblock In \emph{{ICML}}, pp.\  3887--3896, 2020.

\bibitem[Gupta et~al.(2022)Gupta, Medini, Shrivastava, and Smola]{Bliss}
Gupta, G., Medini, T., Shrivastava, A., and Smola, A.~J.
\newblock {BLISS:} {A} billion scale index using iterative re-partitioning.
\newblock In \emph{{KDD}}, pp.\  486--495. {ACM}, 2022.

\bibitem[Indyk \& Xu(2023)Indyk and Xu]{worst-case-of-graph}
Indyk, P. and Xu, H.
\newblock Worst-case performance of popular approximate nearest neighbor search implementations: Guarantees and limitations.
\newblock \emph{CoRR}, abs/2310.19126, 2023.

\bibitem[Japan(2023)]{ngtqg}
Japan, Y.
\newblock Neighborhood graph and tree for indexing high-dimensional data.
\newblock \url{https://github.com/yahoojapan/NGT}, 2023.

\bibitem[Jégou et~al.(2011)Jégou, Douze, and Schmid]{PQ}
Jégou, H., Douze, M., and Schmid, C.
\newblock Product quantization for nearest neighbor search.
\newblock \emph{IEEE Trans. Pattern Anal. Mach. Intell}, 33\penalty0 (1):\penalty0 117--128, 2011.

\bibitem[Kusupati et~al.(2022)Kusupati, Bhatt, Rege, Wallingford, Sinha, Ramanujan, Howard-Snyder, Chen, Kakade, Jain, et~al.]{kusupati2022matryoshka}
Kusupati, A., Bhatt, G., Rege, A., Wallingford, M., Sinha, A., Ramanujan, V., Howard-Snyder, W., Chen, K., Kakade, S., Jain, P., et~al.
\newblock Matryoshka representation learning.
\newblock \emph{NeurIPS}, 35:\penalty0 30233--30249, 2022.

\bibitem[Lei et~al.(2019)Lei, Huang, Kankanhalli, and Tung]{lei2019sublinear}
Lei, Y., Huang, Q., Kankanhalli, M., and Tung, A.
\newblock Sublinear time nearest neighbor search over generalized weighted space.
\newblock In \emph{{ICML}}, pp.\  3773--3781, 2019.

\bibitem[Li et~al.(2023)Li, Feng, Lian, Xie, Liu, Ge, and Chen]{BATLearn}
Li, W., Feng, C., Lian, D., Xie, Y., Liu, H., Ge, Y., and Chen, E.
\newblock Learning balanced tree indexes for large-scale vector retrieval.
\newblock In \emph{{KDD}}, pp.\  1353--1362. {ACM}, 2023.

\bibitem[Lu \& Kudo(2020)Lu and Kudo]{R2LSH}
Lu, K. and Kudo, M.
\newblock {R2LSH:} {A} nearest neighbor search scheme based on two-dimensional projected spaces.
\newblock In \emph{{ICDE}}, pp.\  1045--1056, 2020.

\bibitem[Lu et~al.(2018)Lu, Wang, Xiao, and Song]{IPL}
Lu, K., Wang, H., Xiao, Y., and Song, H.
\newblock Why locality sensitive hashing works: {A} practical perspective.
\newblock \emph{Inf. Process. Lett.}, 136:\penalty0 49--58, 2018.

\bibitem[Lu et~al.(2021)Lu, Kudo, Xiao, and Ishikawa]{hvs}
Lu, K., Kudo, M., Xiao, C., and Ishikawa, Y.
\newblock {HVS}: Hierarchical graph structure based on voronoi diagrams for solving approximate nearest neighbor search.
\newblock \emph{{PVLDB}}, 15\penalty0 (2):\penalty0 246--258, 2021.

\bibitem[Malkov \& Yashunin(2020)Malkov and Yashunin]{hnsw}
Malkov, Y.~A. and Yashunin, D.~A.
\newblock Efficient and robust approximate nearest neighbor search using hierarchical navigable small world graphs.
\newblock \emph{IEEE Trans. Pattern Anal. Mach. Intell}, 42\penalty0 (4):\penalty0 824--836, 2020.

\bibitem[Mu{\~{n}}oz et~al.(2019)Mu{\~{n}}oz, Gon{\c{c}}alves, Dias, and da~Silva~Torres]{HCNNG}
Mu{\~{n}}oz, J. A.~V., Gon{\c{c}}alves, M.~A., Dias, Z., and da~Silva~Torres, R.
\newblock Hierarchical clustering-based graphs for large scale approximate nearest neighbor search.
\newblock \emph{Pattern Recognit.}, 96, 2019.

\bibitem[Pham(2021)]{ceos}
Pham, N.
\newblock Simple yet efficient algorithms for maximum inner product search via extreme order statistics.
\newblock In \emph{{KDD}}, pp.\  1339--1347, 2021.

\bibitem[Pham \& Liu(2022)Pham and Liu]{Falconn++}
Pham, N. and Liu, T.
\newblock Falconn++: A locality-sensitive filtering approach for approximate nearest neighbor search.
\newblock In \emph{{NeurIPS}}, pp.\  31186--31198, 2022.

\bibitem[Prokhorenkova \& Shekhovtsov(2020)Prokhorenkova and Shekhovtsov]{practice-to-theory}
Prokhorenkova, L. and Shekhovtsov, A.
\newblock Graph-based nearest neighbor search: From practice to theory.
\newblock In \emph{{ICML}}, pp.\  7803--7813, 2020.

\bibitem[Qin et~al.(2021)Qin, Wang, Xiao, Zhang, and Wang]{qin2021high}
Qin, J., Wang, W., Xiao, C., Zhang, Y., and Wang, Y.
\newblock High-dimensional similarity query processing for data science.
\newblock In \emph{{KDD}}, pp.\  4062--4063, 2021.

\bibitem[Subramanya et~al.(2019)Subramanya, Devvrit, Simhadri, Krishnaswamy, and Kadekodi]{DiskANN}
Subramanya, S.~J., Devvrit, F., Simhadri, H.~V., Krishnaswamy, R., and Kadekodi, R.
\newblock Rand-nsg: Fast accurate billion-point nearest neighbor search on a single node.
\newblock In \emph{{NeurIPS}}, pp.\  13748--13758, 2019.

\bibitem[Sun et~al.(2023)Sun, Guo, and Kumar]{SOAR}
Sun, P., Guo, R., and Kumar, S.
\newblock Automating nearest neighbor search configuration with constrained optimization.
\newblock \emph{arXiv preprint arXiv:2301.01702}, 2023.

\bibitem[Xu et~al.(2021)Xu, Wang, Wang, and Ma]{TOGG-KMC}
Xu, X., Wang, M., Wang, Y., and Ma, D.
\newblock Two-stage routing with optimized guided search and greedy algorithm on proximity graph.
\newblock \emph{Knowl. Based Syst.}, 229:\penalty0 107305, 2021.

\bibitem[Zilliz(2023)]{glass}
Zilliz.
\newblock Graph library for approximate similarity search.
\newblock \url{https://github.com/zilliztech/pyglass}, 2023.

\end{thebibliography}
